\documentclass[10pt,twocolumn,letterpaper]{article}
\usepackage[accsupp]{axessibility}

\usepackage[algorithms]{wacv}
\usepackage{times}
\usepackage{epsfig}
\usepackage{graphicx}
\usepackage{amsmath}
\usepackage{amssymb}
\usepackage{booktabs}
\usepackage{comment}
\usepackage{soul}

\usepackage{algpseudocode,algorithm,algorithmicx}
\usepackage{caption}
\usepackage{xspace}
\usepackage{siunitx}

\usepackage{tikz}
\usetikzlibrary{spy,backgrounds}

\usepackage{array}
\usepackage{multirow}
\newcolumntype{M}[1]{>{\centering\arraybackslash}m{#1}}

\usepackage[pagebackref,breaklinks,colorlinks]{hyperref}

\usepackage[capitalize]{cleveref}
\crefname{section}{Sec.}{Secs.}
\Crefname{section}{Section}{Sections}
\Crefname{table}{Table}{Tables}
\crefname{table}{Tab.}{Tabs.}

\def\wacvPaperID{2807} %
\def\confName{WACV}
\def\confYear{2025}

\newcommand{\ours}{GStex\xspace}

\newcommand{\opacity}{\alpha}
\newcommand{\scale}{\boldsymbol{\sigma}}
\newcommand{\position}{\boldsymbol{\mu}}
\newcommand{\rotation}{\mathbf{R}}
\newcommand{\ra}{\mathbf{v_1}}
\newcommand{\rb}{\mathbf{v_2}}
\newcommand{\rag}{\mathbf{v}_{\mathbf{1}i}}
\newcommand{\rbg}{\mathbf{v}_{\mathbf{2}i}}
\newcommand{\rc}{\mathbf{v_3}}
\newcommand{\radiance}{\mathbf{c}}
\newcommand{\ray}{\mathbf{r}}
\newcommand{\origin}{\mathbf{o}}
\newcommand{\rayparam}{t}
\newcommand{\dir}{\mathbf{d}}
\newcommand{\gaussian}{G}
\newcommand{\intersect}{\mathbf{p}}
\newcommand{\normal}{\mathbf{n}}
\newcommand{\transmittance}{T}
\newcommand{\radiancetexture}{\radiance^{\mathcal{T}}}
\newcommand{\radiancesh}{\radiance^{\text{SH}}}
\newcommand{\texelsize}{\mathcal{T}_\text{size}}
\newcommand{\texturedgaussian}{G^\mathcal{T}}

\newcommand{\editadd}[1]{#1}
\newcommand{\editremove}[1]{}
\newlength\tmpcolwidth

\begin{document}

\title{\ours: Per-Primitive Texturing of 2D Gaussian Splatting\\for Decoupled Appearance and Geometry Modeling}

\author{Victor Rong$^{1,2}$
\space\space
Jingxiang Chen$^{1}$
\space\space
Sherwin Bahmani$^{1,2}$
\space\space
Kiriakos N. Kutulakos$^{1,2}$
\space\space
David B. Lindell$^{1,2}$\\
\small{\textnormal{$^{1}$University of Toronto\space\space$^{2}$Vector Institute}}\\
\url{https://lessvrong.com/cs/gstex}
}

\twocolumn[{
\maketitle
 \begin{center}
    \centering
\includegraphics[width=1\textwidth]{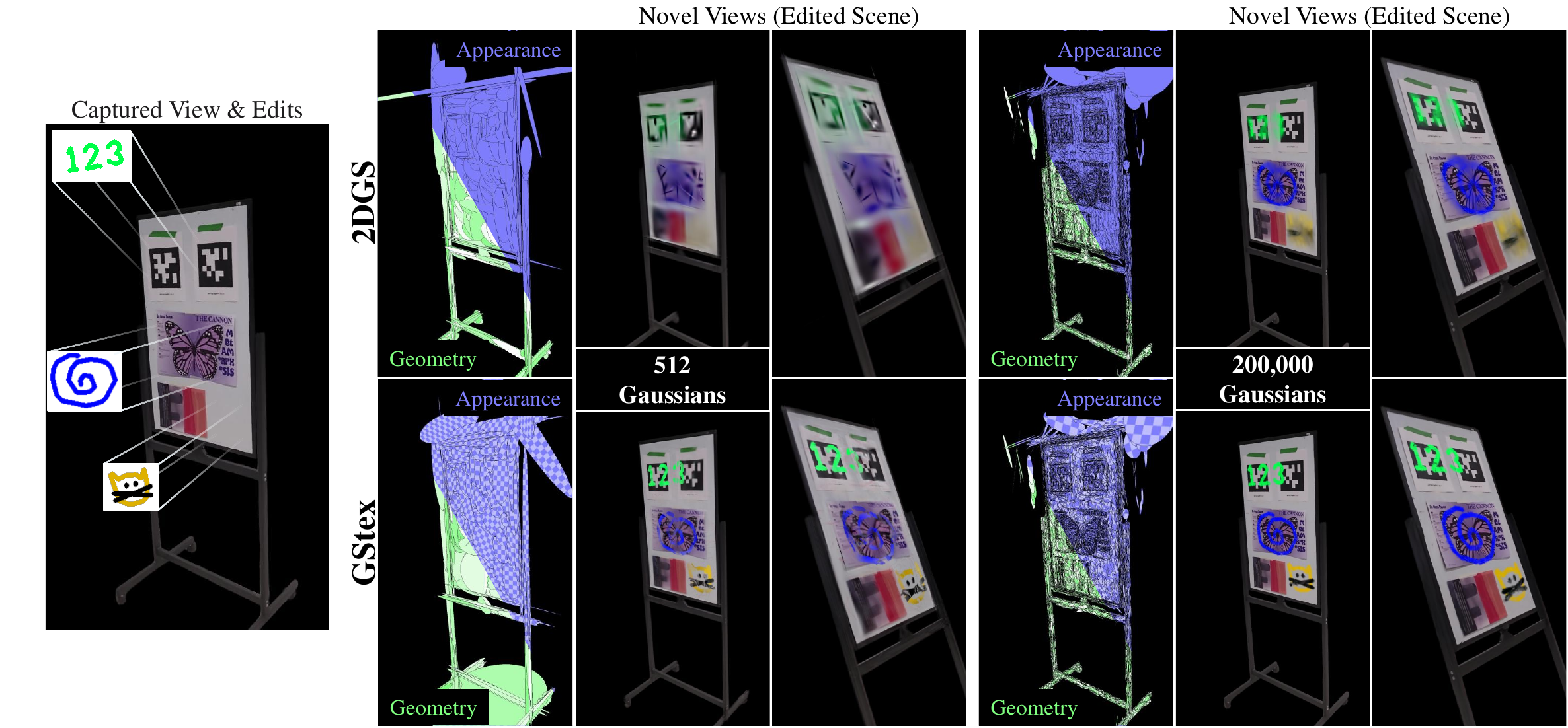}
\captionof{figure}{\textbf{\ours}. We apply 2D Gaussian splatting (2DGS)~\cite{kerbl20233d} and \ours to reconstruct a planar textured scene (a whiteboard; left column) and edit its appearance. (\textbf{Col.\ 2}) In 2DGS, each 2D Gaussian encodes both appearance (blue) and geometry (green), and so these properties are coupled. Our approach decouples appearance and geometry through use of per-Gaussian texture maps. Each tile of the blue checkerboard pattern represents $10 \times 10$ texels. For visualization purposes, we show Gaussians with opacity $>0.5$. (\textbf{Cols.\ 3--4}) When rendering the edited scene from novel views, our representation results in sharper details because it models features that are smaller in size than a single Gaussian. (\textbf{Cols.\ 5--7}) We find that \ours renders edited textures more accurately than 2DGS, even when using 200,000 Gaussians.}
\label{fig:teaser}
\end{center}

}]

\begin{abstract}
Gaussian splatting has demonstrated excellent performance for view synthesis and scene reconstruction. 
The representation achieves photorealistic quality by optimizing the position, scale, color, and opacity of thousands to millions of 2D or 3D Gaussian primitives within a scene.  
However, since each Gaussian primitive encodes both appearance and geometry, these attributes are strongly coupled---thus, high-fidelity appearance modeling requires a large number of Gaussian primitives, even when the scene geometry is simple (e.g., for a textured planar surface).
We propose to texture each 2D Gaussian primitive so that even a single Gaussian can be used to capture appearance details. By employing per-primitive texturing, our appearance representation is agnostic to the topology and complexity of the scene's geometry. We show that our approach, \ours, yields improved visual quality over prior work in texturing Gaussian splats. Furthermore, we demonstrate that our decoupling enables improved novel view synthesis performance compared to 2D Gaussian splatting when reducing the number of Gaussian primitives, and that \ours can be used for scene appearance editing and re-texturing.
\end{abstract}

\section{Introduction}

Gaussian splatting is an effective representation for novel view synthesis, achieving fast training and rendering speeds as well as photorealistic visual quality~\cite{kerbl20233d,huang2023neural}. 
Using a set of posed images of a scene, it efficiently optimizes spatial features (3D position, orientation, scale) and appearance features (view-dependent color) of thousands to millions of Gaussian primitives to accurately model scene appearance.
Explicit scene modeling using Gaussians has proven to be very powerful; Gaussians can be rendered efficiently, and the smooth decay of the Gaussian function itself enables effective optimization of position and other features via gradient descent. 

While Gaussian splatting has proved to be a powerful representation for numerous tasks---including scene reconstruction~\cite{huang2023neural}, physics simulation~\cite{xie2023physgaussian,feng2024gaussian}, and more~\cite{wu20234d}---this commonly used formulation has a major limitation: modeling of appearance and geometry is inherently coupled.
Since each Gaussian represents a single view-dependent color, many Gaussians are needed to model regions of a scene with high-frequency texture or spatially varying reflectance---even if the underlying geometry is simple. Moreover, this coupling makes it challenging to perform edits to the scene appearance that are finer than the size of an individual Gaussian, as demonstrated in Figure~\ref{fig:teaser}.

Conventional graphics representations have a clear separation between appearance and geometry~\cite{shirley2009fundamentals}. 
For example, vertices and faces of a triangle mesh are used to parameterize geometry, and a 2D texture map is used to parameterize appearance.
The geometry and appearance are coupled by associating each face of a mesh with its corresponding texture via UV coordinates. 
Thus, the appearance of a conventional 3D representation can be edited without affecting geometry by directly manipulating the texture map.

Recent work similarly endeavors to decouple the appearance and geometry of Gaussian splatting. Specifically, Xu et al.~\cite{xu2024texture} use a texture mapping method that requires optimizing a surface parameterization.  
The approach assumes that the scene's Gaussians lie on a surface that is topologically equivalent to a sphere; however, this assumption breaks down for scenes with complex geometry. 
Additionally, the constrained surface parameterization leads to a loss of visual quality. 
While surface parameterization is indeed the dominant paradigm for texturing triangle meshes \cite{poranne2017autocuts}, other representations such as subdivision surfaces have found success with per-primitive maps \cite{lee2000displaced,burley2008ptex}. 
We contend that per-primitive texturing integrates better with Gaussian splats, as it permits scenes of arbitrary topology and preserves the independence between Gaussians.

Following this line of reasoning, we propose a novel Gaussian-based representation with per-Gaussian texturing that enables decoupled appearance and geometry modeling for bounded scenes. Our approach uses an initial geometric reconstruction of a scene based on 2D Gaussian primitives~\cite{huang20242d,dai2024high}.
Then, we associate a texture map with each 2D Gaussian to encode spatially varying albedo.
We optimize the texture maps by minimizing the photometric loss between captured scene images and rendered views of the scene, where we use ray-Gaussian intersections and a texture lookup step to render appearance. 
In summary, we make the following contributions.

\begin{itemize}
    \item We propose \ours, a representation and rendering approach for decoupled appearance and geometry modeling based on textured 2D Gaussian primitives.
    \item We evaluate our representation for scene reconstruction and novel view synthesis on a range of synthetic and captured scenes. 
    \item We demonstrate view synthesis performance on par with standard Gaussian splatting methods while significantly improving over alternative approaches for texturing Gaussian splats. Further, we demonstrate that \ours improves perceptual quality over 2D Gaussian splatting~\cite{huang20242d} when reducing the number of Gaussians.
    \item We show additional capabilities enabled by \editremove{our representation}\editadd{\ours} such as making fine-grained edits to appearance through texture painting and procedural re-texturing.
\end{itemize}

\section{Related Work}
Our work is related to representations for appearance and geometry used in computer graphics, including conventional 3D representations, learned 3D representations, and recently proposed methods based on Gaussian splatting.

\begin{figure*}[t!]
\begin{center}
\includegraphics[width=\linewidth]{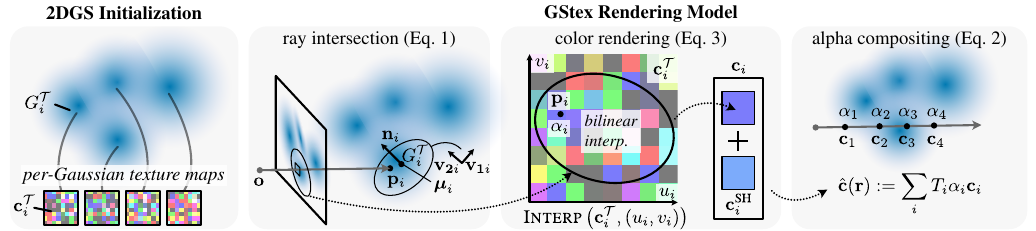}
\end{center}
   \caption{\textbf{Overview of the \ours rendering model.} We initialize the model using 2DGS~\cite{huang20242d} and assign an RGB texture map $\radiancetexture_i$ to each 2D Gaussian primitive $\texturedgaussian_i$. We render a pixel by casting a ray into the scene; the ray passes through a textured 2D Gaussian that is oriented along $\rag$ and $\rbg$ with position $\position_i$ and normal $\normal_i$. The color $\radiance_i$ is given by mapping the intersection point to the uv coordinates of the Gaussian's texture map $\radiancetexture_i$, bilinearly interpolating the resulting texture value, and adding it to the Gaussian's view-dependent color component $\radiancesh_i$ (parameterized by spherical harmonics). The alpha value $\opacity_i$ is given by evaluating the Gaussian function at the intersection point. 
Finally, we alpha composite the rendered colors of each intersected Gaussian, resulting in the rendered pixel color $\hat{\radiance}$.}
\label{fig:render}
\end{figure*}

\paragraph{Conventional 3D representations.}
It is common for 3D representations in computer graphics to decouple geometry and texture. For example, a textured triangle mesh represents geometry using a set of vertices, connected by edges to form triangular faces that approximate a continuous surface; each face of a triangle is mapped to a 2D representation of appearance called a texture map~\cite{botsch2006geometric}.
Finding surface parameterizations of a 3D object is a challenging topological problem, and there is a large body of work tackling specific issues associated with it (e.g., identifying cuts or finding low-distortion mappings~\cite{sorkine2002bounded,poranne2017autocuts,li2018optcuts,xiang2021neutex,srinivasan2023nuvo,levy2002least}). Due to the intrinsic difficulties of surface parameterization, non-parametric texturing has been proposed \cite{yuksel2019rethinking}. These approaches, such as Renderman's Ptex \cite{burley2008ptex,christensen2018renderman} and mesh colors \cite{yuksel2010mesh,yuksel2017mesh} use per-primitive textures, circumventing \editremove{all }topological issues entirely. Though non-parametric textures do not offer editing in UV space, projective painting~\cite{yang2022neumesh,everitt2001projective,segal1992fast,debevec1998efficient} and procedural texture synthesis ~\cite{peachey1985solid,perlin1985image,ebert2002texturing} are sufficient for production-level quality. Inspired by this line of work, our method proposes textured 2D Gaussian primitives, which we efficiently optimize by building on recent work that leverages automatic differentiation and GPU-accelerated rasterization~\cite{huang20242d,kerbl20233d}. Though 2D point-based primitives have long been known in computer graphics~\cite{pfister2000surfels,carceroni2002multi,furukawa2009accurate,henry2012rgb,schops2019surfelmeshing}, they have only recently been used in deep learning frameworks~\cite{mihajlovic2021deepsurfels,huang20242d,dai2024high,gao2023surfelnerf}.%

\paragraph{Learned 3D representations.}
Recent methods seek to learn 3D representations from multi-view images~\cite{mildenhall2021nerf}, point clouds~\cite{deng2022depth,rematas2022urban}, depth maps~\cite{azinovic2022neural}, and more~\cite{attal2021torf,malik2024transient}. 
Such methods represent 3D geometry using signed distance~\cite{curless1996volumetric,park2019deepsdf}, occupancy~\cite{mescheder2019occupancy,saito2019pifu}, or a volumetric representation~\cite{mildenhall2021nerf,lombardi2019neural}, and there are varying approaches to parameterization.
For example, 3D representations can be parameterized using fully-connected networks~\cite{sitzmann2020implicit,tancik2020fourier,barron2021mip}, voxel grids~\cite{yu2021plenoctrees,fridovich2022plenoxels,sun2022direct,newcombe2011kinectfusion}, combinations of feature grids and neural networks~\cite{liu2020neural,martel2021acorn,chen2022tensorf,lombardi2021mixture,sitzmann2019deepvoxels,takikawa2021neural,muller2022instant}\editremove{, learned feature representations based on hash encoding, or}\editadd{, or point-based neural features~\cite{xu2022point,chen2023neurbf}}.
A number of works texture these representations by optimizing a surface parameterization~\cite{xiang2021neutex,srinivasan2023nuvo,das2022learning}, while other works use hybrid methods to disentangle appearance and geometry \cite{yang2022neumesh,groueix2018papier}.
Our approach similarly learns a 3D representation of appearance and geometry, but we represent these explicitly by optimizing textured surfaces comprising 2D Gaussian splats.

\paragraph{Gaussian splatting.}
Our approach builds on Gaussian splatting~\cite{kerbl20233d}, which enables efficient scene reconstruction and rendering for novel view synthesis using fast splatting and rasterization~\cite{zwicker2001surface,zwicker2002ewa}. There are a great number of follow-ups to Gaussian splatting and we refer to Chen et al.~\cite{chen2024survey} for an overview of this rapidly evolving area. %
Most relevant to our work are 2D Gaussian-based methods that target high-quality appearance modeling and surface reconstruction~\cite{dai2024high,huang20242d}, and approaches for appearance editing. Among editing approaches, the vast majority~\cite{zhou2024feature,chen2024gaussianeditor,zhuang2024tip,palandra2024gsedit,wu2024gaussctrl} focus on semantic edits. However, they cannot perform fine-grained edits with details smaller than a Gaussian. Closest to our work is that of Xu et al.~\cite{xu2024texture}, which proposes to learn a surface parameterization for textured Gaussians. In contrast, \ours associates independent texture maps to each Gaussian, bypassing topological issues and achieving better visual quality. 

\subsection{Preliminaries: 2D Gaussian Splatting}
We provide a brief overview of\editremove{ the mathematical model and rendering approach of} 2DGS.
Each Gaussian $\gaussian = \{\radiance, \opacity, \rotation, \position, \scale \}$ is parameterized by its radiance $\radiance \in \mathbb{R}^{3S}$ (where  $S$ is the number of spherical harmonics coefficients used to model view-dependent appearance), opacity $\opacity \in \mathbb{R}$, rotation matrix $\rotation = [\ra, \rb, \rc] \in \mathbb{SO}_3$, mean position $\position \in \mathbb{R}^3$, and scale $\scale \in \mathbb{R}^3$. 
The normal vector of the 2D Gaussian is given as $\normal:=\rc$ and we set the corresponding entry of the scale vector to zero. 

Unlike \editremove{3DGS}\editadd{3D Gaussian splatting (3DGS)~\cite{kerbl20233d}}, in which each Gaussian is splatted using an approximate ellipsoidal weighted average scheme, 2DGS processes Gaussians through ray-primitive intersections. 
More precisely, the intersection between a camera ray $\ray := \origin + \rayparam \dir$ and the $i$th Gaussian $\gaussian_i$ at a $3$D point $\intersect_i$ is given as 
\begin{equation}
    \intersect_i(\ray, \gaussian_i) := \origin + \rayparam \dir,\quad \rayparam = \frac{\normal_i \cdot (\position_i - \origin)}{\normal_i \cdot \dir}.
    \label{eq:intersect}
\end{equation}

The rendered radiance for each ray $\hat{\radiance}(\ray)$ is computed by alpha compositing the radiances of each intersected Gaussian. 
This\editremove{ operation} follows the volume rendering formula~\cite{mildenhall2021nerf,max1995optical}: 
\begin{equation}
    \hat{\radiance}(\ray) := \sum_{i} \transmittance_i \opacity_i(\intersect_i) \radiance_i(\intersect_i, \dir),\quad \transmittance_i := \prod_{j}^{i-1} (1 - \opacity_i(\intersect_i)).
    \label{eq:composite}
\end{equation}
Here, the alpha values $\opacity_i(\intersect_i)$ for the $i$th Gaussian are computed by sampling the corresponding Gaussian function at the intersection point $\intersect_i$.
Similarly, $\radiance_i(\dir)$ is the radiance corresponding to Gaussian $\gaussian_i$ viewed from direction $\dir$.

\section{\ours}
We decouple appearance and geometry in the Gaussian splatting representation using per-Gaussian texture maps, which are optimized jointly with the Gaussian parameters for novel view synthesis. 
In the following, we describe the rendering model and optimization of the representation.

\subsection{Rendering Model}
\label{sub:model}
We provide an overview of our rendering model below and in Figure~\ref{fig:render}. Our approach modifies the radiance parameters $\radiance$ of 2D Gaussian splatting while keeping the other Gaussian parameters the same as described previously. 
Specifically, we decompose the radiance $\radiance$ of a Gaussian into two components~\cite{xu2024texture}. The first is a diffuse texture $\radiancetexture$, which varies spatially across the corresponding Gaussian and is represented as an array of RGB values. The second is a residual term $\radiancesh$ that models the Gaussian's view-dependent appearance using spherical harmonics coefficients. Note that we do not include the constant coefficient in $\radiancesh$.
Our proposed textured 2D Gaussian is thus given as $\texturedgaussian = \{\radiancetexture, \radiancesh, \opacity, \rotation, \position, \scale \}$.

The rendered radiance of the $i$th textured 2D Gaussian varies depending on both the spatial position $\intersect_i$ and incident ray direction $\dir$. 
More formally, we render the radiance as 
\begin{equation}
    \radiance_i(\intersect_i, \dir) = \textsc{Interp}\left(\radiancetexture_i, (u_i(\intersect), v_i(\intersect))\right) + \textsc{SH}\left(\radiancesh_i, \dir\right).
    \label{eq:render}
\end{equation} 
Here, \textsc{Interp} is the bilinear interpolation function and $\radiancetexture_i$ is a $U_i \times V_i$ array of RGB texels, corresponding to the $i$th 2D Gaussian. 
We map the ray intersection point $\intersect_i$ to the array coordinates $(u_i(\intersect), v_i(\intersect)) \in [0, U_i) \times [0, V_i)$ used to \editremove{interpolate}\editadd{query} the texture value.
For the view-dependent term\editremove{ we have} $\radiancesh_i\in\mathbb{R}^{3(S-1)}$,\editremove{ where} we use the same spherical harmonic encoding for view-dependent appearance as 2DGS and calculate radiance from \editremove{the coefficients}\editadd{$\radiancesh_i$} and $\dir$ using $\textsc{SH}$ (a function that encodes RGB spherical harmonics). 
The representation can be seen as a generalization of 2DGS where $\radiancetexture_i$ is a spatially varying version of the zeroth-order spherical harmonic.

The mapping is \editremove{constructed}\editadd{done} so that each texel corresponds to a square in world space of the same size $\texelsize \times \texelsize$. 
To map the ray intersection point $\intersect_i$ to the grid coordinates (i.e., $u_i(\intersect)$, $v_i(\intersect)$) we project $\intersect_i$ onto the two coordinate axes $\rag, \rbg$ of the $i$th 2D Gaussian's rotation matrix $\rotation_i$. 
Then we divide by the world-space texel size $\texelsize$ and shift appropriately.
That is,
\begin{align*}
    u_i(\intersect_i) &= (\rag \cdot \intersect_i) / \texelsize + (U_i-1) / 2,\\
    v_i(\intersect_i) &= (\rbg \cdot \intersect_i)/\texelsize + (V_i-1) / 2.
\end{align*}  
Last, the texture value is retrieved using bilinear interpolation.
We provide a pseudocode description of the rendering model in Algorithm~\ref{alg:rendering}.

\begin{algorithm}[t!]
    \caption{Rendering model \editremove{(Eq. 3)}}\label{alg:rendering}
\begin{algorithmic}[1]
\Function{Render}{}\\
\textbf{\textit{Input: }}\\
\hspace*{\algorithmicindent} $\ray := \origin + \rayparam \dir$ \Comment{camera ray} \\
\hspace*{\algorithmicindent} $\{\texturedgaussian_1, \ldots \texturedgaussian_N\}$ \Comment{textured 2D Gaussians} \\

\vspace{0.5em}
\textbf{\textit{Calculate ray-Gaussian intersections and radiances}}
\For{$i$ in $(1, \ldots, N)$}
\State $\intersect_i \leftarrow \textsc{Intersect}(\ray, \texturedgaussian_i)$
\Comment{(Eq.~\ref{eq:intersect})}
\State $\radiance_i \leftarrow \textsc{GetRadiance}(\intersect_i, \dir)$
\Comment{(Eq.~\ref{eq:render})}
\EndFor\\

\vspace{0.5em}
\textbf{\textit{Render ray color}}
\State $\hat{\radiance} \leftarrow \textsc{Composite}(\{\intersect_i\}_{i=1}^N, \{\radiance_i\}_{i=1}^N, \dir)$
\Comment{(Eq.~\ref{eq:composite})}
\State Return $\hat{\radiance}$
\EndFunction
\end{algorithmic}
\end{algorithm}

\subsection{Optimization}
\label{sub:train}

Our optimization procedure consists of two main stages: (1) obtain an initial representation using 2DGS, and (2) augment the initial representation with texture information and continue the optimization. 
In the first stage, we initialize the Gaussian positions using a sparse point cloud from COLMAP~\cite{schonberger2016structure}, and we optimize the representation for a set number of iterations following 2DGS~\cite{huang20242d}. 

After obtaining this initial reconstruction, we proceed to the second stage of optimization. 
Here,\editremove{ we set} the dimensions $U_i, V_i$ of each 2D Gaussian's texture map \editadd{are} based on the size of the Gaussian.   
Specifically, \editadd{we set }the texture map dimensions \editremove{are allocated }as $U_i = \lceil 6\scale_{1i}/\texelsize \rceil$ and $V_i = \lceil 6\scale_{2i}/\texelsize \rceil$, where we choose plus-minus three standard deviations to cover the main region of support for each Gaussian. 
As the texture maps of each Gaussian have different dimensions, we flatten them into $(U_i V_i) \times 3$ tensors and concatenate them into a single tensor, akin to a jagged array. 
Then, we set the initial values of $\radiancetexture_i$ to the radiance corresponding to the learned zeroth-order spherical harmonic coefficient for each Gaussian. The other spherical harmonic coefficients $\radiancesh_i$ are copied from the 2DGS model.
This procedure results in the initial \ours model.

We continue the optimization \editremove{after disabling}\editadd{with} the Gaussian culling and densification steps \editadd{disabled}~\cite{kerbl20233d}, which \editremove{allows finetuning of all model parameters}\editadd{lets us finetune the model parameters}, including $\radiancetexture_i$, in a straightforward fashion. During backpropagation, we apply a stop gradient so that the Gaussian positions $\position_i$ are not affected by gradients from the interpolated texture values (as the positions can be sensitive to high frequency texture variations). We evaluate the effect of the stop gradient in the supplementary.

Since the size of the 2D Gaussians changes during the second stage of optimization, we periodically update the number of texels allocated to each Gaussian\editremove{ and re-initialize the texture maps}. This procedure prevents texels from becoming stretched or distorted. The values in the \editremove{re-initialized}\editadd{resized} texture maps are obtained by bilinearly interpolating the previous texture maps, which keeps the model's appearance consistent and prevents \editremove{disrupting }the optimization \editadd{from being disrupted}.

Following \editremove{3D Gaussian splatting (3DGS)}\editadd{3DGS}~\cite{kerbl20233d}, both stages of the optimization use the following loss function
\begin{equation*}
    \mathcal{L} = \sum_\ray(1 - \lambda) \mathcal{L}_1\left(\hat{\radiance}(\ray), \radiance(\ray)\right) + \lambda\mathcal{L}_{\text{D-SSIM}}\left(\hat{\radiance}(\ray), \radiance(\ray)\right),
    \label{eq:loss}
\end{equation*}
where we compute the L1 difference and the D-SSIM between the rendered radiance $\hat{\radiance}(\ray)$ and ground truth radiance $\radiance(\ray)$.
The weighting term $\lambda$ is set to 0.2, as in 3DGS.

\subsection{Appearance Editing}
\label{sec:editing}
We introduce two ways of editing the appearance of our representation based on classical texturing techniques. Further details on how we adapt these to \ours are included in the supplementary.

\paragraph{Texture painting \cite{debevec1996modeling,cook20073d}.} We support editing 3D appearance by editing the appearance of a rendered 2D image from a particular viewpoint. 
Specifically, we transfer the edited 2D image onto the \ours representation by shooting rays from the corresponding camera origin through the center of each pixel of the edited image. We associate each ray to the RGBA value $(\hat{\radiance}, \alpha)$ of the pixel it passes through, and for each texel of the \ours model, we aggregate the RGBA values of the rays that intersect it.  The new texel value is computed using a weighted combination of these aggregated RGBA values. To make the editing procedure occlusion-aware, we first calculate a depth map and only update texels corresponding to Gaussians along each ray that are close to the computed depth. We provide a mathematical description of this procedure in the supplementary.

\paragraph{Procedural textures \cite{perlin1985image,peachey1985solid}.} Procedural textures can be computed as a function of 3D position. 
We apply procedural texturing to \ours by computing the world coordinates of each texel and setting the RGB value of each texel to the corresponding output of the texturing function.

\begin{table}[t]
\caption{Metrics for novel view synthesis and rendering performance on the synthetic Blender dataset and DTU dataset preprocessed by Huang et al.~\cite{huang20242d}. We report PSNR $\uparrow$, SSIM $\uparrow$, LPIPS $\downarrow$, rendering frames per second (FPS) $\uparrow$, and memory to store parameters (Mem) $\downarrow$, averaged across all scenes. \label{tab:nvs_all}}
\vspace{-1em}
  \begin{center}
     \resizebox{\columnwidth}{!}{%
         \addtolength{\tabcolsep}{-0.43em}

\begin{tabular}{lcccccccccc}
\toprule
& \multicolumn{5}{c}{Blender \cite{mildenhall2021nerf}} & \multicolumn{5}{c}{DTU \cite{jensen2014large}}\\
Method & PSNR & SSIM & LPIPS & FPS & Mem& PSNR & SSIM & LPIPS & FPS & Mem\\ %
\cmidrule(l{2pt}r{2pt}){1-1}\cmidrule(l{2pt}r{2pt}){2-6}\cmidrule(l{2pt}r{2pt}){7-11}
3DGS & 33.34 & 0.969 & 0.030 & 345 & 65MB & 32.87 & 0.956 & 0.052 & 420 & 80MB\\
2DGS & 33.15 & 0.968 & 0.024 & 168 & 27MB & 32.22 & 0.940 & 0.090 & 75 & 41MB\\
\cmidrule(l{2pt}r{2pt}){1-1}\cmidrule(l{2pt}r{2pt}){2-6}\cmidrule(l{2pt}r{2pt}){7-11}
Texture-GS & 28.97 & 0.938 & 0.055 & 100 & 88MB & 30.53 & 0.920 & 0.083 & 53 & 88MB\\
\ours & 33.25 & 0.969 & 0.024 & 129 & 39MB & 32.87 & 0.956 & 0.038 & 111 & 53MB \\
\bottomrule
\end{tabular}
}
\end{center}
\vspace{-1em}
\end{table}

\subsection{Implementation Details}
\label{sub:impl}
We implement our method by modifying the Nerfstudio framework and gsplat codebase \cite{nerfstudio,ye2023mathematical}. For both stages of optimization we use the Adam optimizer, and we follow the learning rate schedules used in 2DGS \cite{huang20242d}. The texel values are optimized with a learning rate of $\num{1e-3}$.
We omit the normal and depth regularization terms used in 2DGS to maximize photometric quality and to be consistent with their experiments on the Blender dataset.
In all our novel view synthesis experiments we first optimize using 2DGS for 15000 iterations, and then we conduct the second optimization stage for an additional 15000 iterations.

During the second stage of optimization, we set the texel size $\texelsize$ by fixing the total number of texels, and then dividing the total area of all Gaussians by the number of texels (area is computed for each Gaussian as $36\scale_1\scale_2$). As the width and height of each texture must be integer, this procedure does not give the exact desired number of texels. 
We perform binary search to find a texel size that results in a total number of texels within $0.1\%$ of the desired quantity.

Following 3DGS, we ignore ray--Gaussian intersections with opacity below $\frac{1}{255}$. For any ray--Gaussian intersections that fall outside the support of the texture map (i.e., three standard deviations in each dimension) we use constant extrapolation to determine the texture. This extrapolation procedure has little impact on the rendered appearance because the alpha value is negligible at three standard deviations away from the mean (i.e., the Gaussian is transparent).

\section{Experiments}
We show experiments that highlight the benefits of the proposed representation for decoupled modeling of appearance and geometry without requiring a global parameterization of the surface~\cite{xu2024texture}. 
Specifically, we evaluate the performance of the proposed method for novel view synthesis with varying numbers of Gaussian primitives, and we demonstrate applications of appearance editing and re-texturing. All experiments were run on an NVIDIA RTX A6000 48 GB GPU.
\begin{figure*}[t]
	\setlength\tmpcolwidth{.24\linewidth}
	\setlength{\tabcolsep}{0pt}
    \renewcommand{\arraystretch}{0}
    \centering
	\begin{tabular}{cccc}
    Texture-GS & \ours & Texture-GS & \ours\\   
        \begin{tikzpicture}[spy using outlines={rectangle,magnification=3,size=1.75cm}]   
        	\node[anchor=south west,inner sep=0]  {\includegraphics[width=\tmpcolwidth,trim={0cm 3cm 0cm 0cm},clip]{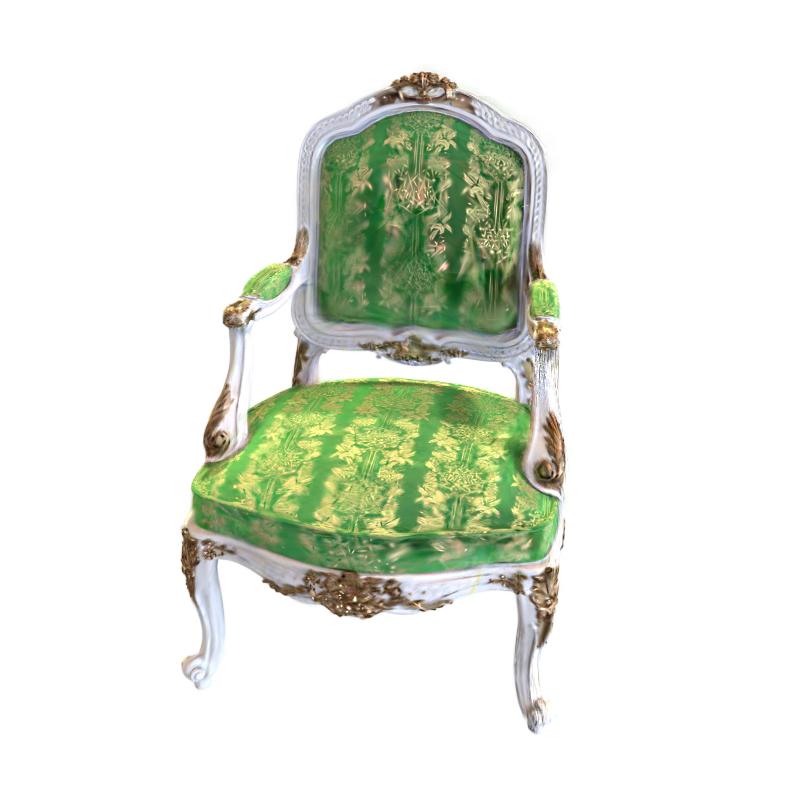}};
        	\spy[blue] on (2.2, 2.85) in node at (1cm,-0.5cm);
        	\spy[green] on (2.6, 1.05) in node at (3cm,-0.5cm);
        \end{tikzpicture}&
        \begin{tikzpicture}[spy using outlines={rectangle,magnification=3,size=1.75cm}]   
        	\node[anchor=south west,inner sep=0]  {\includegraphics[width=\tmpcolwidth,trim={0cm 3cm 0cm 0cm},clip]{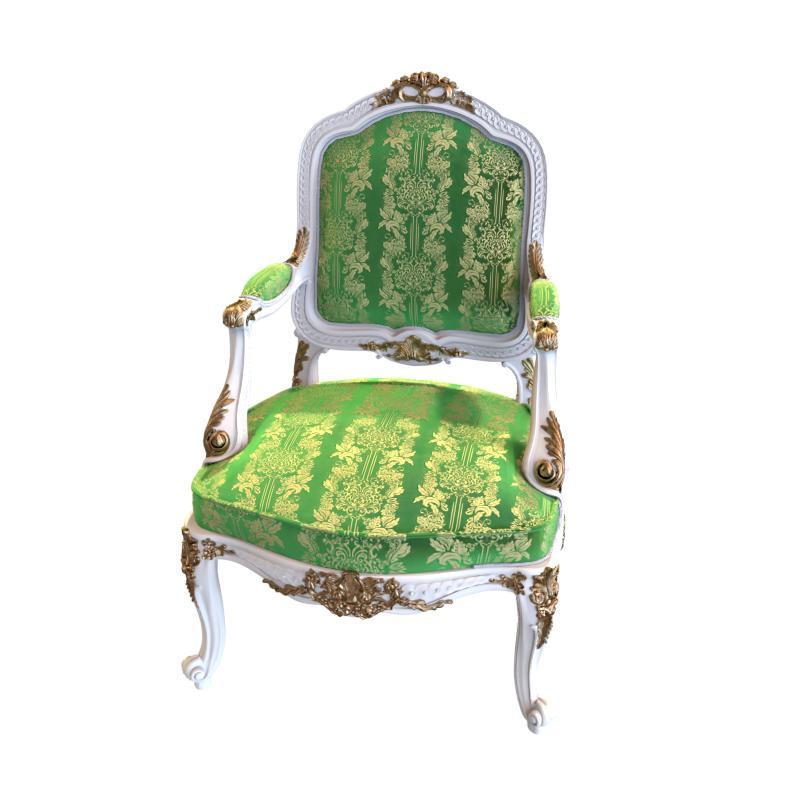}};
        	\spy[blue] on (2.2, 2.85) in node at (1cm,-0.5cm);
        	\spy[green] on (2.6, 1.05) in node at (3cm,-0.5cm);
        \end{tikzpicture}&
        \begin{tikzpicture}[spy using outlines={rectangle,magnification=3,size=1.75cm}]   
        	\node[anchor=south west,inner sep=0]  {\includegraphics[width=\tmpcolwidth,trim={0cm 0cm 0cm 6cm},clip]{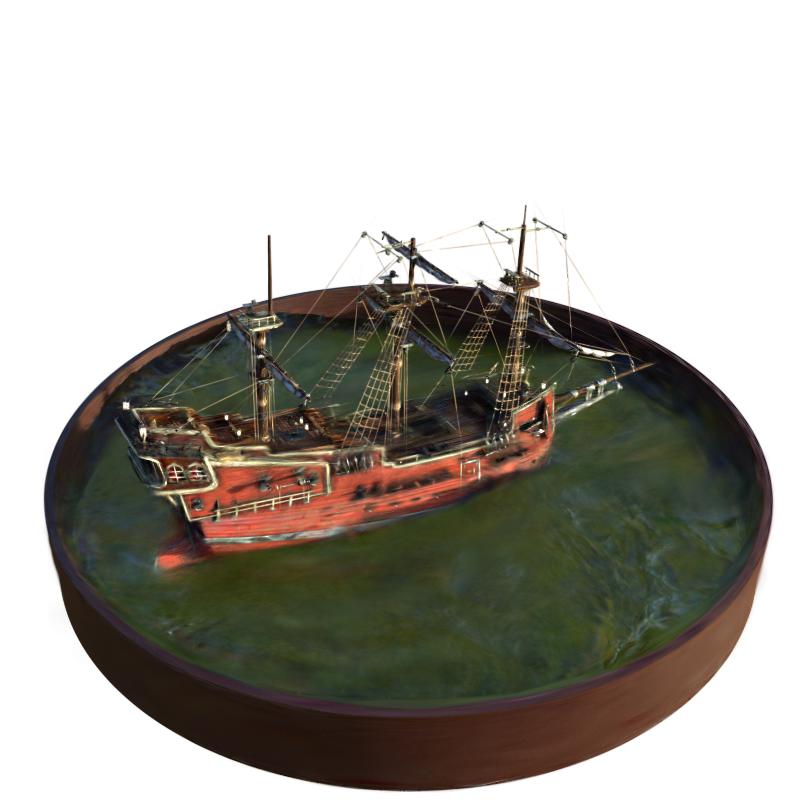}};
        	\spy[blue] on (1.0, 1.35) in node at (1cm,-0.5cm);
        	\spy[green] on (2.4, 1.75) in node at (3cm,-0.5cm);
        \end{tikzpicture}&
        \begin{tikzpicture}[spy using outlines={rectangle,magnification=3,size=1.75cm}]   
        	\node[anchor=south west,inner sep=0]  {\includegraphics[width=\tmpcolwidth,trim={0cm 0cm 0cm 6cm},clip]{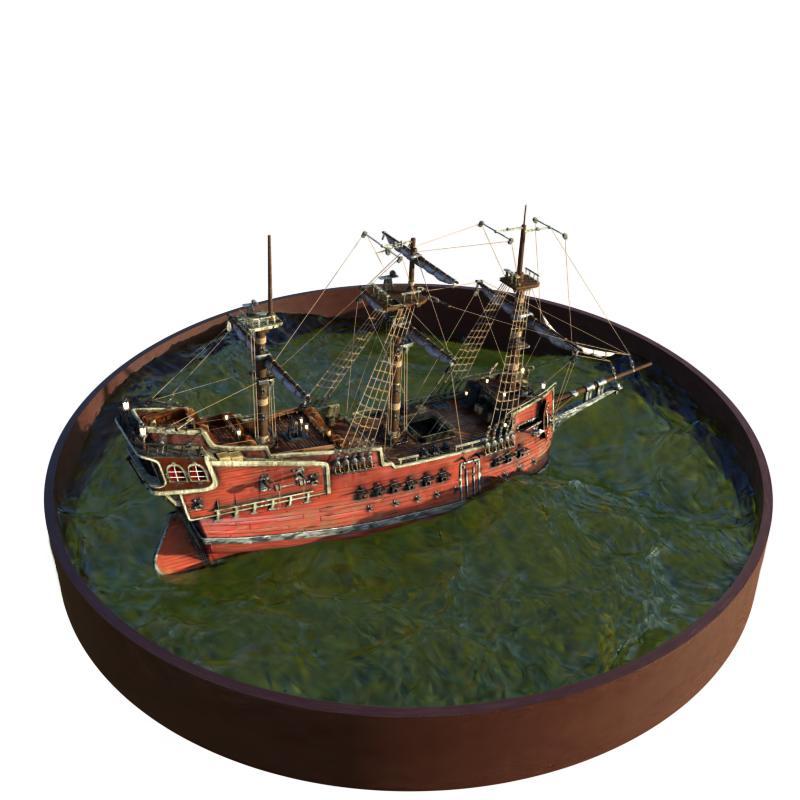}};
        	\spy[blue] on (1.0, 1.35) in node at (1cm,-0.5cm);
        	\spy[green] on (2.4, 1.75) in node at (3cm,-0.5cm);
        \end{tikzpicture}\\
	\end{tabular}
    \caption{\textbf{Comparison to Texture-GS.} We compare renders of our method and Texture-GS~\cite{xu2024texture} on scenes from the Blender dataset, which generally have finer geometry compared to the DTU dataset. High-frequency scene textures are rendered more clearly using \ours compared to Texture-GS.\label{fig:texture_gs_comparison}}
\end{figure*}

\begin{figure}[t]
\setlength{\tmpcolwidth}{1.65in}
\setlength{\tabcolsep}{0pt}
\renewcommand{\arraystretch}{0}
\centering
\includegraphics[width=3.3in,trim={5px 5px 5px 5px},clip]{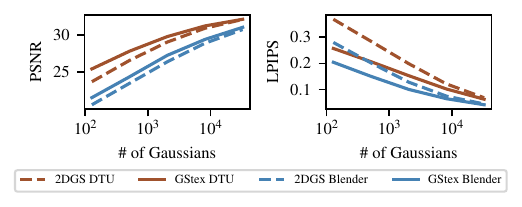}
\vspace{-2em}
\caption{\textbf{Novel view synthesis for varying numbers of Gaussians.} We perform novel view synthesis using our representation (solid) and 2DGS (dashed) initialized with varying numbers of Gaussians ranging from $128$ to $32768$. Densification and culling are turned off to ensure a consistent number of Gaussians. The plots show metrics averaged across all Blender synthetic scenes (blue) and DTU scenes (red). \label{fig:lod_plot}}
\vspace{-1em}
\end{figure}

\subsection{Novel View Synthesis}
We compare the performance of \ours for novel view synthesis to 3DGS~\cite{kerbl20233d}, 2DGS~\cite{huang20242d}, and Texture-GS~\cite{xu2024texture} on the Blender~\cite{mildenhall2021nerf} dataset and the COLMAP-preprocessed DTU dataset ~\cite{huang20242d,jensen2014large}.

\paragraph{Dataset and baselines.}
For the Blender dataset, we use the provided train/test split at the original $800 \times 800$ resolution with a white background.
We initialize the Gaussians in each method by uniformly sampling $100000$ points in $[-1.3, 1.3]^3$ and uniformly sampling random RGB color values.
For the DTU dataset, we use the COLMAP dataset train/test split suggested by Barron et al.~\cite{barron2021mip}, selecting every eighth image as a test image. 
Following Huang et al.~\cite{huang20242d}, we downsample the images by a factor of $2$ (to a resolution of $777 \times 581$). We also use the provided alpha values to set the background to black. 
The COLMAP point clouds provided by Huang et al.~\cite{huang20242d} are used for initialization.
For all methods, we evaluate using the codebases released by the corresponding authors (see the supplementary for additional details), and we use the recommended densification and culling hyperparameters.
Texture-GS~\cite{xu2024texture}, which optimizes both a UV-parameterization and associated texture, suggests a texture size of $1024 \times 1024$, and so we evaluate \ours with $10^6$ texels to ensure a fair comparison.

\begin{figure}
\setlength{\tmpcolwidth}{1.65in}
\setlength{\tabcolsep}{0pt}
\renewcommand{\arraystretch}{0}
\centering
\includegraphics[width=3.3in,trim={5px 5px 5px 5px},clip]{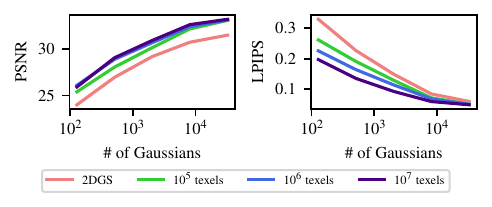}
\vspace{-2em}
\caption{\textbf{Ablation over number of texels and Gaussians.} We evaluate the effect of the number of texels as the number of Gaussians varies. Densification and culling are turned off to ensure a consistent number of Gaussians. This experiment is performed on the \textit{whiteboard} scene (Figure~\ref{fig:teaser}). \label{fig:sweep_plot}}
\end{figure}

We train our approach and 2DGS such that the initial iterations of optimization are shared. 
More precisely, during the 2DGS experiments, we export the Gaussian model after 15,000 iterations and 30,000 iterations of optimization. The result of 2DGS after 15,000 iterations is used to initialize the \ours model. We then train our approach for another 15,000 iterations, as described in Section~\ref{sub:impl}. This yields two models, 2DGS and \ours, each having trained for 30,000 iterations, as is commonly done in Gaussian splatting evaluations \cite{kerbl20233d}. For both models, densification and culling is turned off after iteration 15,000, and so the number of Gaussians does not change \editremove{in the last 15,000 iterations}\editadd{afterwards}. As a result, the number of Gaussians used in each 2DGS and \ours \editremove{scene}\editadd{model} is the same. \editadd{The numbers of Gaussians for each method and scene is included in the supplementary.}

\begin{figure*}[!h]
    \newcommand{\vlabel}[1]{\vspace{2pt}#1\vspace{2pt}}
    \setlength\tmpcolwidth{1.0in}
	\setlength{\tabcolsep}{0pt}
    \renewcommand{\arraystretch}{1}
	\centering
	\begin{tabular}{M{0.2in}M{\tmpcolwidth}M{\tmpcolwidth}M{\tmpcolwidth}M{\tmpcolwidth}M{\tmpcolwidth}M{\tmpcolwidth}}
        & \vlabel{2DGS} & \vlabel{\ours} & \vlabel{2DGS} & \vlabel{\ours} & \vlabel{2DGS} & \vlabel{\ours}\\
        \multirow{3}{*}{\rotatebox{90}{$\xleftarrow{\makebox[1.9in]{Number of Gaussians}}$}}&
        \begin{tikzpicture}[spy using outlines={rectangle,magnification=3,size=1.0cm}]
        	\node[anchor=south west,inner sep=0]  {\includegraphics[width=\tmpcolwidth,trim={20px 80px 20px 80px},clip]{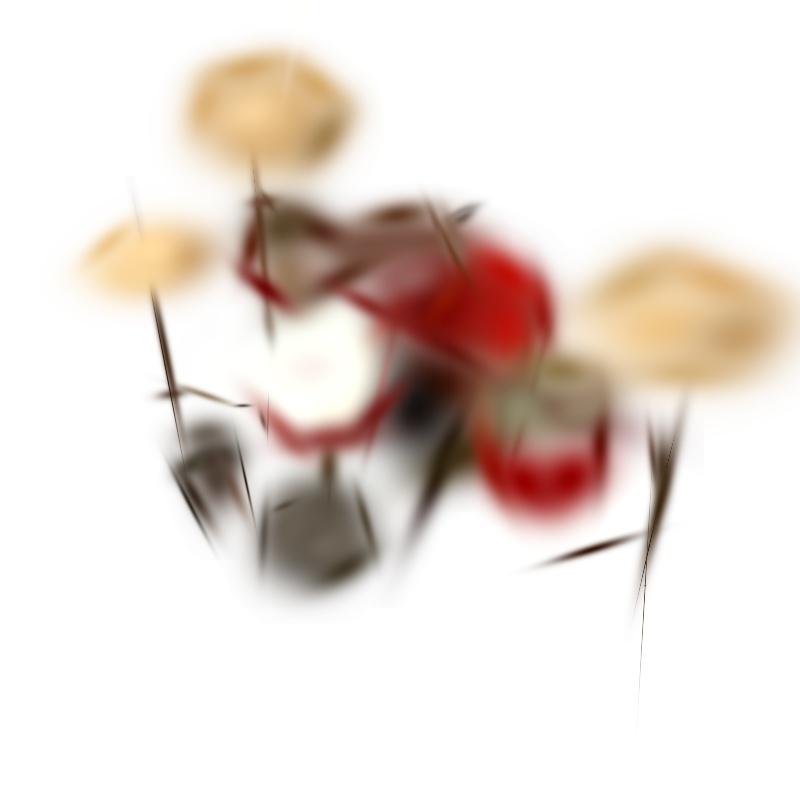}};
        	\spy[blue] on (1.2, 1.05) in node at (0.6cm,-0cm);
            \spy[green] on (2.1, 1.25) in node at (2.0cm,-0cm);
        \end{tikzpicture}
        &
        \begin{tikzpicture}[spy using outlines={rectangle,magnification=3,size=1.0cm}]
        	\node[anchor=south west,inner sep=0]  {\includegraphics[width=\tmpcolwidth,trim={20px 80px 20px 80px},clip]{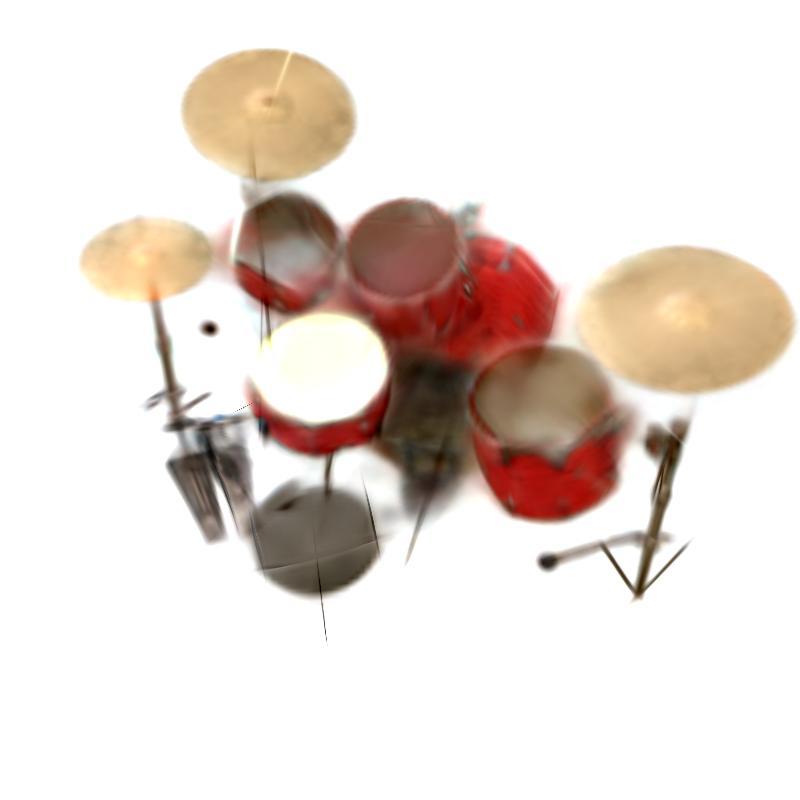}};
        	\spy[blue] on (1.2, 1.05) in node at (0.6cm,-0cm);
            \spy[green] on (2.1, 1.25) in node at (2.0cm,-0cm);
        \end{tikzpicture}
        &
        \begin{tikzpicture}[spy using outlines={rectangle,magnification=3,size=1.0cm}]
        	\node[anchor=south west,inner sep=0]  {\includegraphics[width=\tmpcolwidth,trim={20px 80px 20px 80px},clip]{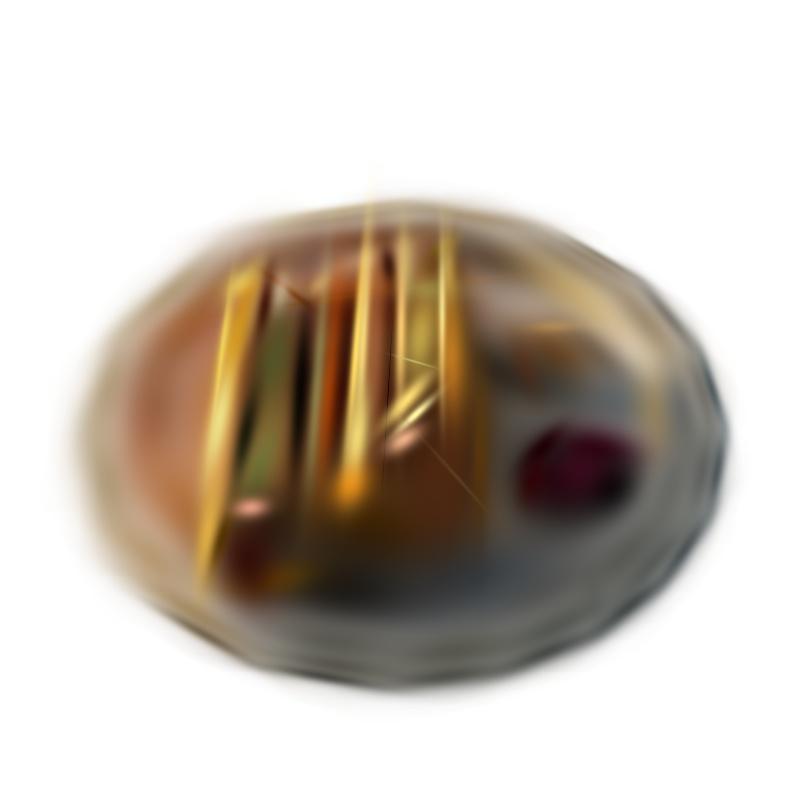}};
        	\spy[blue] on (0.95, 1.05) in node at (0.6cm,-0cm);
            \spy[green] on (1.3, 1.25) in node at (2.0cm,-0cm);
        \end{tikzpicture}
        &
        \begin{tikzpicture}[spy using outlines={rectangle,magnification=3,size=1.0cm}]
        	\node[anchor=south west,inner sep=0]  {\includegraphics[width=\tmpcolwidth,trim={20px 80px 20px 80px},clip]{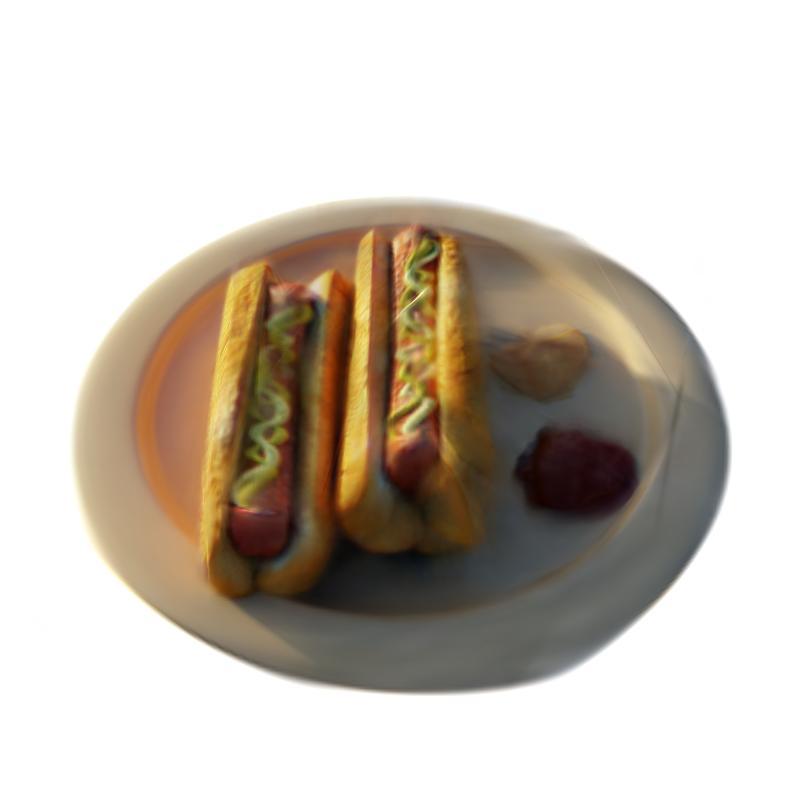}};
        	\spy[blue] on (0.95, 1.05) in node at (0.6cm,-0cm);
            \spy[green] on (1.3, 1.25) in node at (2.0cm,-0cm);
        \end{tikzpicture}
        &
        \begin{tikzpicture}[spy using outlines={rectangle,magnification=3,size=1.0cm}]
        	\node[anchor=south west,inner sep=0]  {\includegraphics[width=\tmpcolwidth,trim={20px 80px 20px 80px},clip]{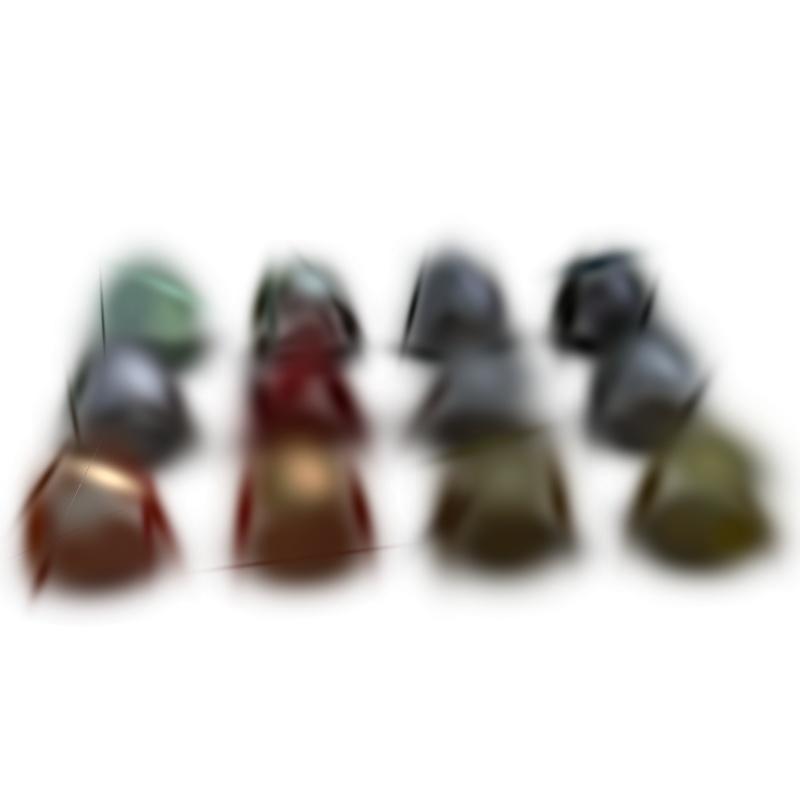}};
        	\spy[blue] on (0.95, 1.05) in node at (0.6cm,-0cm);
            \spy[green] on (1.65, 0.8) in node at (2.0cm,-0cm);
        \end{tikzpicture}
        &
        \begin{tikzpicture}[spy using outlines={rectangle,magnification=3,size=1.0cm}]
        	\node[anchor=south west,inner sep=0]  {\includegraphics[width=\tmpcolwidth,trim={20px 80px 20px 80px},clip]{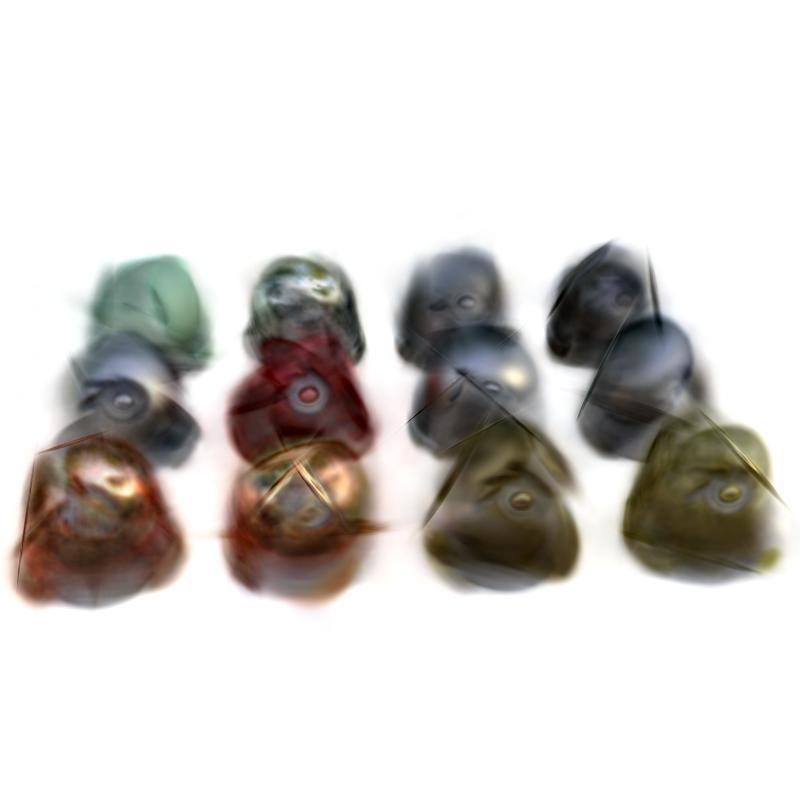}};
        	\spy[blue] on (0.95, 1.05) in node at (0.6cm,-0cm);
            \spy[green] on (1.65, 0.8) in node at (2.0cm,-0cm);
        \end{tikzpicture}\\
        &
        \begin{tikzpicture}[spy using outlines={rectangle,magnification=3,size=1.0cm}]
        	\node[anchor=south west,inner sep=0]  {\includegraphics[width=\tmpcolwidth,trim={20px 80px 20px 80px},clip]{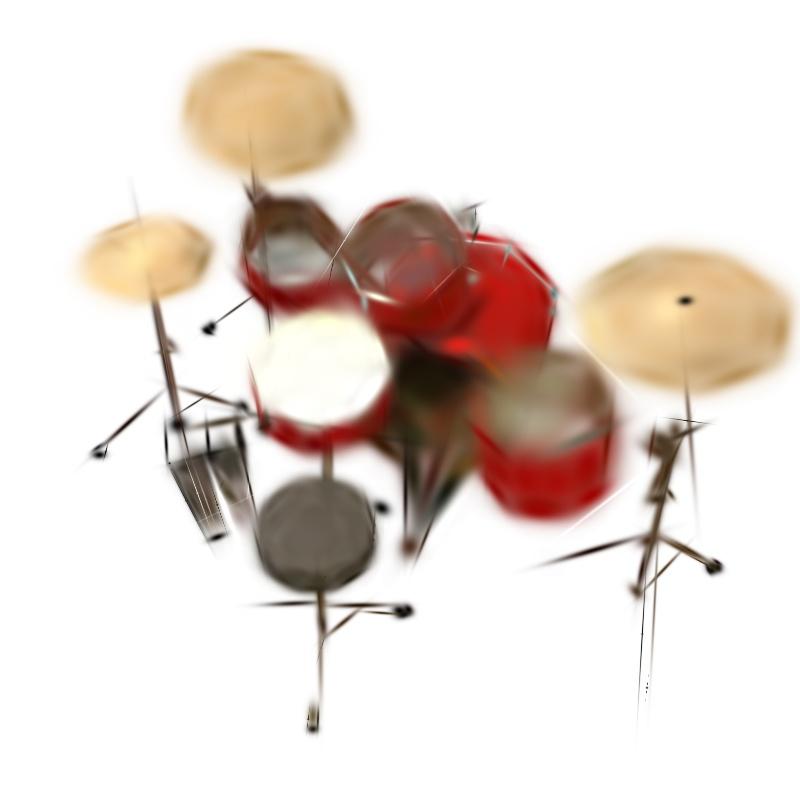}};
        	\spy[blue] on (1.2, 1.05) in node at (0.6cm,-0cm);
            \spy[green] on (2.1, 1.25) in node at (2.0cm,-0cm);
        \end{tikzpicture}
        &
        \begin{tikzpicture}[spy using outlines={rectangle,magnification=3,size=1.0cm}]
        	\node[anchor=south west,inner sep=0]  {\includegraphics[width=\tmpcolwidth,trim={20px 80px 20px 80px},clip]{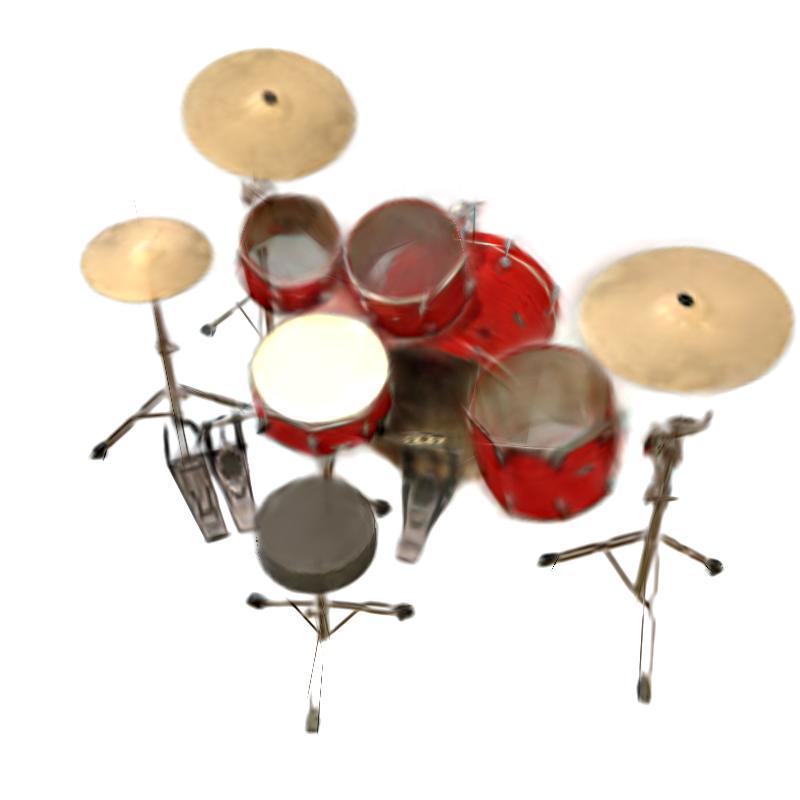}};
        	\spy[blue] on (1.2, 1.05) in node at (0.6cm,-0cm);
            \spy[green] on (2.1, 1.25) in node at (2.0cm,-0cm);
        \end{tikzpicture}
        &
        \begin{tikzpicture}[spy using outlines={rectangle,magnification=3,size=1.0cm}]
        	\node[anchor=south west,inner sep=0]  {\includegraphics[width=\tmpcolwidth,trim={20px 80px 20px 80px},clip]{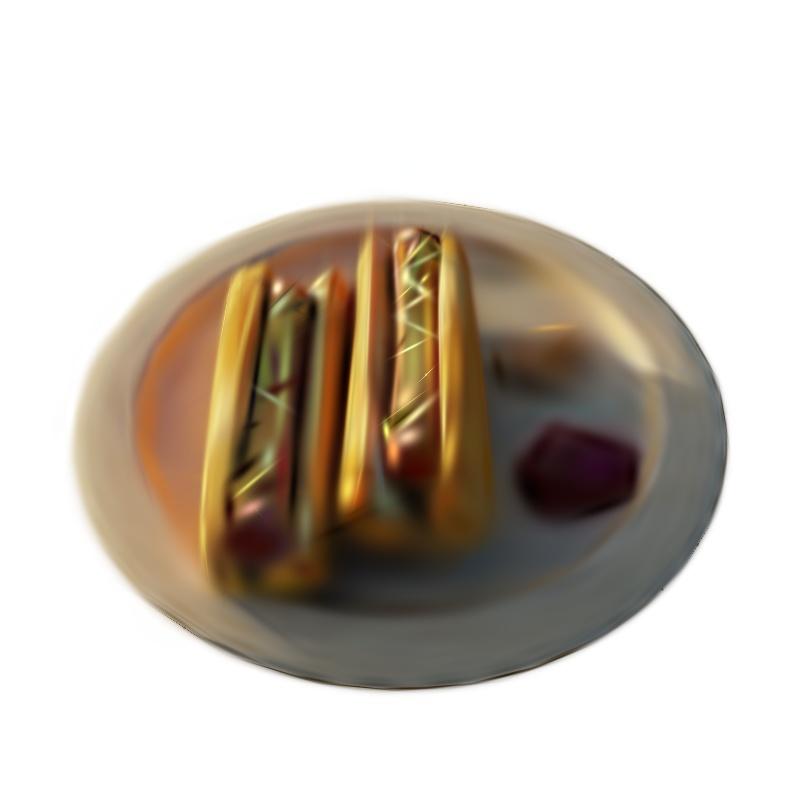}};
        	\spy[blue] on (0.95, 1.05) in node at (0.6cm,-0cm);
            \spy[green] on (1.3, 1.25) in node at (2.0cm,-0cm);
        \end{tikzpicture}
        &
        \begin{tikzpicture}[spy using outlines={rectangle,magnification=3,size=1.0cm}]
        	\node[anchor=south west,inner sep=0]  {\includegraphics[width=\tmpcolwidth,trim={20px 80px 20px 80px},clip]{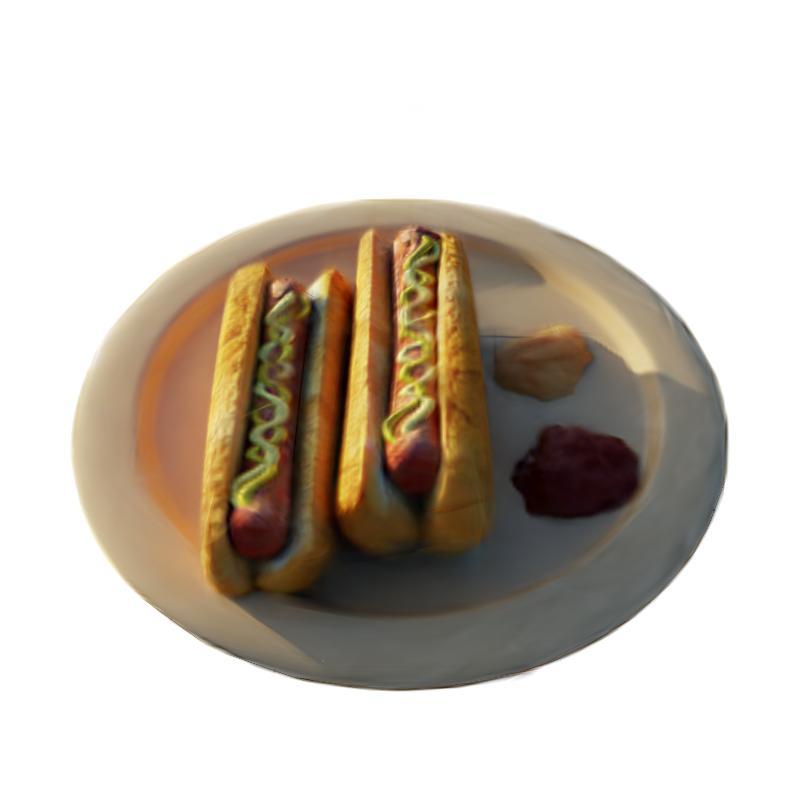}};
        	\spy[blue] on (0.95, 1.05) in node at (0.6cm,-0cm);
            \spy[green] on (1.3, 1.25) in node at (2.0cm,-0cm);
        \end{tikzpicture}
        &
        \begin{tikzpicture}[spy using outlines={rectangle,magnification=3,size=1.0cm}]
        	\node[anchor=south west,inner sep=0]  {\includegraphics[width=\tmpcolwidth,trim={20px 80px 20px 80px},clip]{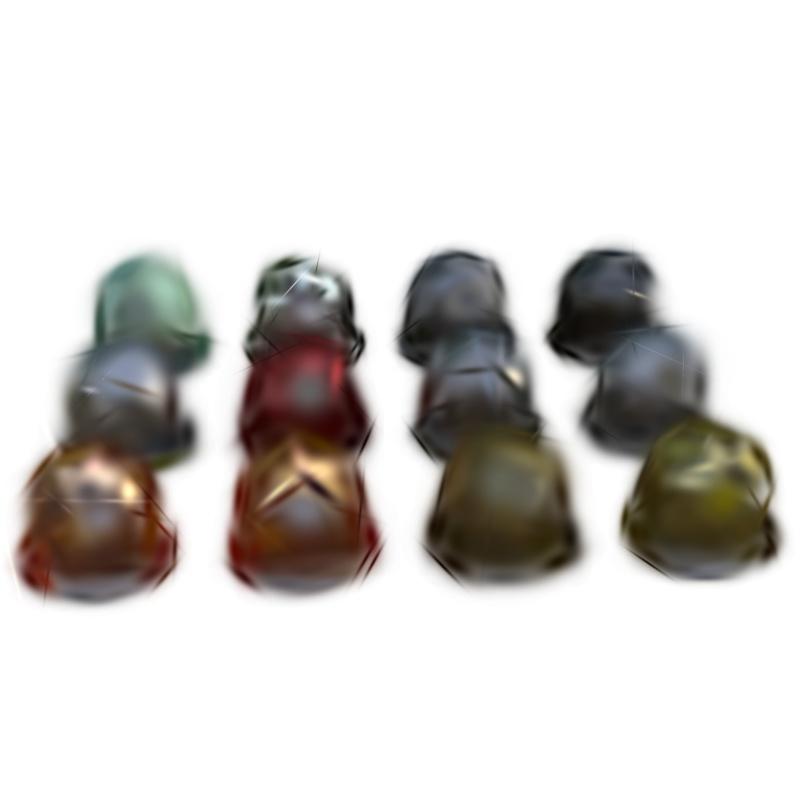}};
        	\spy[blue] on (0.95, 1.05) in node at (0.6cm,-0cm);
            \spy[green] on (1.65, 0.8) in node at (2.0cm,-0cm);
        \end{tikzpicture}
        &
        \begin{tikzpicture}[spy using outlines={rectangle,magnification=3,size=1.0cm}]
        	\node[anchor=south west,inner sep=0]  {\includegraphics[width=\tmpcolwidth,trim={20px 80px 20px 80px},clip]{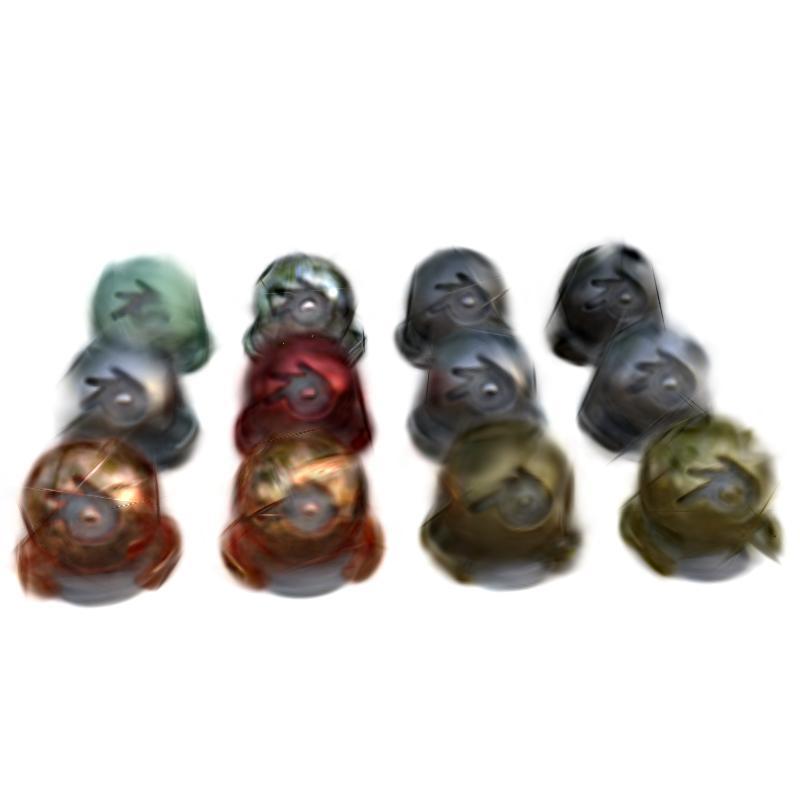}};
        	\spy[blue] on (0.95, 1.05) in node at (0.6cm,-0cm);
            \spy[green] on (1.65, 0.8) in node at (2.0cm,-0cm);
        \end{tikzpicture}\\
        &
        \begin{tikzpicture}[spy using outlines={rectangle,magnification=3,size=1.0cm}]
        	\node[anchor=south west,inner sep=0]  {\includegraphics[width=\tmpcolwidth,trim={20px 80px 20px 80px},clip]{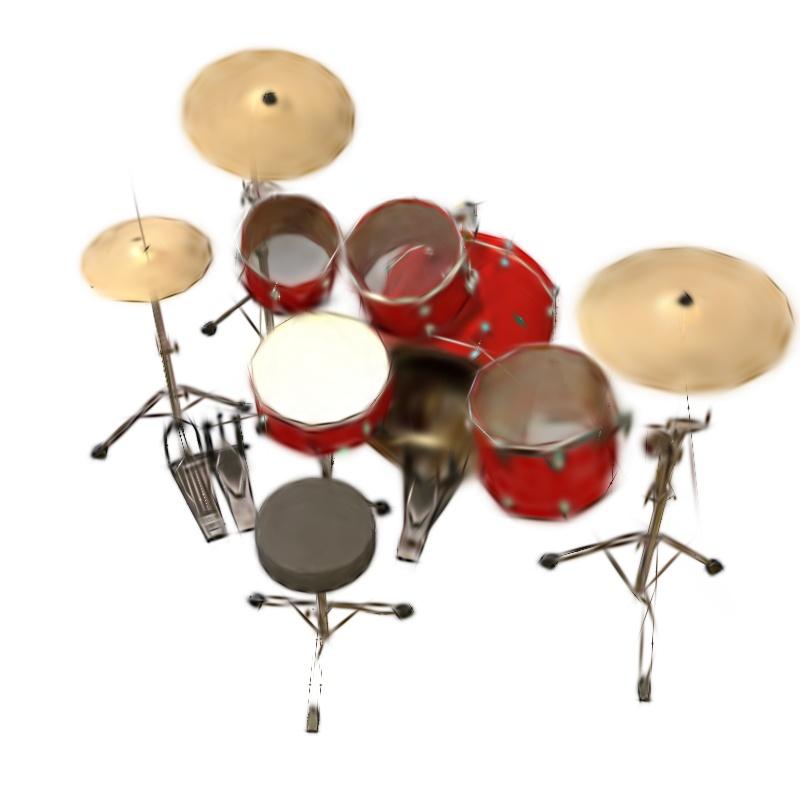}};
        	\spy[blue] on (1.2, 1.05) in node at (0.6cm,-0cm);
            \spy[green] on (2.1, 1.25) in node at (2.0cm,-0cm);
        \end{tikzpicture}
        &
        \begin{tikzpicture}[spy using outlines={rectangle,magnification=3,size=1.0cm}]
        	\node[anchor=south west,inner sep=0]  {\includegraphics[width=\tmpcolwidth,trim={20px 80px 20px 80px},clip]{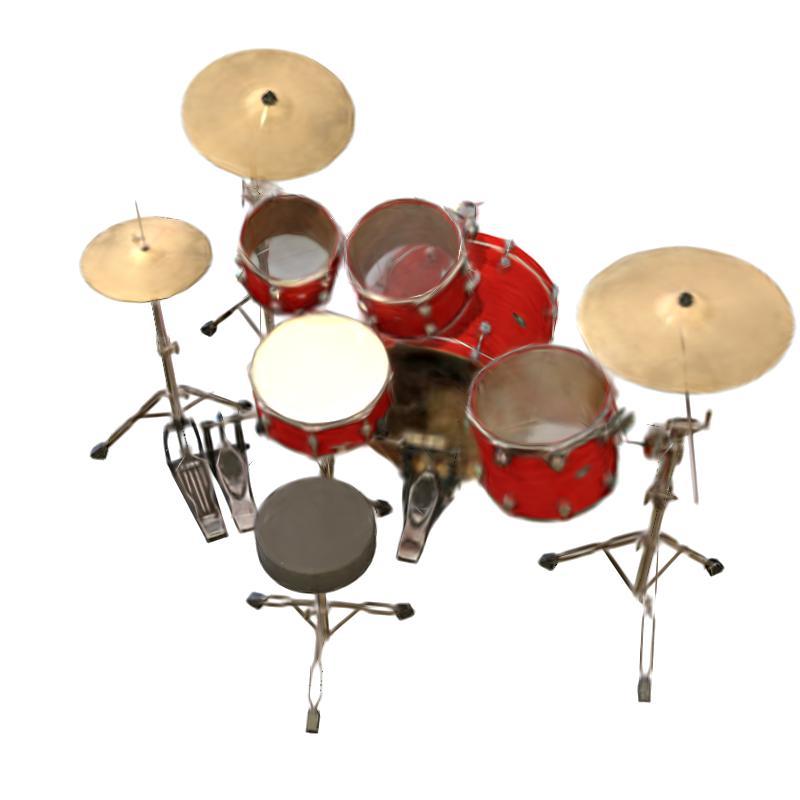}};
        	\spy[blue] on (1.2, 1.05) in node at (0.6cm,-0cm);
            \spy[green] on (2.1, 1.25) in node at (2.0cm,-0cm);
        \end{tikzpicture}
        &
        \begin{tikzpicture}[spy using outlines={rectangle,magnification=3,size=1.0cm}]
        	\node[anchor=south west,inner sep=0]  {\includegraphics[width=\tmpcolwidth,trim={20px 80px 20px 80px},clip]{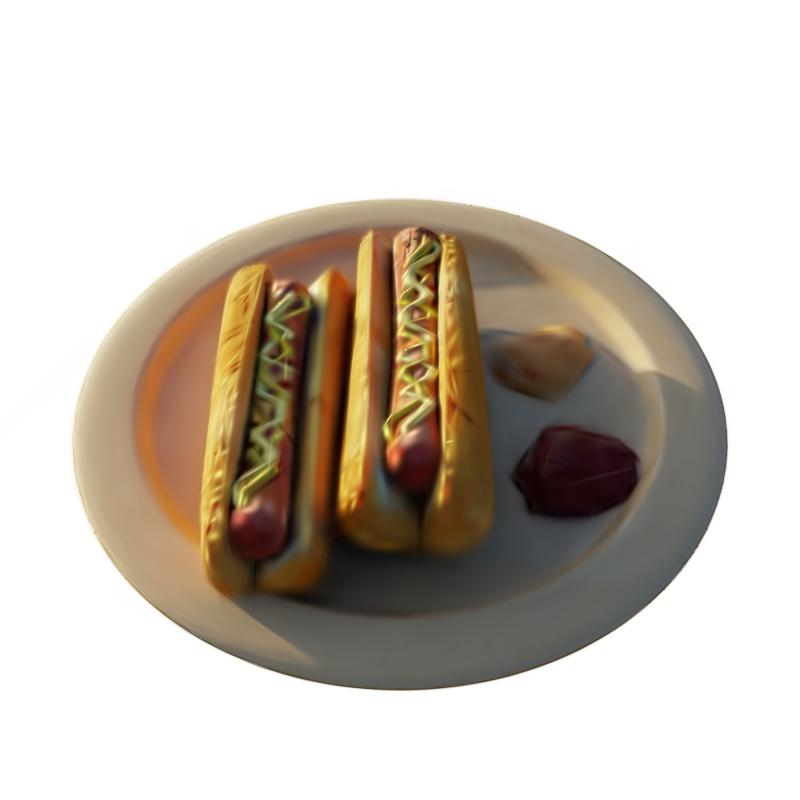}};
        	\spy[blue] on (0.95, 1.05) in node at (0.6cm,-0cm);
            \spy[green] on (1.3, 1.25) in node at (2.0cm,-0cm);
        \end{tikzpicture}
        &
        \begin{tikzpicture}[spy using outlines={rectangle,magnification=3,size=1.0cm}]
        	\node[anchor=south west,inner sep=0]  {\includegraphics[width=\tmpcolwidth,trim={20px 80px 20px 80px},clip]{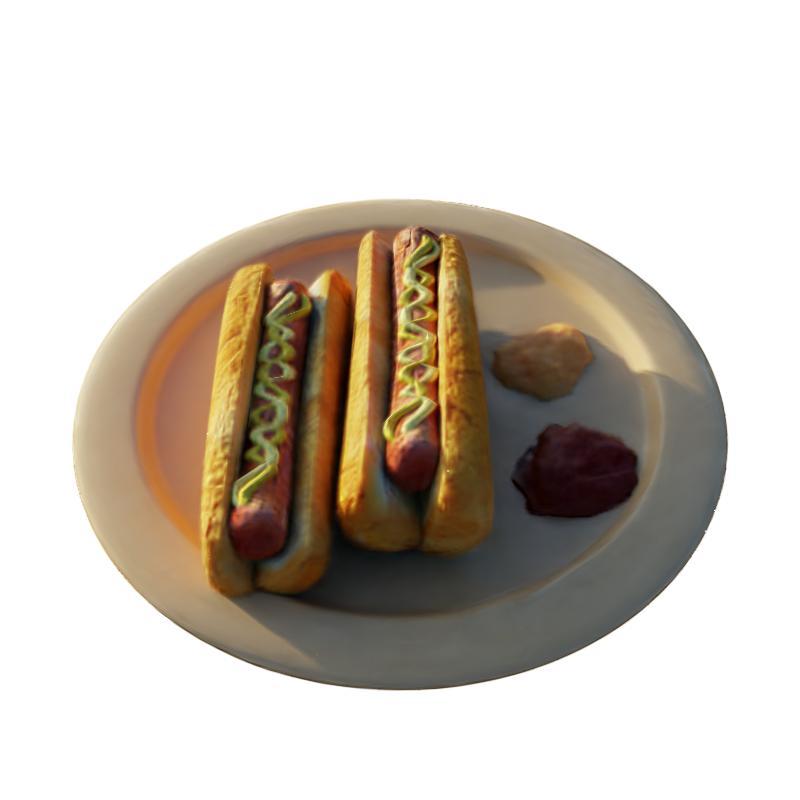}};
        	\spy[blue] on (0.95, 1.05) in node at (0.6cm,-0cm);
            \spy[green] on (1.3, 1.25) in node at (2.0cm,-0cm);
        \end{tikzpicture}
        &
        \begin{tikzpicture}[spy using outlines={rectangle,magnification=3,size=1.0cm}]
        	\node[anchor=south west,inner sep=0]  {\includegraphics[width=\tmpcolwidth,trim={20px 80px 20px 80px},clip]{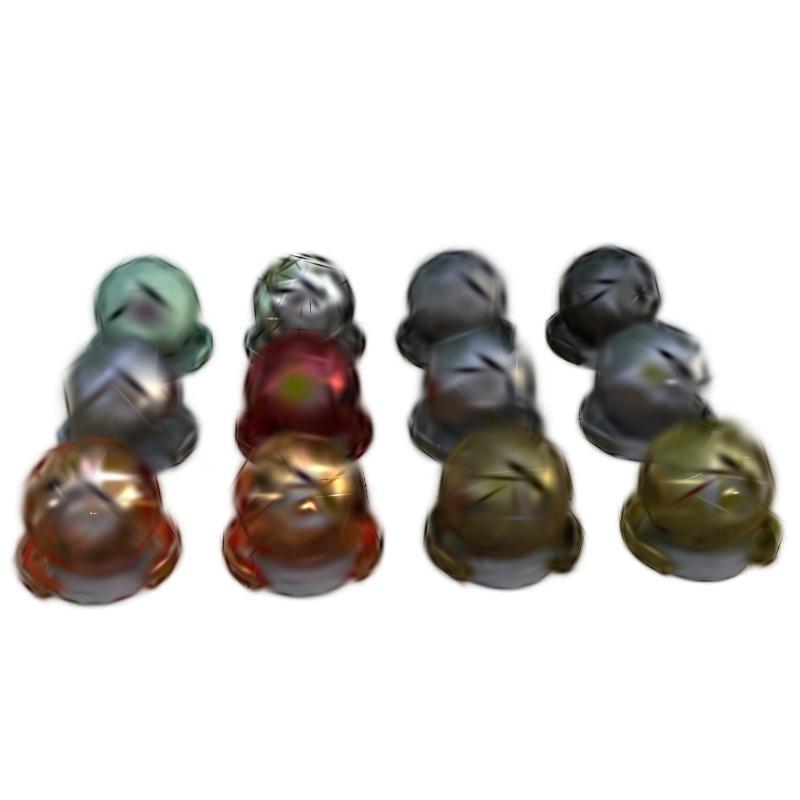}};
        	\spy[blue] on (0.95, 1.05) in node at (0.6cm,-0cm);
            \spy[green] on (1.65, 0.8) in node at (2.0cm,-0cm);
        \end{tikzpicture}
        &
        \begin{tikzpicture}[spy using outlines={rectangle,magnification=3,size=1.0cm}]
        	\node[anchor=south west,inner sep=0]  {\includegraphics[width=\tmpcolwidth,trim={20px 80px 20px 80px},clip]{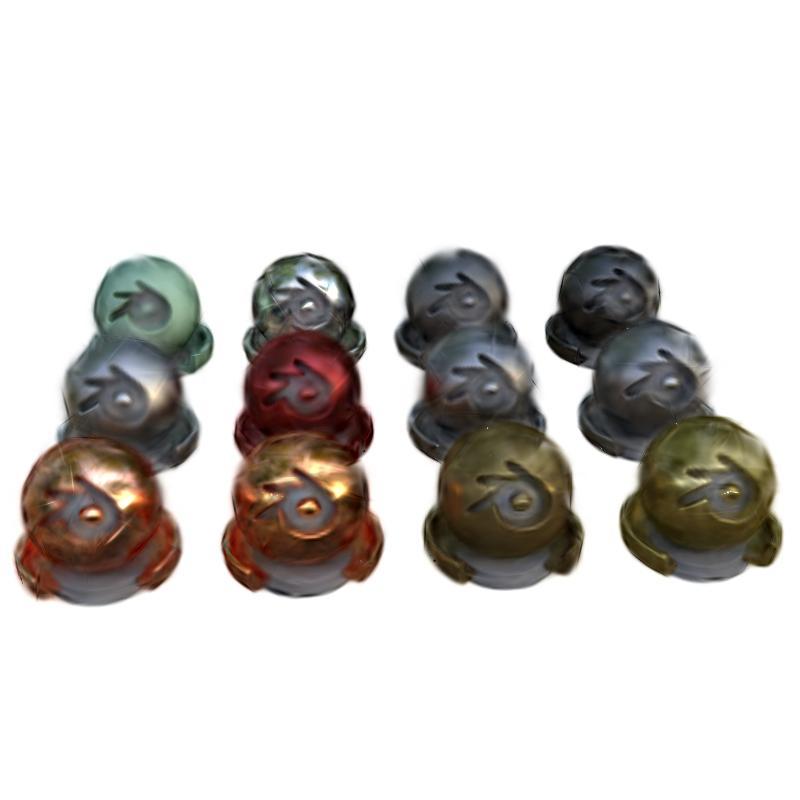}};
        	\spy[blue] on (0.95, 1.05) in node at (0.6cm,-0cm);
            \spy[green] on (1.65, 0.8) in node at (2.0cm,-0cm);
        \end{tikzpicture}
	\end{tabular}
    \setlength\tmpcolwidth{1.0in}
	\setlength{\tabcolsep}{0pt}
    \renewcommand{\arraystretch}{1}
	\centering
	\begin{tabular}{M{0.2in}M{\tmpcolwidth}M{\tmpcolwidth}M{\tmpcolwidth}M{\tmpcolwidth}M{\tmpcolwidth}M{\tmpcolwidth}}
        & \vlabel{2DGS} & \vlabel{\ours} & \vlabel{2DGS} & \vlabel{\ours} & \vlabel{2DGS} & \vlabel{\ours}\\
        \multirow{3}{*}{\rotatebox{90}{$\xleftarrow{\makebox[1.9in]{Number of Gaussians}}$}}
        &
        \begin{tikzpicture}[spy using outlines={rectangle,magnification=3,size=1.0cm}]
        	\node[anchor=south west,inner sep=0]  {\includegraphics[width=\tmpcolwidth,trim={40px 160px 40px 0px},clip]{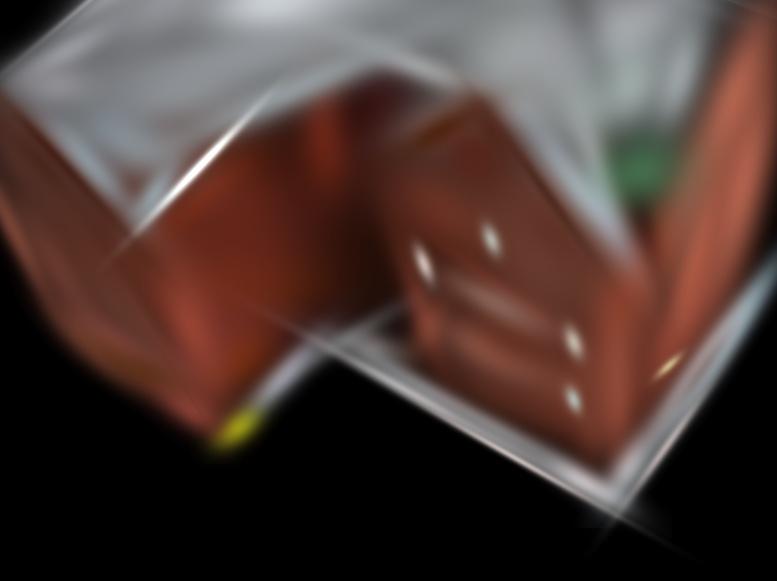}};
        	\spy[blue] on (1.5, 0.75) in node at (0.6cm,-0.4cm);
            \spy[green] on (1.9, 0.95) in node at (2.0cm,-0.4cm);
        \end{tikzpicture}
        & \begin{tikzpicture}[spy using outlines={rectangle,magnification=3,size=1.0cm}]
        	\node[anchor=south west,inner sep=0]  {\includegraphics[width=\tmpcolwidth,trim={40px 160px 40px 0px},clip]{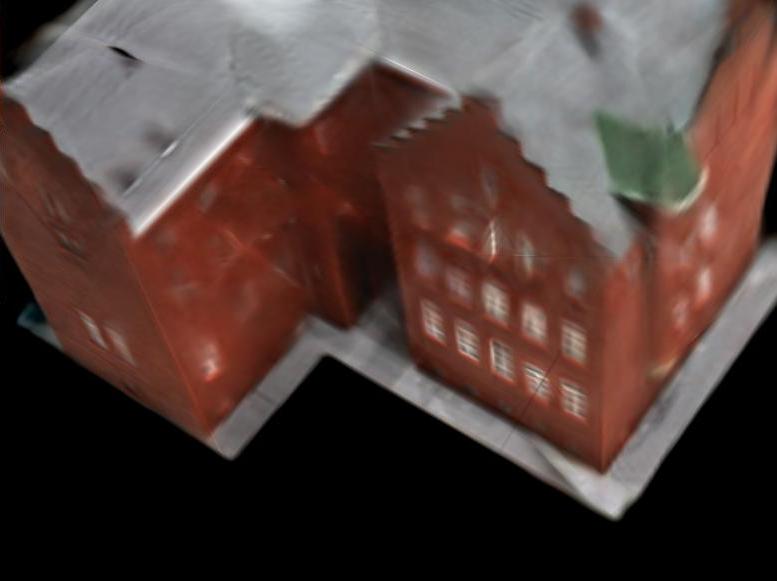}};
        	\spy[blue] on (1.5, 0.75) in node at (0.6cm,-0.4cm);
            \spy[green] on (1.9, 0.95) in node at (2.0cm,-0.4cm);
        \end{tikzpicture}
        &
        \begin{tikzpicture}[spy using outlines={rectangle,magnification=3,size=1.0cm}]
        	\node[anchor=south west,inner sep=0]  {\includegraphics[width=\tmpcolwidth,trim={40px 160px 40px 0px},clip]{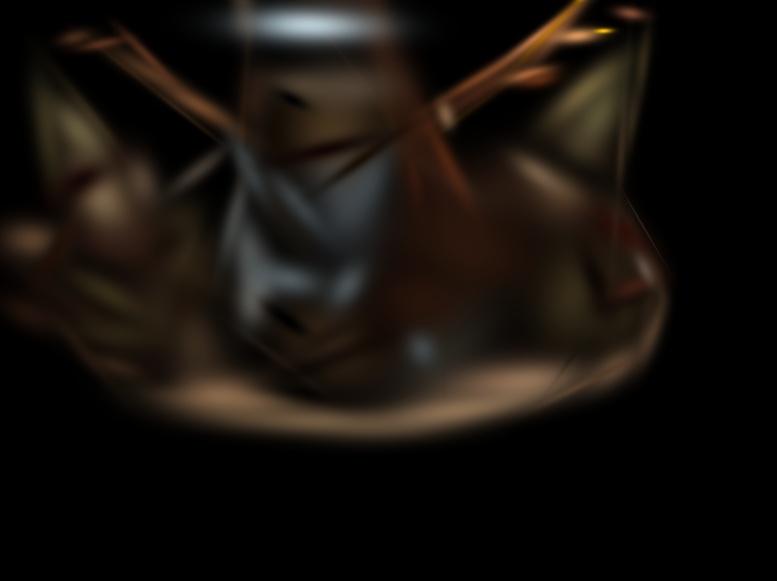}};
        	\spy[blue] on (0.25, 0.85) in node at (0.6cm,-0.4cm);
            \spy[green] on (1.05, 1.25) in node at (2.0cm,-0.4cm);
        \end{tikzpicture}
        & \begin{tikzpicture}[spy using outlines={rectangle,magnification=3,size=1.0cm}]
        	\node[anchor=south west,inner sep=0]  {\includegraphics[width=\tmpcolwidth,trim={40px 160px 40px 0px},clip]{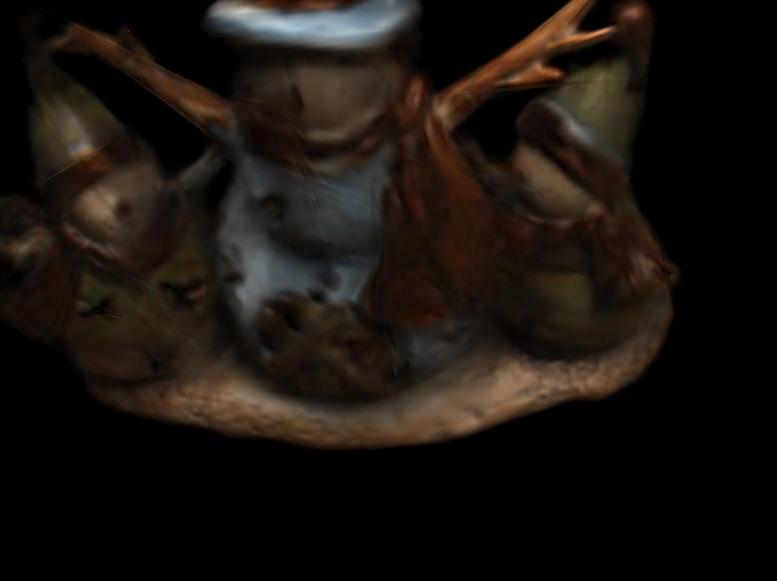}};
        	\spy[blue] on (0.25, 0.85) in node at (0.6cm,-0.4cm);
            \spy[green] on (1.05, 1.25) in node at (2.0cm,-0.4cm);
        \end{tikzpicture}
        &
        \begin{tikzpicture}[spy using outlines={rectangle,magnification=3,size=1.0cm}]
        	\node[anchor=south west,inner sep=0]  {\includegraphics[width=\tmpcolwidth,trim={40px 160px 40px 0px},clip]{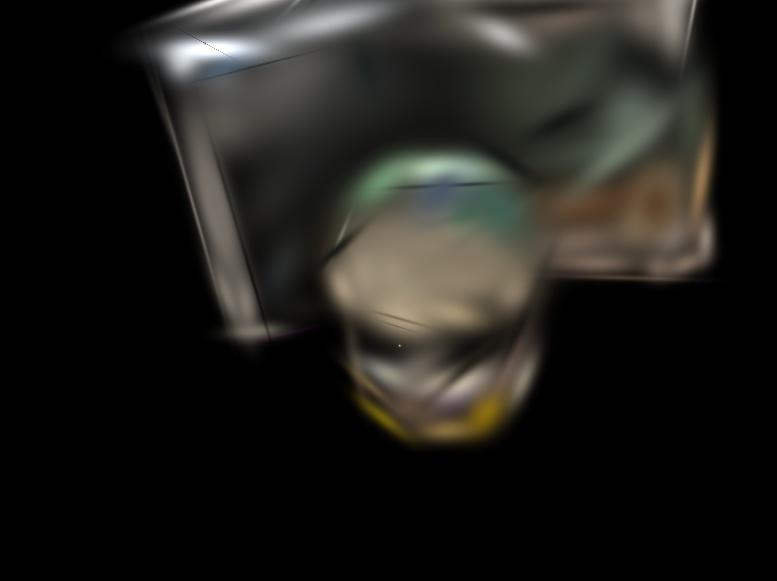}};
        	\spy[blue] on (0.8, 1.25) in node at (0.6cm,-0.4cm);
            \spy[green] on (1.4, 0.9) in node at (2.0cm,-0.4cm);
        \end{tikzpicture}
        & \begin{tikzpicture}[spy using outlines={rectangle,magnification=3,size=1.0cm}]
        	\node[anchor=south west,inner sep=0]  {\includegraphics[width=\tmpcolwidth,trim={40px 160px 40px 0px},clip]{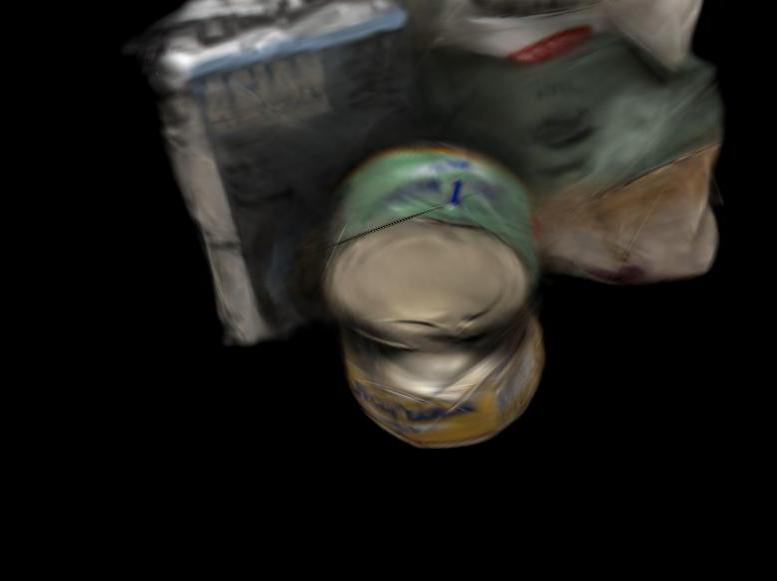}};
        	\spy[blue] on (0.8, 1.25) in node at (0.6cm,-0.4cm);
            \spy[green] on (1.4, 0.9) in node at (2.0cm,-0.4cm);
        \end{tikzpicture}\\
        &
        \begin{tikzpicture}[spy using outlines={rectangle,magnification=3,size=1.0cm}]
        	\node[anchor=south west,inner sep=0]  {\includegraphics[width=\tmpcolwidth,trim={40px 160px 40px 0px},clip]{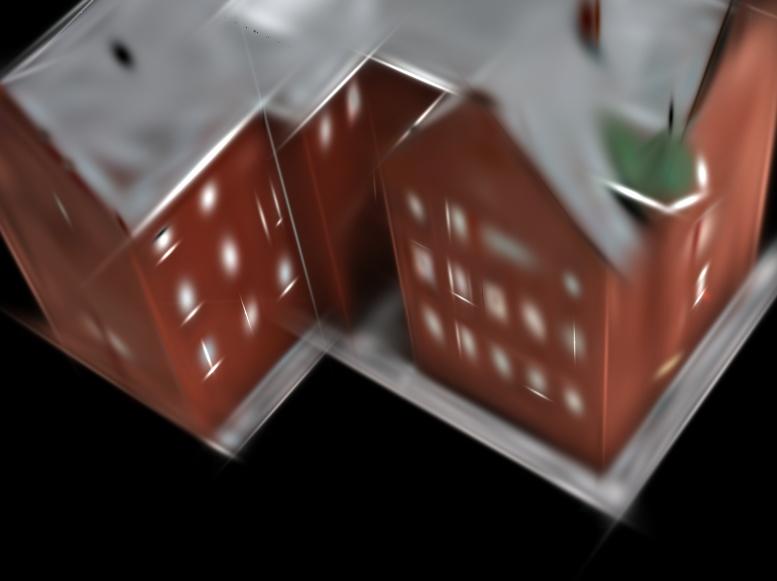}};
        	\spy[blue] on (1.5, 0.75) in node at (0.6cm,-0.4cm);
            \spy[green] on (1.9, 0.95) in node at (2.0cm,-0.4cm);
        \end{tikzpicture}
        & \begin{tikzpicture}[spy using outlines={rectangle,magnification=3,size=1.0cm}]
        	\node[anchor=south west,inner sep=0]  {\includegraphics[width=\tmpcolwidth,trim={40px 160px 40px 0px},clip]{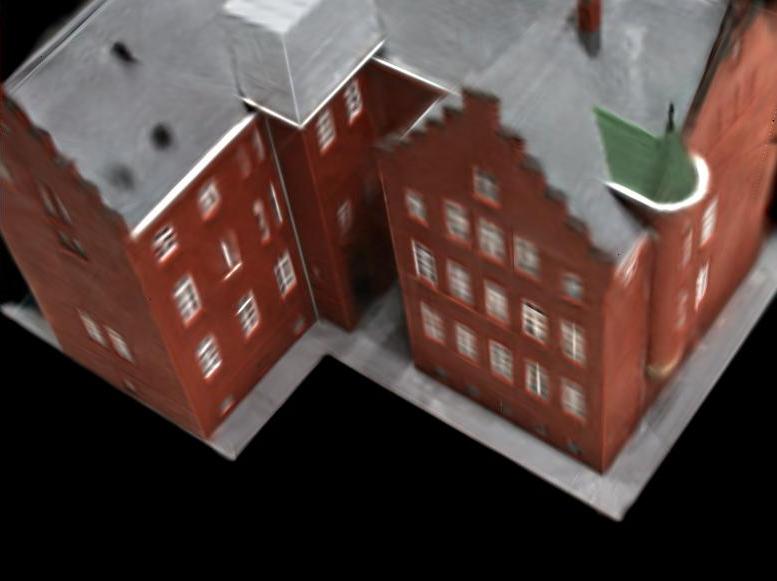}};
        	\spy[blue] on (1.5, 0.75) in node at (0.6cm,-0.4cm);
            \spy[green] on (1.9, 0.95) in node at (2.0cm,-0.4cm);
        \end{tikzpicture}
        &
        \begin{tikzpicture}[spy using outlines={rectangle,magnification=3,size=1.0cm}]
        	\node[anchor=south west,inner sep=0]  {\includegraphics[width=\tmpcolwidth,trim={40px 160px 40px 0px},clip]{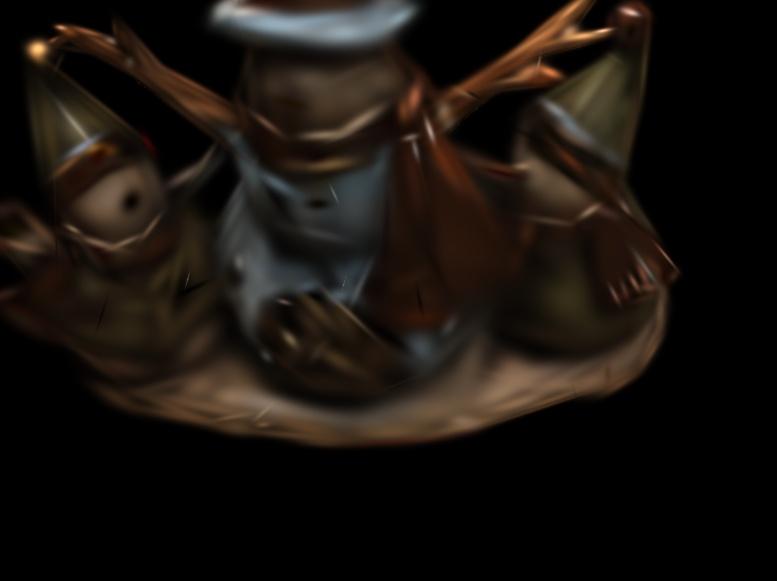}};
        	\spy[blue] on (0.25, 0.85) in node at (0.6cm,-0.4cm);
            \spy[green] on (1.05, 1.25) in node at (2.0cm,-0.4cm);
        \end{tikzpicture}
        & \begin{tikzpicture}[spy using outlines={rectangle,magnification=3,size=1.0cm}]
        	\node[anchor=south west,inner sep=0]  {\includegraphics[width=\tmpcolwidth,trim={40px 160px 40px 0px},clip]{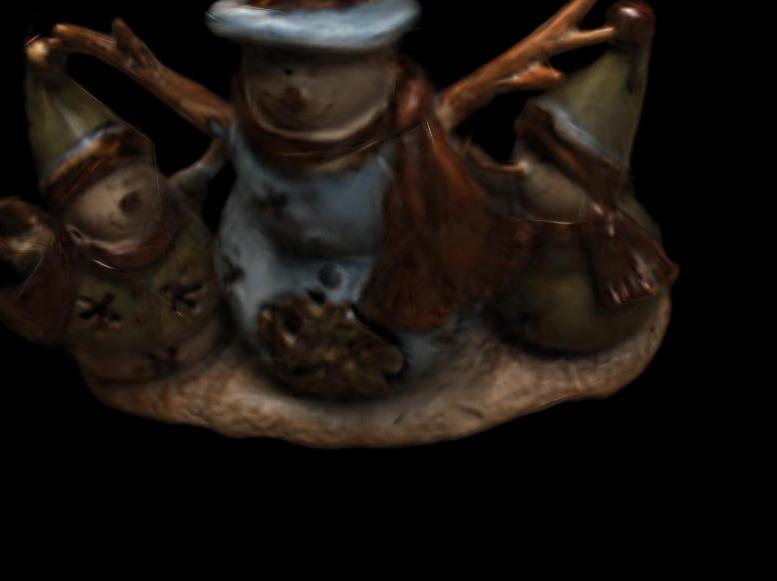}};
        	\spy[blue] on (0.25, 0.85) in node at (0.6cm,-0.4cm);
            \spy[green] on (1.05, 1.25) in node at (2.0cm,-0.4cm);
        \end{tikzpicture}
        &
        \begin{tikzpicture}[spy using outlines={rectangle,magnification=3,size=1.0cm}]
        	\node[anchor=south west,inner sep=0]  {\includegraphics[width=\tmpcolwidth,trim={40px 160px 40px 0px},clip]{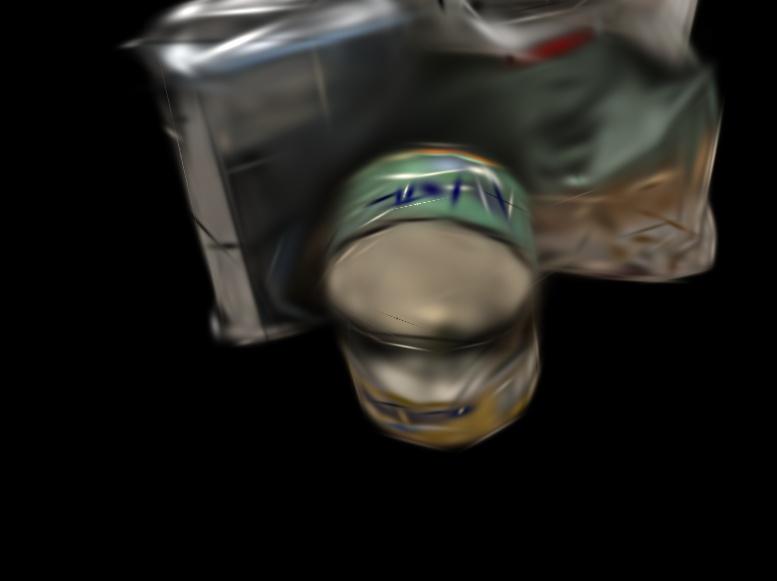}};
        	\spy[blue] on (0.8, 1.25) in node at (0.6cm,-0.4cm);
            \spy[green] on (1.4, 0.9) in node at (2.0cm,-0.4cm);
        \end{tikzpicture}
        & \begin{tikzpicture}[spy using outlines={rectangle,magnification=3,size=1.0cm}]
        	\node[anchor=south west,inner sep=0]  {\includegraphics[width=\tmpcolwidth,trim={40px 160px 40px 0px},clip]{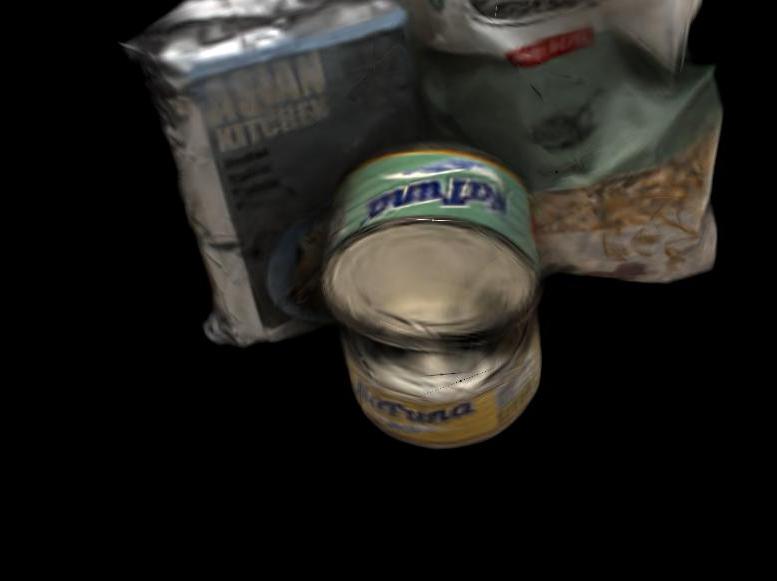}};
        	\spy[blue] on (0.8, 1.25) in node at (0.6cm,-0.4cm);
            \spy[green] on (1.4, 0.9) in node at (2.0cm,-0.4cm);
        \end{tikzpicture}\\
        &
        \begin{tikzpicture}[spy using outlines={rectangle,magnification=3,size=1.0cm}]
        	\node[anchor=south west,inner sep=0]  {\includegraphics[width=\tmpcolwidth,trim={40px 160px 40px 0px},clip]{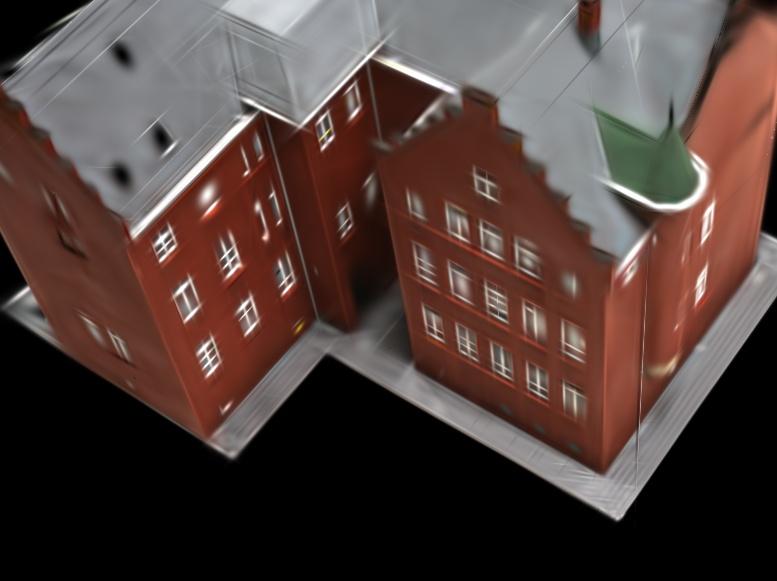}};
        	\spy[blue] on (1.5, 0.75) in node at (0.6cm,-0.4cm);
            \spy[green] on (1.9, 0.95) in node at (2.0cm,-0.4cm);
        \end{tikzpicture}
        & \begin{tikzpicture}[spy using outlines={rectangle,magnification=3,size=1.0cm}]
        	\node[anchor=south west,inner sep=0]  {\includegraphics[width=\tmpcolwidth,trim={40px 160px 40px 0px},clip]{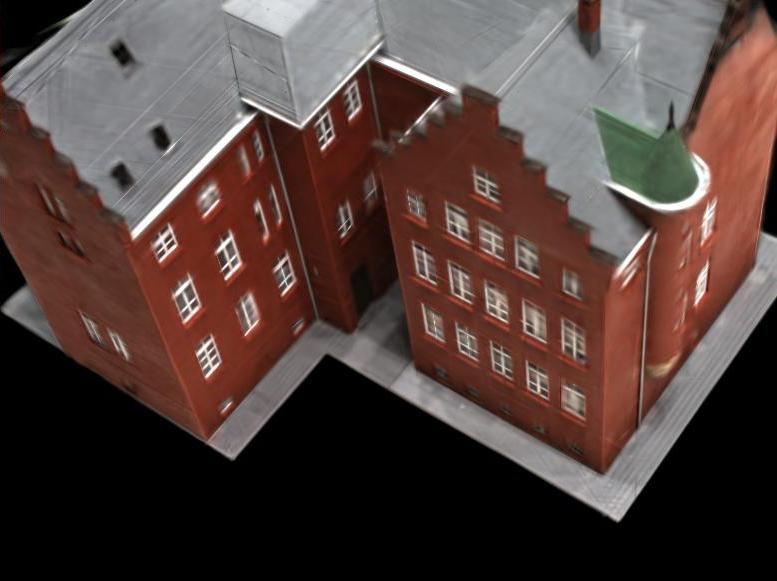}};
        	\spy[blue] on (1.5, 0.75) in node at (0.6cm,-0.4cm);
            \spy[green] on (1.9, 0.95) in node at (2.0cm,-0.4cm);
        \end{tikzpicture}
        &
        \begin{tikzpicture}[spy using outlines={rectangle,magnification=3,size=1.0cm}]
        	\node[anchor=south west,inner sep=0]  {\includegraphics[width=\tmpcolwidth,trim={40px 160px 40px 0px},clip]{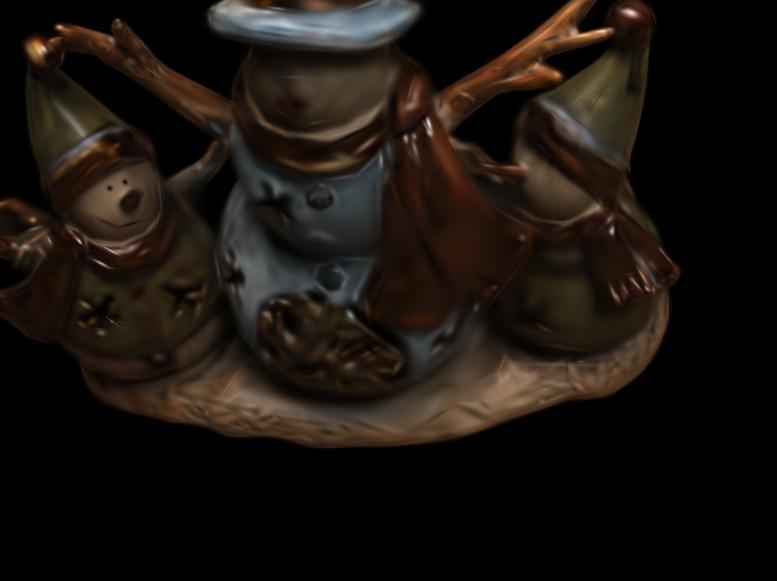}};
        	\spy[blue] on (0.25, 0.85) in node at (0.6cm,-0.4cm);
            \spy[green] on (1.05, 1.25) in node at (2.0cm,-0.4cm);
        \end{tikzpicture}
        & \begin{tikzpicture}[spy using outlines={rectangle,magnification=3,size=1.0cm}]
        	\node[anchor=south west,inner sep=0]  {\includegraphics[width=\tmpcolwidth,trim={40px 160px 40px 0px},clip]{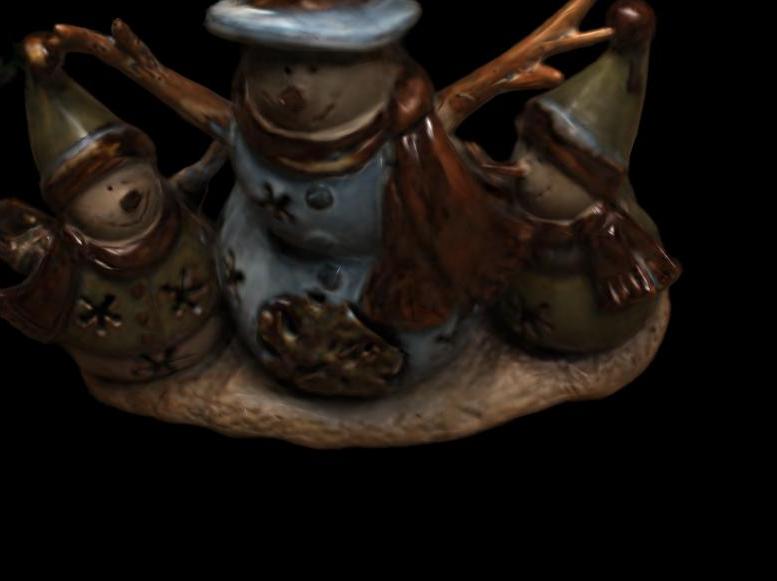}};
        	\spy[blue] on (0.25, 0.85) in node at (0.6cm,-0.4cm);
            \spy[green] on (1.05, 1.25) in node at (2.0cm,-0.4cm);
        \end{tikzpicture}
        &
        \begin{tikzpicture}[spy using outlines={rectangle,magnification=3,size=1.0cm}]
        	\node[anchor=south west,inner sep=0]  {\includegraphics[width=\tmpcolwidth,trim={40px 160px 40px 0px},clip]{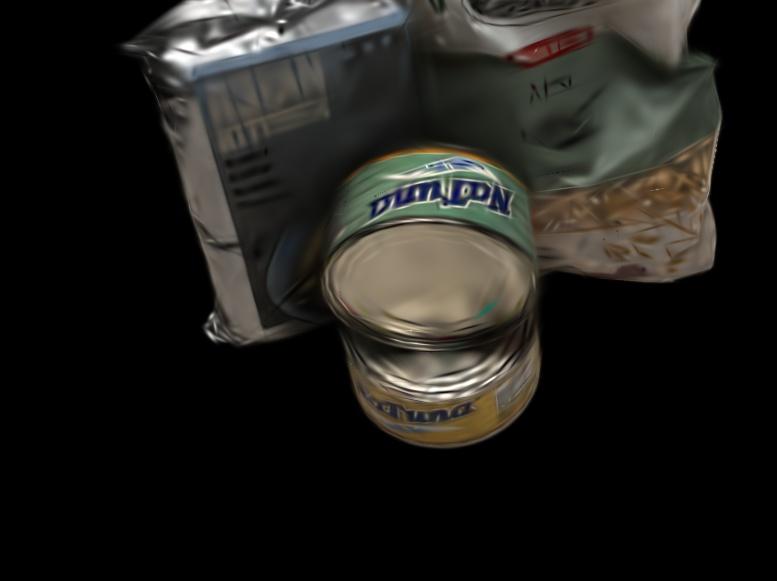}};
        	\spy[blue] on (0.8, 1.25) in node at (0.6cm,-0.4cm);
            \spy[green] on (1.4, 0.9) in node at (2.0cm,-0.4cm);
        \end{tikzpicture}
        & \begin{tikzpicture}[spy using outlines={rectangle,magnification=3,size=1.0cm}]
        	\node[anchor=south west,inner sep=0]  {\includegraphics[width=\tmpcolwidth,trim={40px 160px 40px 0px},clip]{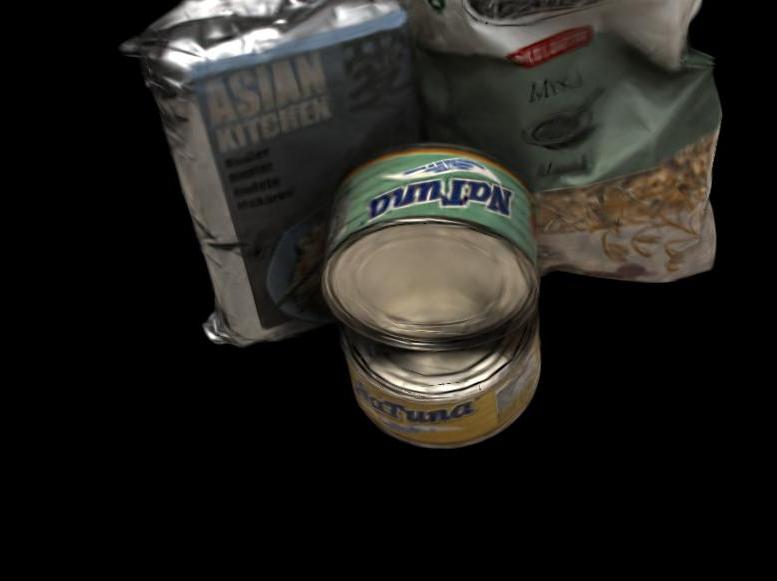}};
        	\spy[blue] on (0.8, 1.25) in node at (0.6cm,-0.4cm);
            \spy[green] on (1.4, 0.9) in node at (2.0cm,-0.4cm);
        \end{tikzpicture}
    \end{tabular}
        \vspace{-1em}
	\caption{\textbf{Novel view synthesis for discrete levels of detail.} We show test set renders of \ours and 2DGS models in three settings of levels of detail: with 128, 512, and 2048 Gaussians. We show our results on a subset of the Blender and DTU datasets.\label{fig:lod_all}}
    \vspace{-1em}
\end{figure*}

\paragraph{Results.} For this nominal setting, our approach improves over Texture-GS and performs comparably to 2DGS and 3DGS (Table~\ref{tab:nvs_all}). \editremove{In terms of performance, our}\editadd{Our} approach has a modest memory overhead compared to 2DGS resulting from the added texture parameters, and we achieve slightly faster or slower rendering framerates depending on the \editadd{scene's} number of Gaussians and optimized opacities. Qualitative comparisons between Texture-GS and \ours are shown in Figure~\ref{fig:texture_gs_comparison}. While the majority of the Texture-GS model's appearance is accurate, its surface texture parameterization constrains the model and causes blurring in parts of the scene.

\begin{figure*}[t]
	\setlength\tmpcolwidth{.32\linewidth}
	\setlength{\tabcolsep}{0pt}
	\centering
	\begin{tabular}{ccc}
		\multicolumn{3}{c}{$\xrightarrow{\makebox[5in]{Number of Texels}}$}\\
        \begin{tikzpicture}[spy using outlines={rectangle,magnification=2,size=1.75cm}]   
        	\node[anchor=south west,inner sep=0]  {\includegraphics[width=\tmpcolwidth,trim={1cm 5.5cm 1cm 3cm},clip]{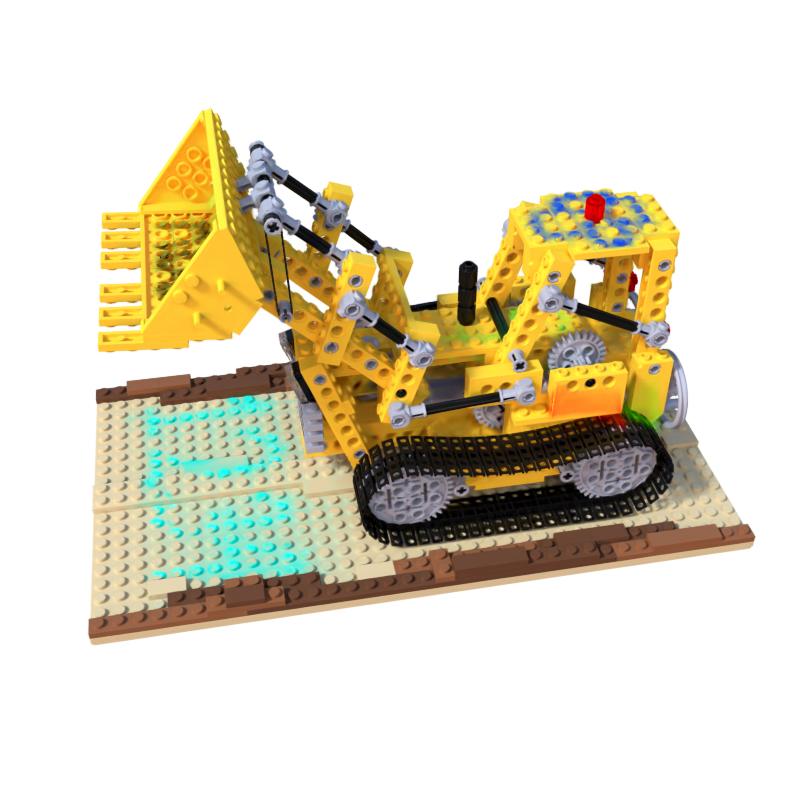}};
        	\spy[blue] on (4.4, 1.85) in node at (1cm,1cm);
        \end{tikzpicture} &
        \begin{tikzpicture}[spy using outlines={rectangle,magnification=2,size=1.75cm}]   
        	\node[anchor=south west,inner sep=0]  {\includegraphics[width=\tmpcolwidth,trim={1cm 5.5cm 1cm 3cm},clip]{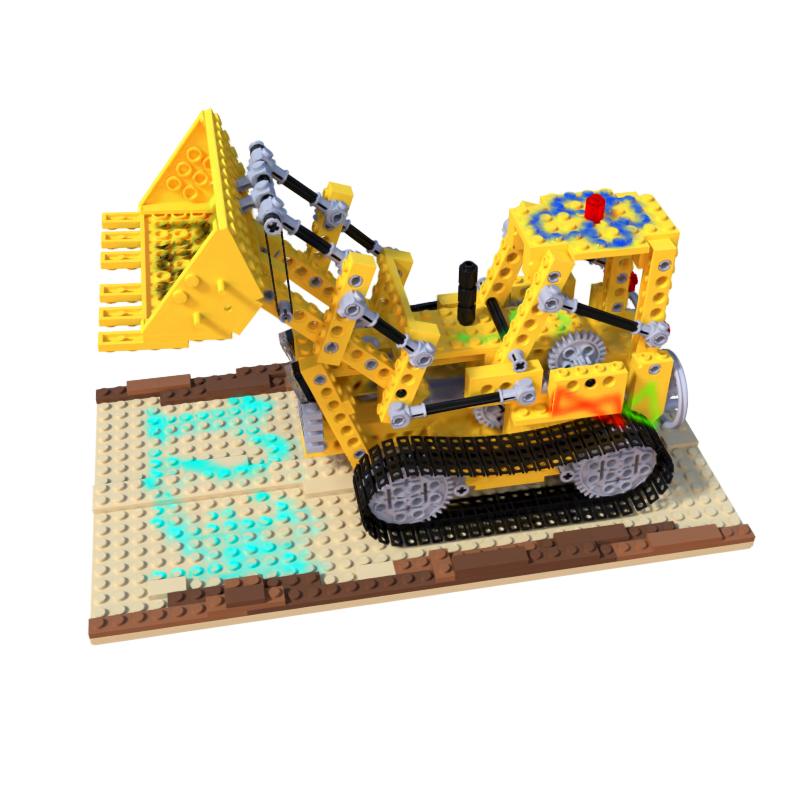}};
        	\spy[blue] on (4.4, 1.85) in node at (1cm,1cm);
        \end{tikzpicture} &
        \begin{tikzpicture}[spy using outlines={rectangle,magnification=2,size=1.75cm}]   
        	\node[anchor=south west,inner sep=0]  {\includegraphics[width=\tmpcolwidth,trim={1cm 5.5cm 1cm 3cm},clip]{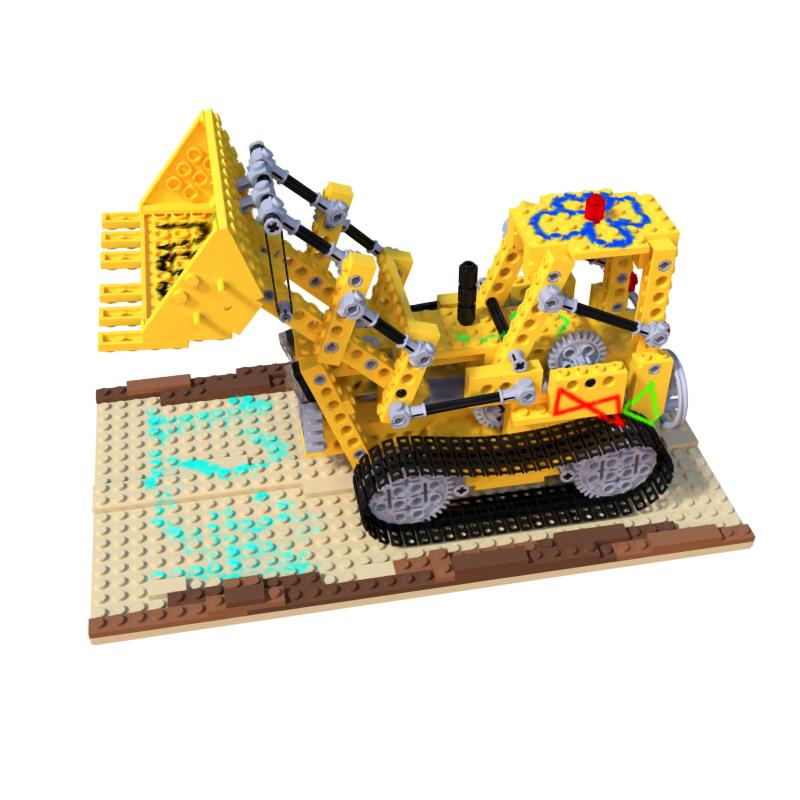}};
        	\spy[blue] on (4.4, 1.85) in node at (1cm,1cm);
        \end{tikzpicture}
	\end{tabular}
        \vspace{-1em}
	\caption{\textbf{Texture painting.} We use texture painting to draw on the texels of a Lego \ours model from multiple angles. We examine the editing performance as the number of texels is varied. From left to right, we use zero (equivalent to 2DGS), $10^6$, and $10^7$ texels, respectively. The number of texels affects the quality of the edit, particularly in regions with a low density of Gaussians, as highlighted.\label{fig:edit}}
 \vspace{-1em}
\end{figure*}

\begin{figure}[h]
	\setlength\tmpcolwidth{.24\linewidth}
	\setlength{\tabcolsep}{0pt}
	\centering
	\begin{tabular}{cccc}
        \includegraphics[width=\tmpcolwidth,trim={4cm 0cm 4cm 0cm},clip]{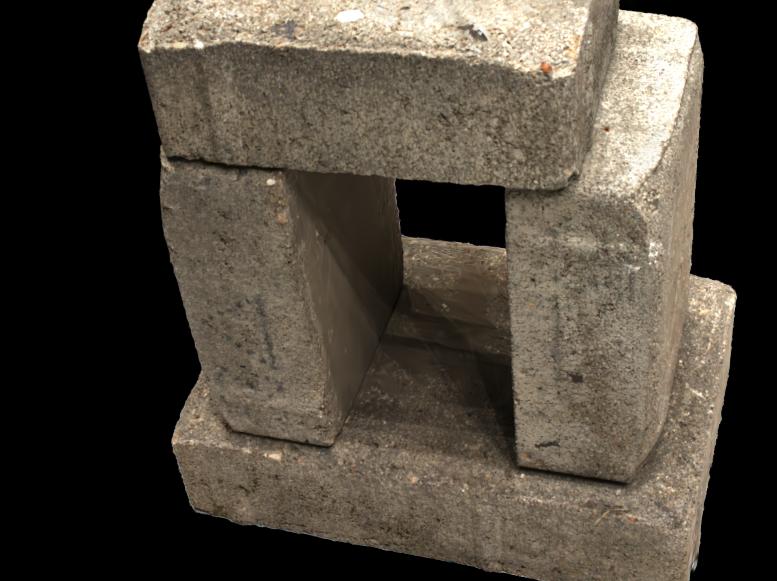} & \includegraphics[width=\tmpcolwidth,trim={4cm 0cm 4cm 0cm},clip]{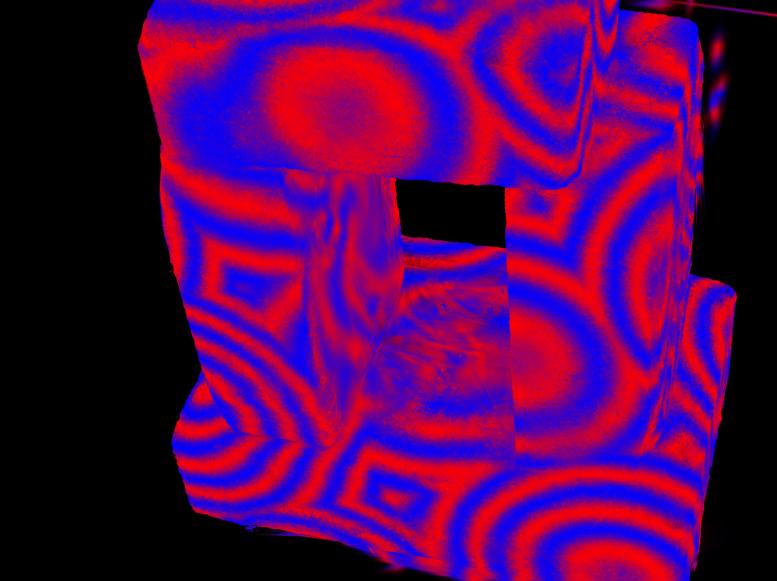} &
        \includegraphics[width=\tmpcolwidth,trim={4cm 0cm 4cm 0cm},clip]{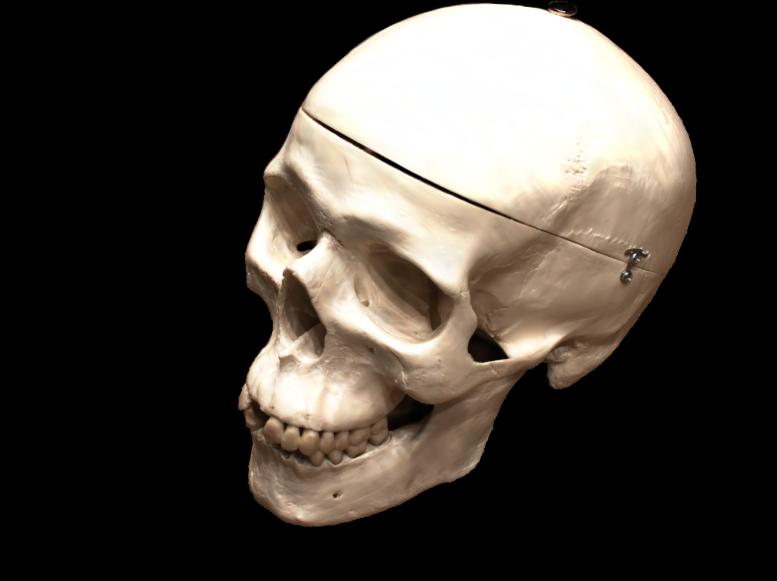} &
        \includegraphics[width=\tmpcolwidth,trim={4cm 0cm 4cm 0cm},clip]{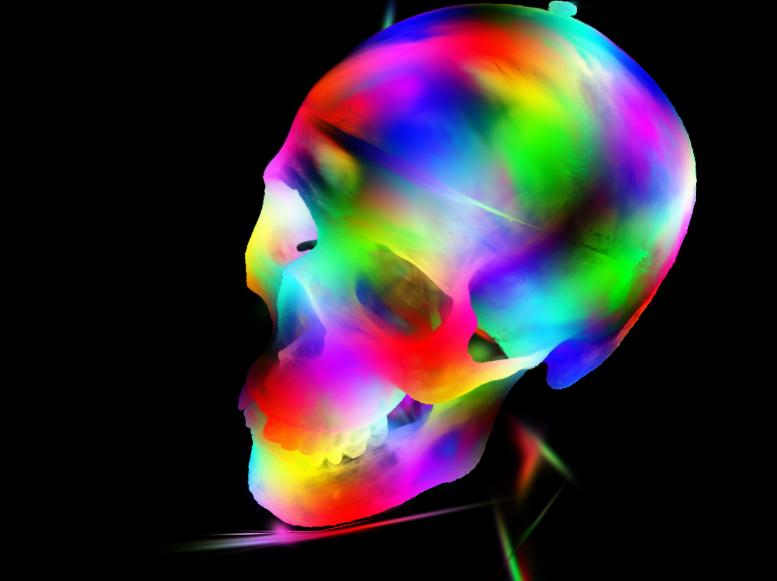}\\
        \includegraphics[width=\tmpcolwidth,trim={4cm 0cm 4cm 0cm},clip]{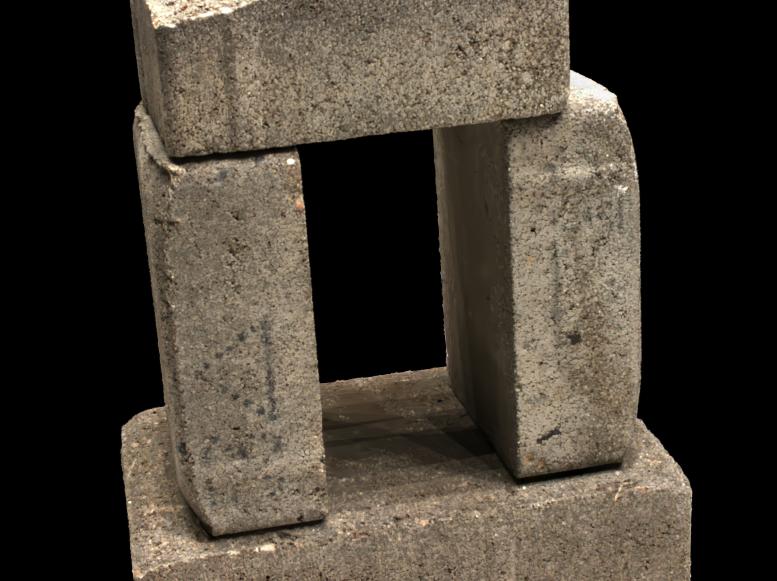} & \includegraphics[width=\tmpcolwidth,trim={4cm 0cm 4cm 0cm},clip]{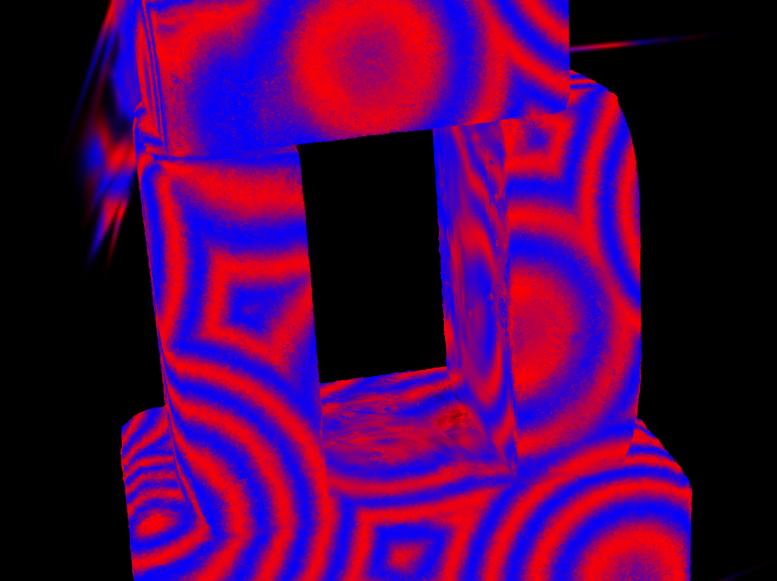} &
        \includegraphics[width=\tmpcolwidth,trim={0cm 0cm 8cm 0cm},clip]{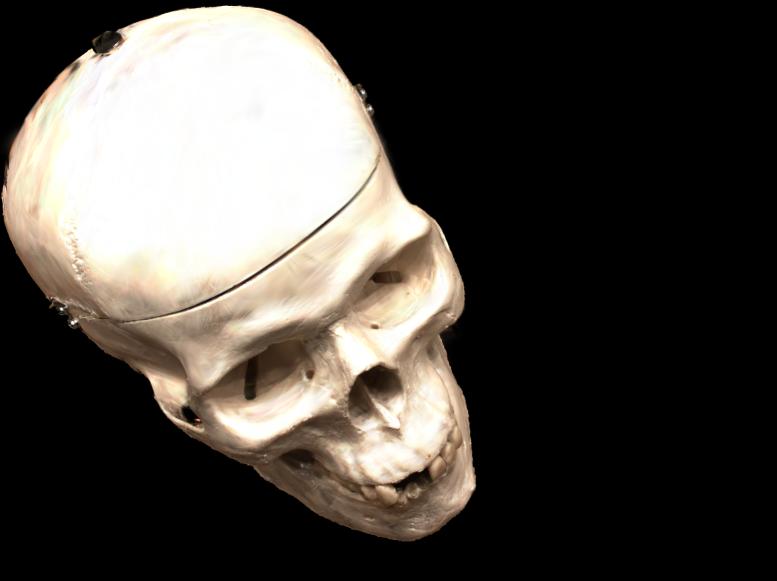} &
        \includegraphics[width=\tmpcolwidth,trim={0cm 0cm 8cm 0cm},clip]{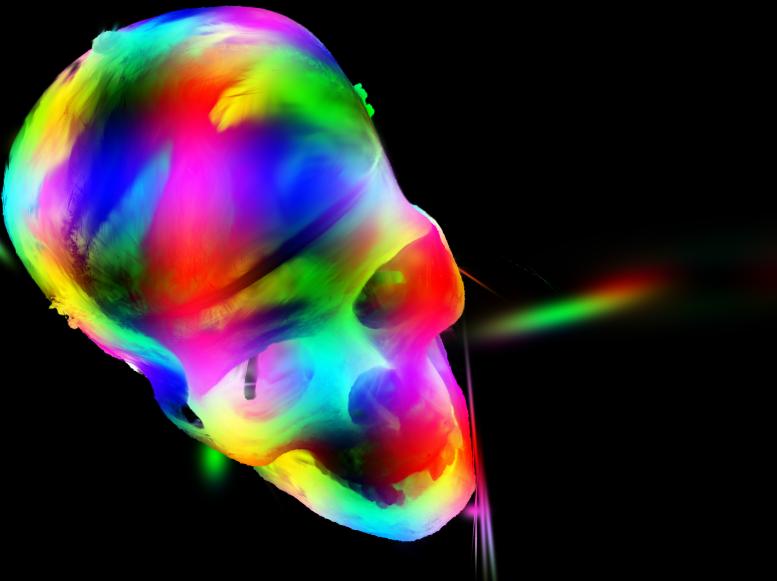}
	\end{tabular}
        \vspace{-1em}
	\caption{\textbf{Procedural texturing.} We demonstrate retexturing of DTU scan40 and scan65 \ours models using procedural textures. The original renders and retextured renders are shown from two different viewpoints.\label{fig:retexture}}
\end{figure}

\subsection{Geometric Level of Detail} 
We demonstrate \editremove{the performance of our method}\editadd{our method's performance} for discrete levels of geometric detail~\cite{luebke2002level} by varying the number of Gaussians used in the representation. 
For each Blender and DTU scene, we train five different models each with $128, 512, 2048, 8192,$ and $32768$ Gaussians for \ours and 2DGS. 
In order to control the number of Gaussians, we turn off densification and culling during training. 
As such, the number of Gaussians is based on the initial point cloud model, which is the same for both methods. We obtain the initial point cloud by sampling Gaussian means from a trained 2DGS model and exporting their positions and colors corresponding to the zeroth-order spherical harmonics coefficient. These are used to initialize the Gaussian means and colors for both 2DGS and \ours, while the other parameters are initialized according to 2DGS's default initialization.
\editremove{Both the \ours and 2DGS}\editadd{All} models are trained for 7,000 iterations\editremove{, which was found to be sufficient for convergence}. 
The number of texels for the \ours models is fixed at $10^6$.

In Figure~\ref{fig:lod_plot}, we show that the visual fidelity of the proposed method improves over 2DGS across a range of Gaussian primitives and scenes for both datasets. This trend is supported by Figure~\ref{fig:lod_all}, in which we compare renders of novel views from \ours and 2DGS. As the number of Gaussians decreases, the 2DGS results become increasingly blurry whereas our approach retains more\editremove{ high-frequency} details.

\subsection{Appearance Editing}
We demonstrate how the appearance of \ours models can be edited so that details smaller than individual Gaussians can be represented, following the methods presented in Section~\ref{sec:editing}.  When applying this editing framework to a 2DGS model (see Figure~\ref{fig:teaser}), details are lost if the density of Gaussians is too low to model the contours of the edit. On the other hand, \ours captures details independent of the number and layout of the Gaussians. 

In Figure~\ref{fig:edit}, we show the result of texture painting applied to a model from a Blender synthetic scene. As the number of texels increases, the visual fidelity improves. 
The same trend is observed when varying the number of texels and Gaussians for the \textit{whiteboard} scene of Figure~\ref{fig:teaser}. 
We quantitatively assess this effect in terms of PSNR and LPIPS in Figure~\ref{fig:sweep_plot} and show that using texels results in significant improvements relative to 2DGS.
Finally, we show that \ours supports procedural re-texturing in Figure~\ref{fig:retexture}.

\section{Conclusion}
Our work improves upon 2D Gaussian splatting by adding texture maps to each Gaussian primitive, resulting in a representation where appearance and geometry are encoded separately. 
Our work could help to improve representations of large-scale 3D scenes by enabling fewer Gaussian primitives to be used without sacrificing rendering fidelity. 
It could also enable integration of Gaussian splatting with conventional graphics techniques, such as bump mapping or level-of-detail rendering, which can take advantage of our explicit representations of texture and geometry. 

We envision multiple avenues for future work. 
Anti-aliasing and mip-mapping~\cite{akenine2019real} are important elements of texturing which we have not yet addressed. \editadd{For unbounded scenes, our method needs a large number of texels, as discussed further in the supplementary. Overcoming this limitation is important to bring \ours to more general scenes.} Our method could be made more memory efficient by optimizing in a latent space~\cite{thies2019deferred} or using a sparse representation~\cite{peyre2009sparse}. 
Finally, it may be possible to support rendering methods that adjust the texture resolution based on the camera frustum~\cite{crow1984summed,lindstrom1996real}.

\paragraph{Acknowledgements} DBL and KNK acknowledge support from the Natural Sciences and Engineering Research Council of Canada (NSERC) under the RGPIN, RTI, and Alliance programs. DBL \editremove{also acknowledges support from}\editadd{was also supported by} the Canada Foundation for Innovation and the Ontario Research Fund. VR was supported by the Wolfond Scholarship Program.

{\small
\bibliographystyle{ieee_fullname}
\bibliography{main}
}

\clearpage
\renewcommand\thesection{S\arabic{section}}
\renewcommand\thefigure{S\arabic{figure}}
\renewcommand\thetable{S\arabic{table}}

\setcounter{section}{0}
\setcounter{figure}{0}
\setcounter{table}{0}

\section{Video Results}
Video results can be viewed on our project page website: \url{https://lessvrong.com/cs/gstex}. We include novel view synthesis renderings for all key figures of our paper. The videos are best viewed on the Chrome browser.

\section{Implementation Details}

\paragraph{Texture implementation.} Let $T_j = \sum_{i=1}^j U_iV_i$ be the number of texels of the first $j$ Gaussians. We represent our texture as a $T_n\times 3$ tensor, where the $i$ Gaussian's texture map can be found flattened from indices $T_{i-1}$ to $T_i - 1$. During rendering via the CUDA kernel, along with the typical Gaussian parameters passed in 2DGS, the texture along with arrays $T_i, U_i,$ and $V_i$ are passed in as well. When a ray intersects Gaussian $i$ in the CUDA kernel, knowledge of $T_{i-1}$ is needed to locate the subarray corresponding to the $i$th Gaussian's texture map, and $U_i$ and $V_i$ are necessary to simulate reshaping into a 2D grid and querying into the correct indices. Note that the $i$th Gaussian's texture map are not loaded into shared memory like with the other Gaussian parameters. The individual texture maps are simply too large. We do, however, load $T_{i-1}, U_i, V_i$ into shared memory.

\paragraph{Texture painting.}
We elaborate on our texture painting method, which casts an edited image onto the textured Gaussian splats from a given viewpoint. For each texel of the \ours model, we aggregate the RGBA values of the rays which hit it so as to propose an updated texel color. To accommodate the transparency of Gaussians, we weight these aggregated values by the transmittance of each ray when it hits the texel, as well as the coefficient from bilinear interpolation. In actuality, some pixels of the edited image may be unedited, or have low alpha from filtering of a stroke. As such, the final texel colors should be interpolated between the original texel colors and the aggregated edited texel color. More precisely, for texel $\tau$ that is intersected with rays of RGBA values $(\hat{\radiance}_i, \alpha_i)$ with combined (between interpolation and transmittance) weight $\omega_i$,  we compute
\begin{align*}
\tau_{\text{edit}} &= \sum \hat{\radiance}_i \alpha_i \omega_i,\\
w_{0} &= \sum \alpha_i \omega_i,\\
w_{1} &= \sum (1-\alpha_i) \omega_i.
\end{align*}
Then the updated texel color $\tau_{\text{new}}$ interpolates between the edited texel value $\tau_{\text{edit}}$ and the original texel value $\tau_{\text{orig}}$ with weight $w_0 / (w_0 + w_1)$: 
\begin{equation*}
    \tau_{\text{new}} = \frac{w_{0} \tau_{\text{edit}} + w_{1} \tau_{\text{orig}}}{w_{0} + w_{1}}.
\end{equation*}

When updating the texture values, we ensure that only texels within $\num{1e-2}$ (in normalized depth coordinates) of the median depth from the corresponding view are altered. After this editing process, we can visualize the edited scene from any viewpoint in a 3D-consistent fashion.

\begin{table}[t]
  \begin{center}
     \resizebox{\columnwidth}{!}{%

\begin{tabular}{lcccccccccc}
\toprule
& \multicolumn{3}{c}{Blender} & \multicolumn{3}{c}{DTU}\\
Method & PSNR & SSIM & LPIPS & PSNR & SSIM & LPIPS\\ %
\cmidrule(l{2pt}r{2pt}){1-1}\cmidrule(l{2pt}r{2pt}){2-4}\cmidrule(l{2pt}r{2pt}){5-7}
\ours & 33.25 & 0.969 & 0.024 & 32.87 & 0.956 & 0.038\\
\cmidrule(l{2pt}r{2pt}){1-1}\cmidrule(l{2pt}r{2pt}){2-4}\cmidrule(l{2pt}r{2pt}){5-7}
w/o stop gradient & 33.07 & 0.968 & 0.024 & 32.82 & 0.956 & 0.039\\
reset never & 33.23 & 0.969 & 0.024 & 32.94 & 0.956 & 0.037\\
reset every & 33.01 & 0.967 & 0.026 & 32.88 & 0.955 & 0.040\\
\bottomrule
\end{tabular}
}
\end{center}
\caption{Ablation experiments for novel view synthesis and rendering performance on the synthetic Blender dataset and DTU dataset. The first row, \ours, refers to our method exactly as described in the paper. The next three rows give various metrics upon adjusting certain design choices. We report PSNR $\uparrow$, SSIM $\uparrow$, LPIPS $\downarrow$. \label{tab:ablations}}
\end{table}

\section{Additional Experiments}

\begin{table*}[t]
  \centering
        \setlength\tabcolsep{8.0pt}
    \begin{tabular}{lccccccccc}
    \toprule
    \multirow{2}{*}{Method} & \multicolumn{3}{c}{Outdoor} & \multicolumn{4}{c}{Indoor} &\\
    \cmidrule(l{4pt}r{4pt}){2-4}\cmidrule(l{4pt}r{4pt}){5-8}
    & Bicycle & Garden & Stump & Bonsai & Counter & Kitchen & Room & Mean\\
    \midrule
    3DGS & 25.18 & 27.23 & 26.55 & 32.13 & 28.99 & 31.31 & 31.36 & 28.96\\
    2DGS & 24.79 & 26.74 & 26.19 & 31.32 & 28.08 & 30.40 & 30.57 & 28.30\\
    \cmidrule(l{2pt}r{2pt}){2-8}\cmidrule(l{2pt}r{2pt}){9-9}
    \ours ($10^7$) & 24.84 & 27.05 & 26.30 & 31.90 & 28.62 & 31.00 & 31.25 & 28.71\\
    \ours ($10^8$) & 24.91 & 27.15 & 26.36 & 32.05 & 28.68 & 31.12 & 31.37 & 28.81\\
    \midrule
    3DGS & 0.763 & 0.862 & 0.771 & 0.940 & 0.906 & 0.925 & 0.917 & 0.869\\
    2DGS & 0.742 & 0.850 & 0.759 & 0.933 & 0.895 & 0.919 & 0.911 & 0.858\\
    \cmidrule(l{2pt}r{2pt}){2-8}\cmidrule(l{2pt}r{2pt}){9-9}
    \ours ($10^7$) & 0.747 & 0.857 & 0.763 & 0.938 & 0.902 & 0.924 & 0.916 & 0.864\\
    \ours ($10^8$) & 0.752 & 0.861 & 0.766 & 0.939 & 0.905 & 0.926 & 0.919 & 0.867\\
    \midrule
    3DGS & 0.212 & 0.109 & 0.216 & 0.205 & 0.202 & 0.127 & 0.221 & 0.185\\
    2DGS & 0.214 & 0.098 & 0.192 & 0.147 & 0.183 & 0.111 & 0.188 & 0.162\\
    \cmidrule(l{2pt}r{2pt}){2-8}\cmidrule(l{2pt}r{2pt}){9-9}
    \ours ($10^7$) & 0.201 & 0.088 & 0.176 & 0.136 & 0.169 & 0.102 & 0.177 & 0.150\\
    \ours ($10^8$) & 0.192 & 0.083 & 0.172 & 0.125 & 0.158 & 0.097 & 0.167 & 0.142\\
    \midrule
    3DGS & 6000K & 5700K & 4900K & 1300K & 1200K & 1800K & 1500K & 3200K\\
    2DGS & 5300K & 3200K & 3500K & 810K & 660K & 850K & 890K & 2200K\\
    \cmidrule(l{2pt}r{2pt}){2-8}\cmidrule(l{2pt}r{2pt}){9-9}
    \ours ($10^7$) & 5300K & 3200K & 3500K & 810K & 660K & 850K & 890K & 2200K\\
    \ours ($10^8$) & 5300K & 3200K & 3500K & 810K & 660K & 850K & 890K & 2200K\\
    \bottomrule
    \end{tabular}
    \caption{\editadd{Novel view synthesis metrics for individual scenes on the MipNeRF-360 dataset. We report PSNR $\uparrow$, SSIM $\uparrow$, LPIPS $\downarrow$, and number of Gaussians, respectively.}}
    \label{tab:nvs_360}
\end{table*}

\subsection{Evaluation Over Design Choices.}
We show the results of three of our design choices in Table~\ref{tab:ablations}. In the first, we check the stop on the gradient between texture colors and Gaussian geometries. Allowing the gradient (``w/o stop gradient") decreases the visual quality slightly. In the other two rows, we evaluate the effect of resetting the texel size more or less often. In ``reset never", we do not change the texel size at all after initialization. In ``reset every", we change it after every initialization. In the actual method, \ours, we reset every 100 iterations. There is not a strong trend in either direction, suggesting that the choice of reset frequency is not significant.

\subsection{Large-Scale Scenes}
\editadd{Though our focus is on object-centric scenes, our method still functions for large-scale scenes such as the MipNeRF-360 dataset~\cite{barron2021mip}. We evaluate our method in Table~\ref{tab:nvs_360}. We use the same experimental set-up as with Table 1 of the main paper except that we use $10^7$ and $10^8$ texels rather than $10^6$, as the number of Gaussians in the scenes are much greater. During the initial 2DGS training, normal regularization was enabled with a coefficient of $0.05$ while it was disabled during \ours training. No distortion regularization was used in either stage. Similarly to the Blender and DTU scenes, we observe minor improvements in visual metrics compared to 2DGS. Even with $10^8$ texels, the texels may be fairly large compared to the details in training and test images, as shown in Figures~\ref{fig:m360_outdoor} and~\ref{fig:m360_indoor}. In unbounded scenes especially, there is large variation in the range between camera views and individual Gaussians. As a result, the texels of foreground Gaussians cover a disproportionate area in the renders while background texels are undersampled and exhibit aliasing artifacts.}

\section{Experiment Details}
We give additional details for the experiments presented in the paper.

\subsection{Captured Scene}
To validate the usefulness of \ours on real data, we capture our own scene, \textit{whiteboard}. We choose a standing whiteboard with four visually apparent pieces of paper taped onto one side, and red marker drawings on another. We captured a move-around video with a Galaxy S21 mobile phone. During capture, we kept ISO and shutter speed fixed to minimize photometric inconsistency. From the video, we selected frames that represent wide range of viewpoints while minimizing motion blur, and we utilize SwinIR \cite{liang2021swinir} to remove compression artifacts. Segmentation masks are generated using Segment Anything \cite{kirillov2023segany} with manual point prompts, and the output masks are further manually painted to remove visible holes and islands. After running COLMAP on our data, 126 images remained, which we undistorted and split into a train and test sets following the intervals of 8 practice typically used for real-world captures \cite{barron2021mip}.

\subsection{Novel View Synthesis}
Full results to the novel view synthesis are listed in Tables~\ref{tab:supp_nvs_syn} and~\ref{tab:supp_nvs_dtu}.
We use the official implementations of 3DGS \cite{kerbl20233d}, 2DGS \cite{huang20242d}, and Texture-GS \cite{xu2024texture}. We use the default hyperparameters for all. For 2DGS, their codebase had updated since the paper release and had tweaked the densification condition. More specifically, the version of the codebase which we worked with intentionally zeroed out gradients for Gaussians when they were sufficiently small to use their anti-aliasing blur. This noticeably reduced the number of Gaussians that were densified at a small cost to photometric quality, which the authors recommend.

For our evaluation of Texture-GS, we use their default training pipeline which has three stages composed of 30,000, 20,000, and 40,000 iterations, totalling to 90,000 iterations. Though this is $3\times$ that of \ours as well as 2DGS and 3DGS, we find that the visual quality of Texture-GS is noticeably behind others.

To calculate all metrics, we first quantize render RGBs to unsigned 8-bit integers before converting back to the float range $[0, 1]$ and applying the metric calculations. For LPIPS, we use version 0.1 with AlexNet activations.

\subsection{Geometric Level of Detail}
We give additional qualitative results for the geometric level of detail experiment. In Figures~\ref{fig:lod_supp_blender} and~\ref{fig:lod_supp_dtu}, we compare 2DGS and \ours in discrete level of detail for the remaining scenes not shown in Figure 6 of the main paper.

\subsection{Appearance Editing}

\begin{figure}
    \includegraphics[width=3.3in]{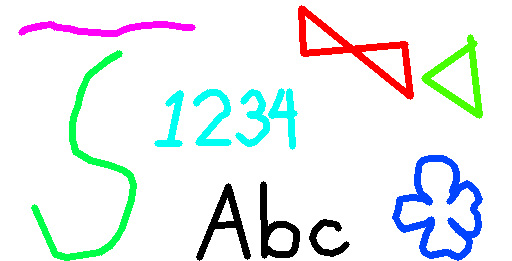}
    \caption{We drew various shapes and cast them from different views onto the Lego model. \label{fig:canvas}}
\end{figure}

\paragraph{Texture painting.} In Figure~\ref{fig:canvas}, we include the 2D images drawn on the Lego models in Figure 7 of the main paper.

\paragraph{Procedural textures.} We evaluate the colors resulting from the procedural textures at the centers of each texel in world coordinates. In our implementation, we do this within the PyTorch framework. In Figure 8 of the main paper, we demonstrate the procedural texture
\begin{alignat*}{2}
    d(x,y,z) &= ||(x, y, z) - (\lfloor x\rceil, \lfloor y\rceil, \lfloor z\rceil) ||_2\\
    R_1(x, y, z) &= 0.5 (\sin(d) + 1),\\
    G_2(x, y, z) &= 0,\\
    B_2(x, y, z) &= 0.5\left(1 - \sin(d)\right),
\end{alignat*}
where $\lfloor t \rceil$ is $t$ rounded to the nearest integer, creating circle-like patterns of blue and red. We also have the simple 
\begin{alignat*}{2}
    R_2(x, y, z) &= 0.5\left(\sin(x) + 1\right),\\
    G_2(x, y, z) &= 0.5\left(\sin(y) + 1\right),\\
    B_2(x, y, z) &= 0.5\left(\sin(z) + 1\right),
\end{alignat*}inducing axis-aligned stripes of color.

\begin{figure*}
    \newcommand{\vlabel}[1]{\vspace{3pt}#1\vspace{1pt}}
    \setlength\tmpcolwidth{3.2in}
	\setlength{\tabcolsep}{1pt}
    \renewcommand{\arraystretch}{0.5}
	\centering
	\begin{tabular}{M{0.0in}M{\tmpcolwidth}M{\tmpcolwidth}}
        & \vlabel{Render} & \vlabel{Texels}\\
        &
        \includegraphics[width=\tmpcolwidth,trim={97px 62px 60px 20px},clip]{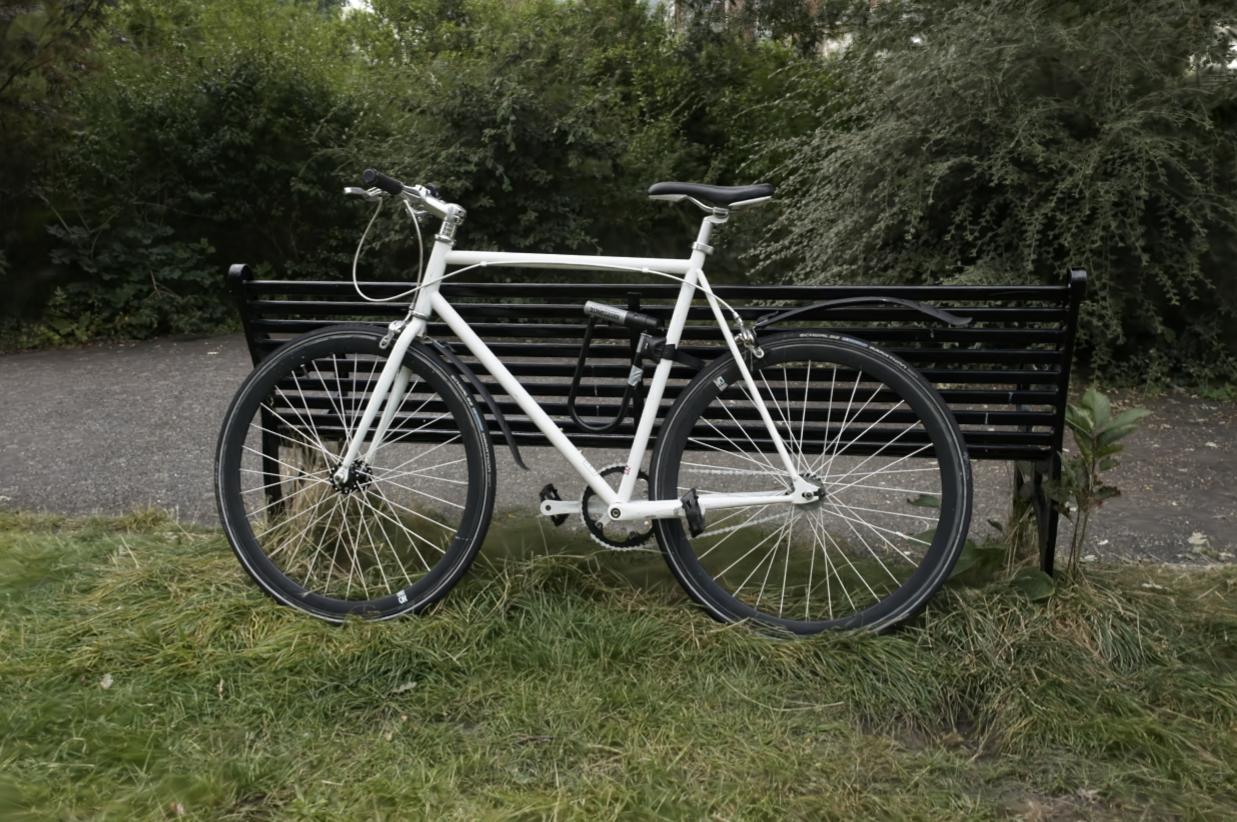}
        &
        \includegraphics[width=\tmpcolwidth,trim={97px 62px 60px 20px},clip]{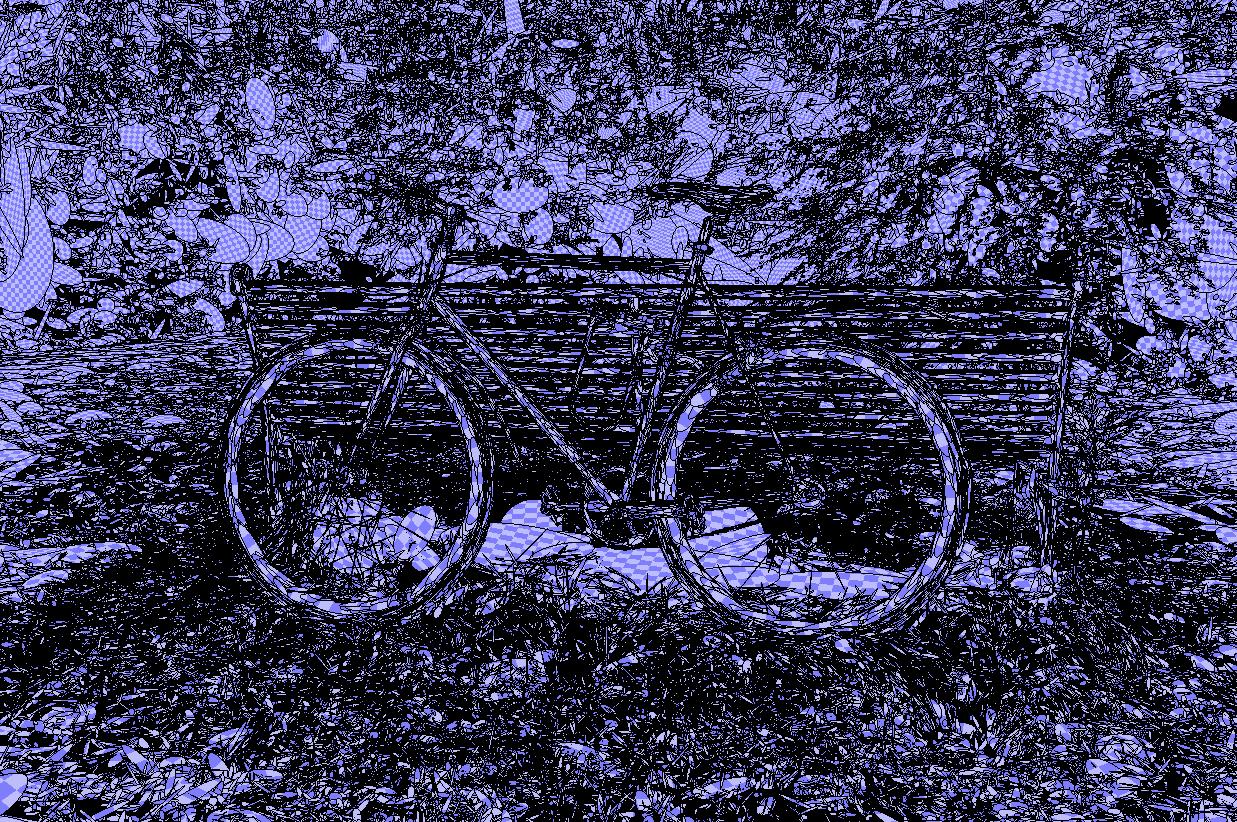}\\
        &
        \includegraphics[width=\tmpcolwidth,trim={156px 20px 60px 80px},clip]{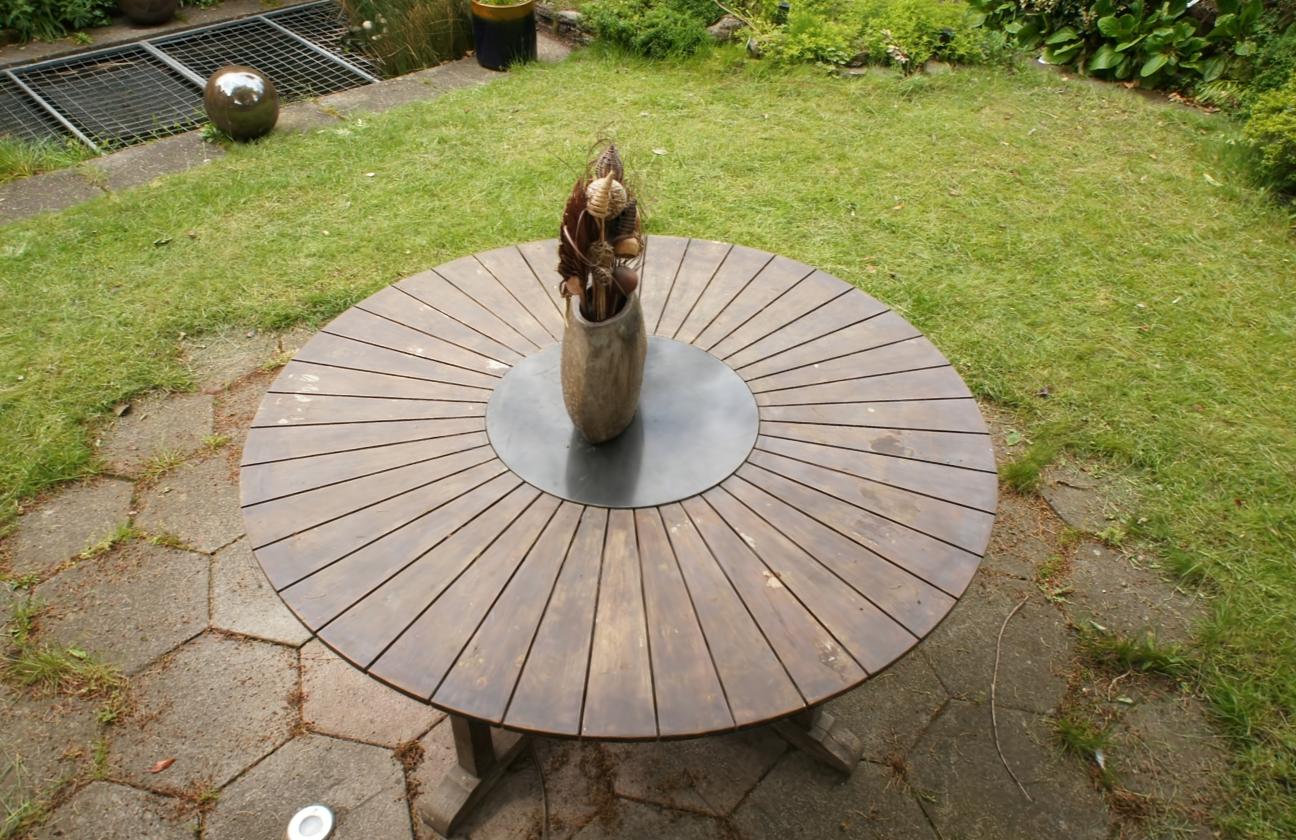}
        &
        \includegraphics[width=\tmpcolwidth,trim={156px 20px 60px 80px},clip]{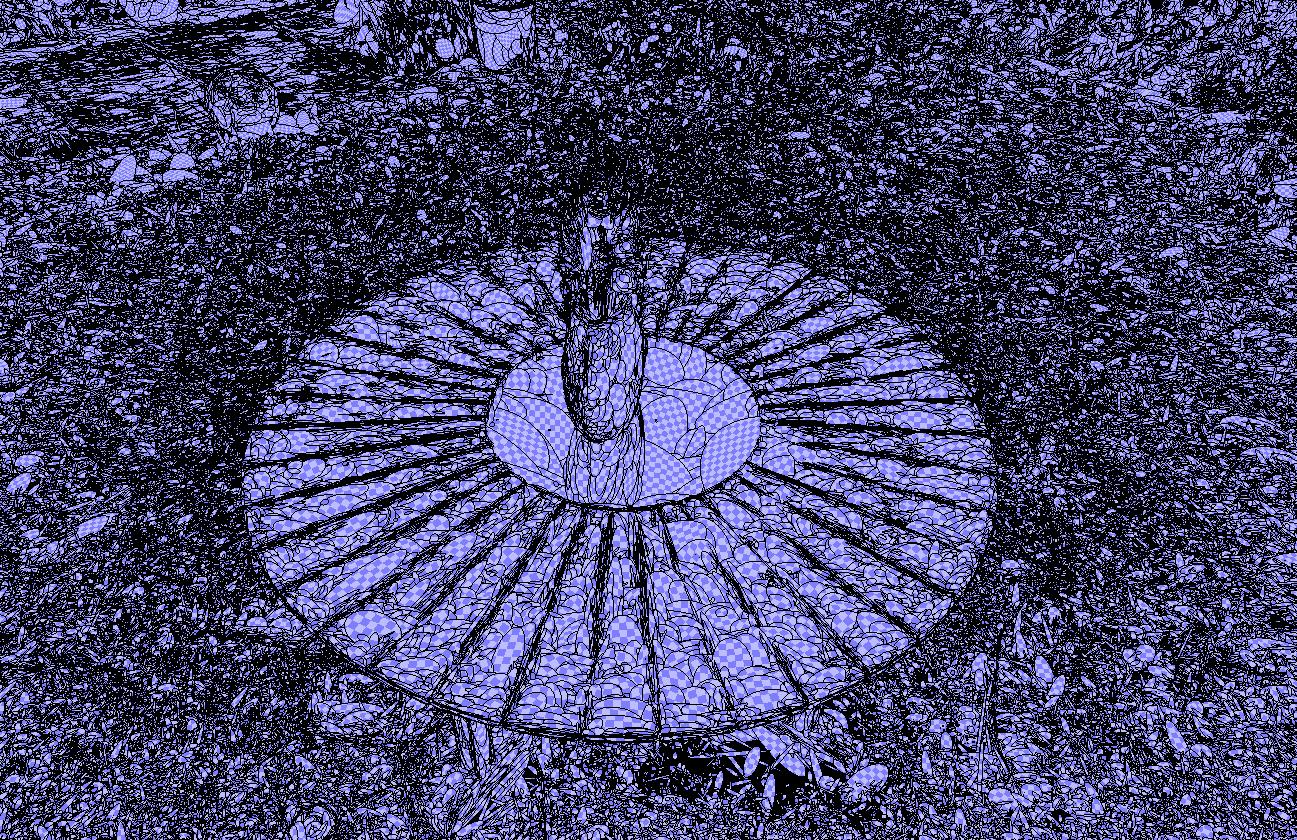}\\
        &
        \includegraphics[width=\tmpcolwidth,trim={104px 65px 60px 20px},clip]{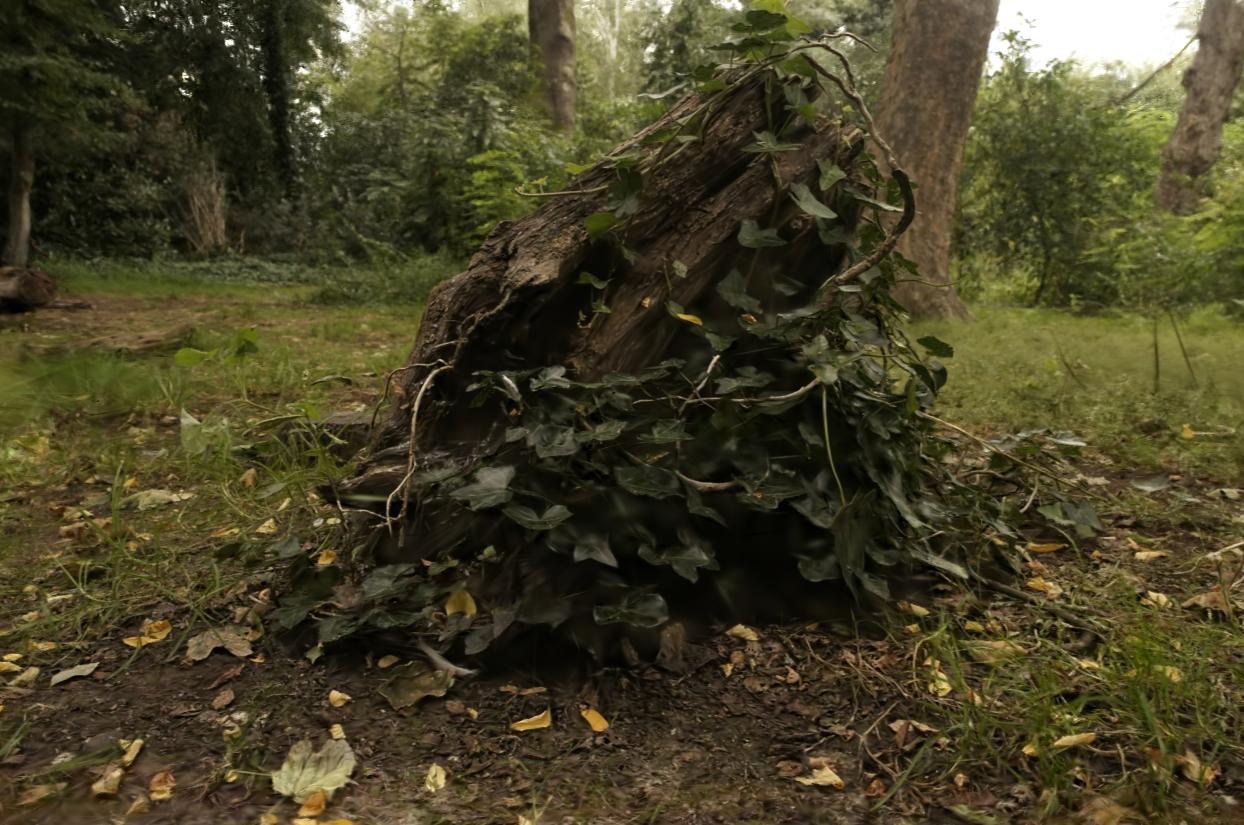}
        &
        \includegraphics[width=\tmpcolwidth,trim={104px 65px 60px 20px},clip]{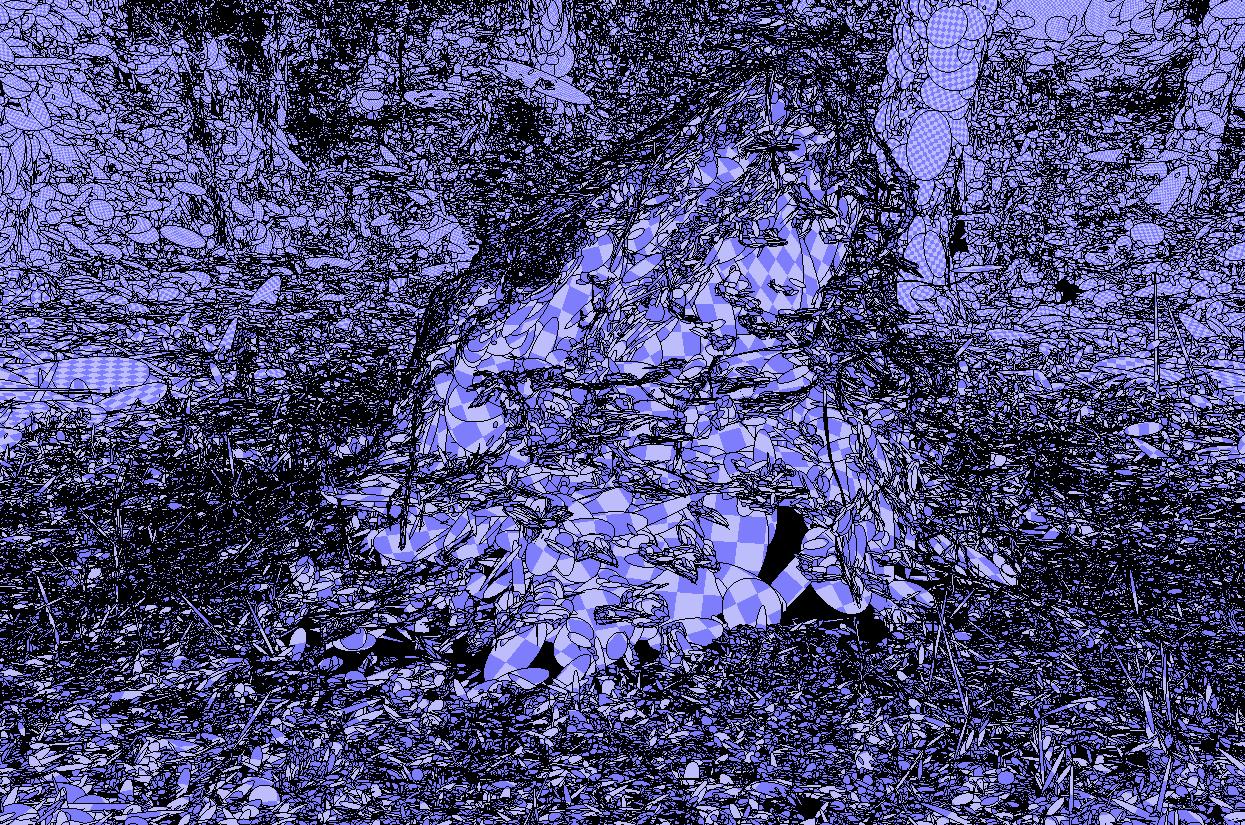}\\
    \end{tabular}
	\caption{\textbf{MipNeRF-360 outdoor scenes.} We provide test-set renders and texel visualizations of \ours with $10^8$ texels applied to the outdoor scenes of MipNeRF-360. Each cell of the checkered Gaussians indicates a single texel. Note that the texel visualizations omit Gaussians with opacity $< 0.5$.\label{fig:m360_outdoor}}
    \vspace{-1em}
\end{figure*}

\begin{figure*}
    \newcommand{\vlabel}[1]{\vspace{3pt}#1\vspace{1pt}}
    \setlength\tmpcolwidth{3.2in}
	\setlength{\tabcolsep}{1pt}
    \renewcommand{\arraystretch}{0.5}
	\centering
	\begin{tabular}{M{0.0in}M{\tmpcolwidth}M{\tmpcolwidth}}
        & \vlabel{Render} & \vlabel{Texels}\\
        &
        \includegraphics[width=\tmpcolwidth,trim={0px 150px 0px 0px},clip]{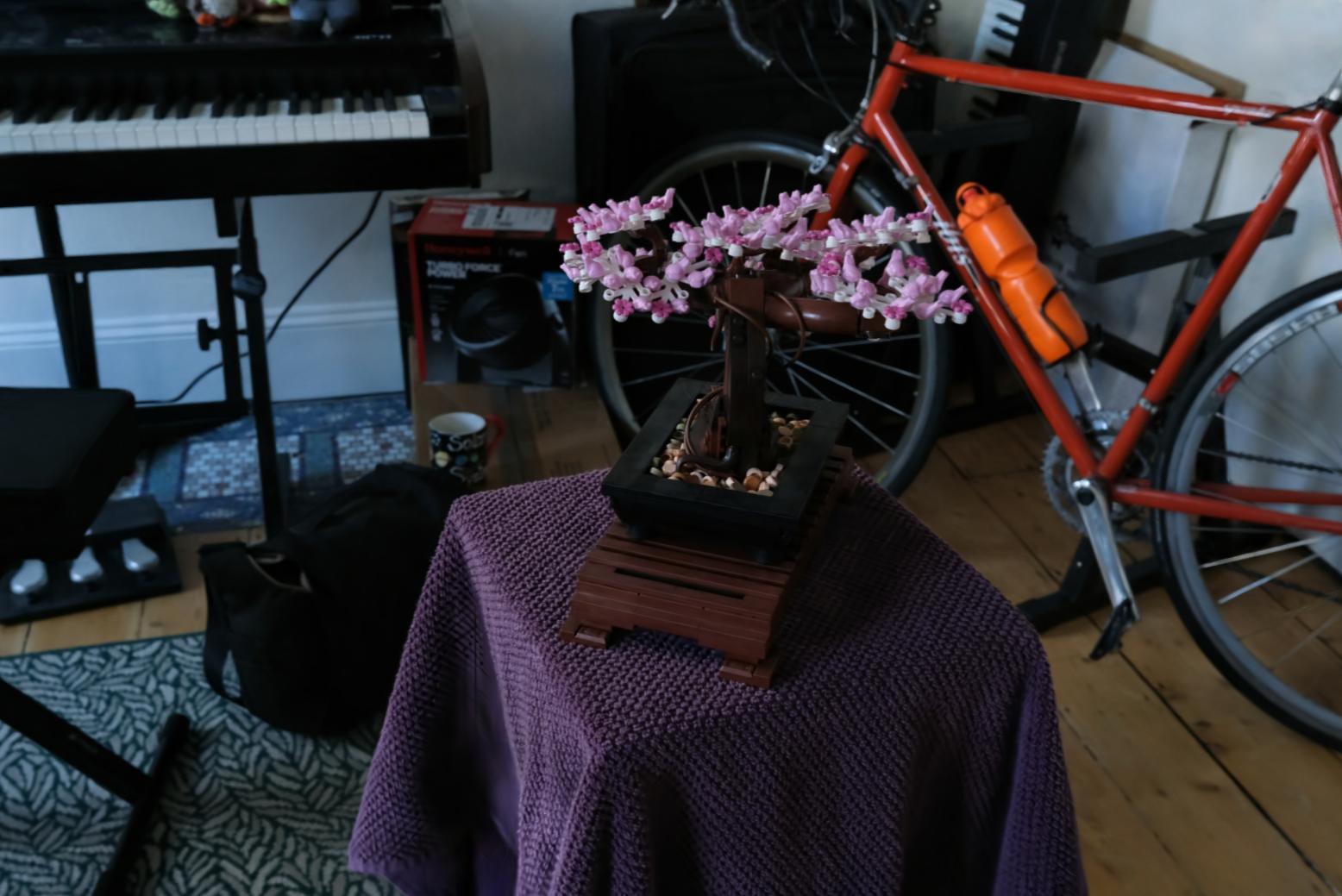}
        &
        \includegraphics[width=\tmpcolwidth,trim={0px 150px 0px 0px},clip]{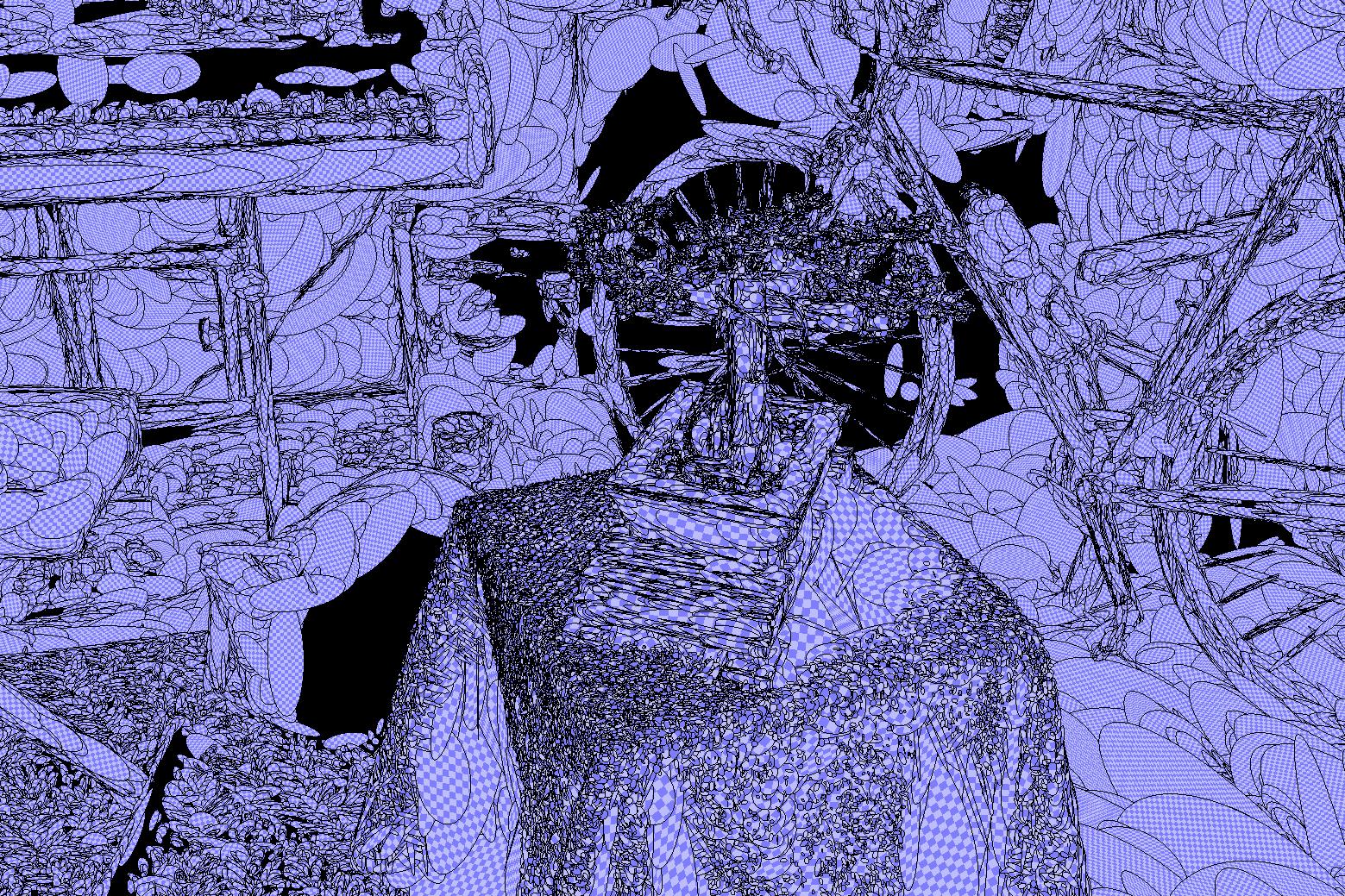}\\
        &
        \includegraphics[width=\tmpcolwidth,trim={0px 150px 0px 0px},clip]{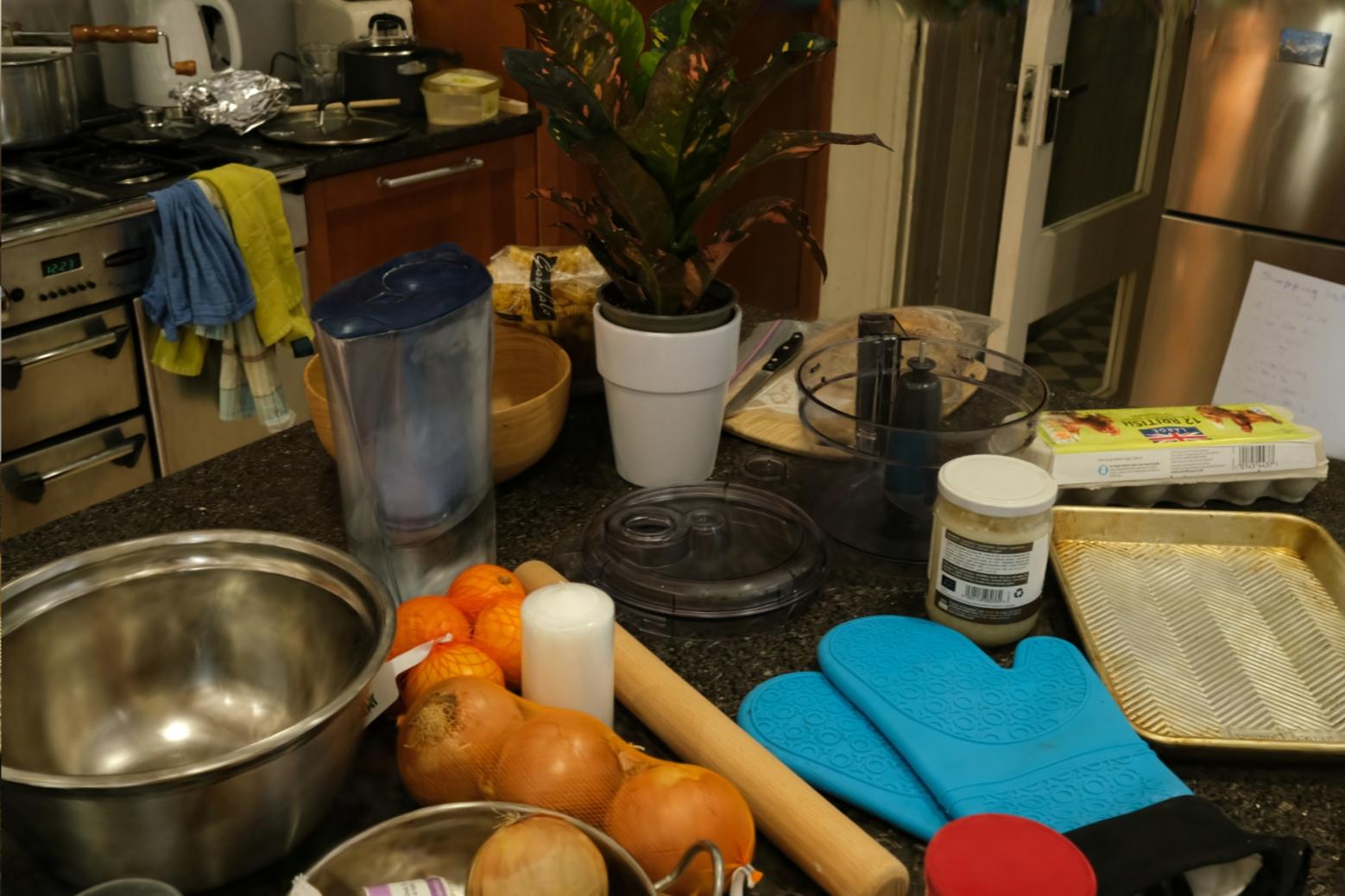}
        &
        \includegraphics[width=\tmpcolwidth,trim={0px 150px 0px 0px},clip]{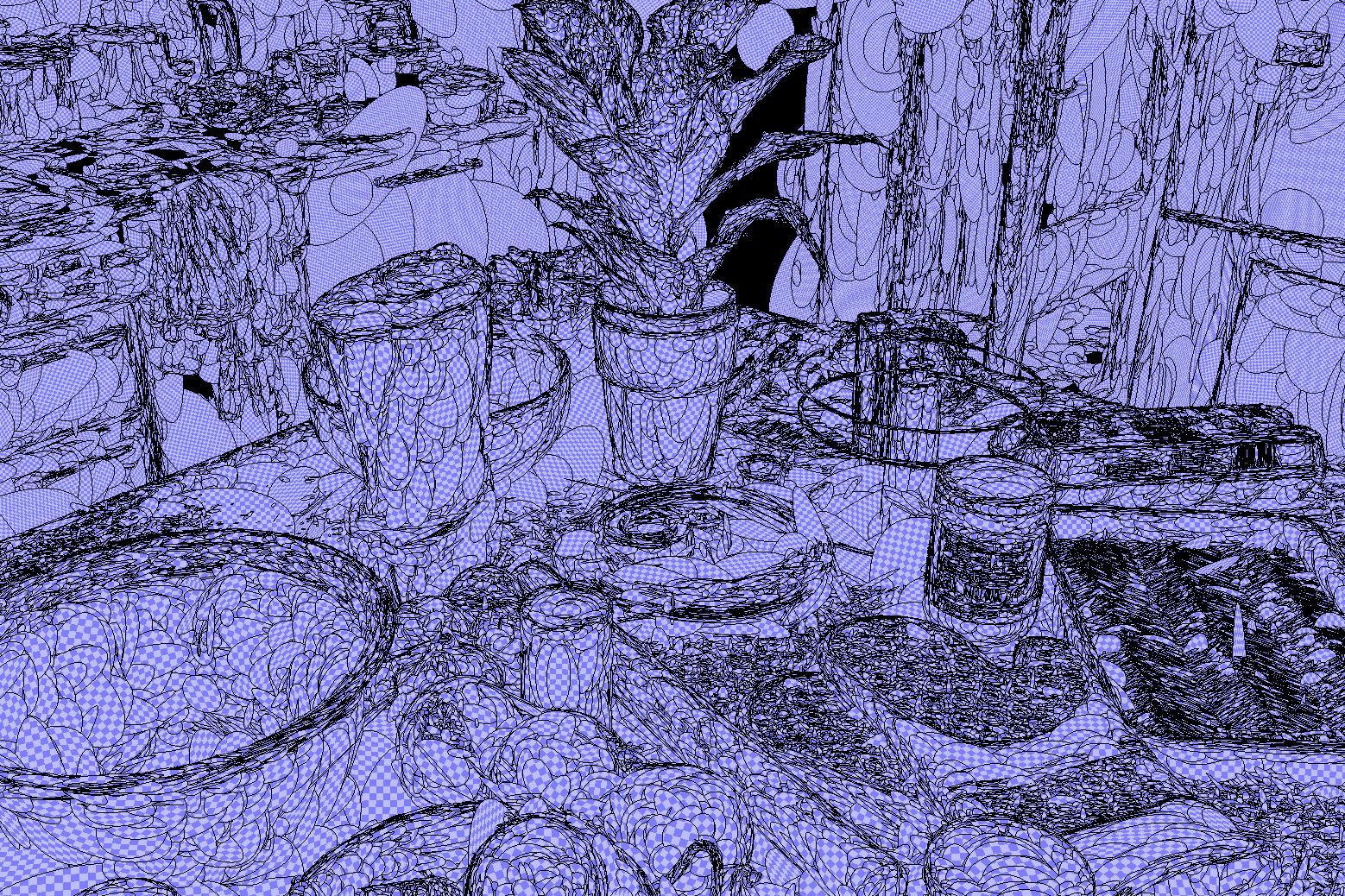}\\
        &
        \includegraphics[width=\tmpcolwidth,trim={0px 150px 0px 0px},clip]{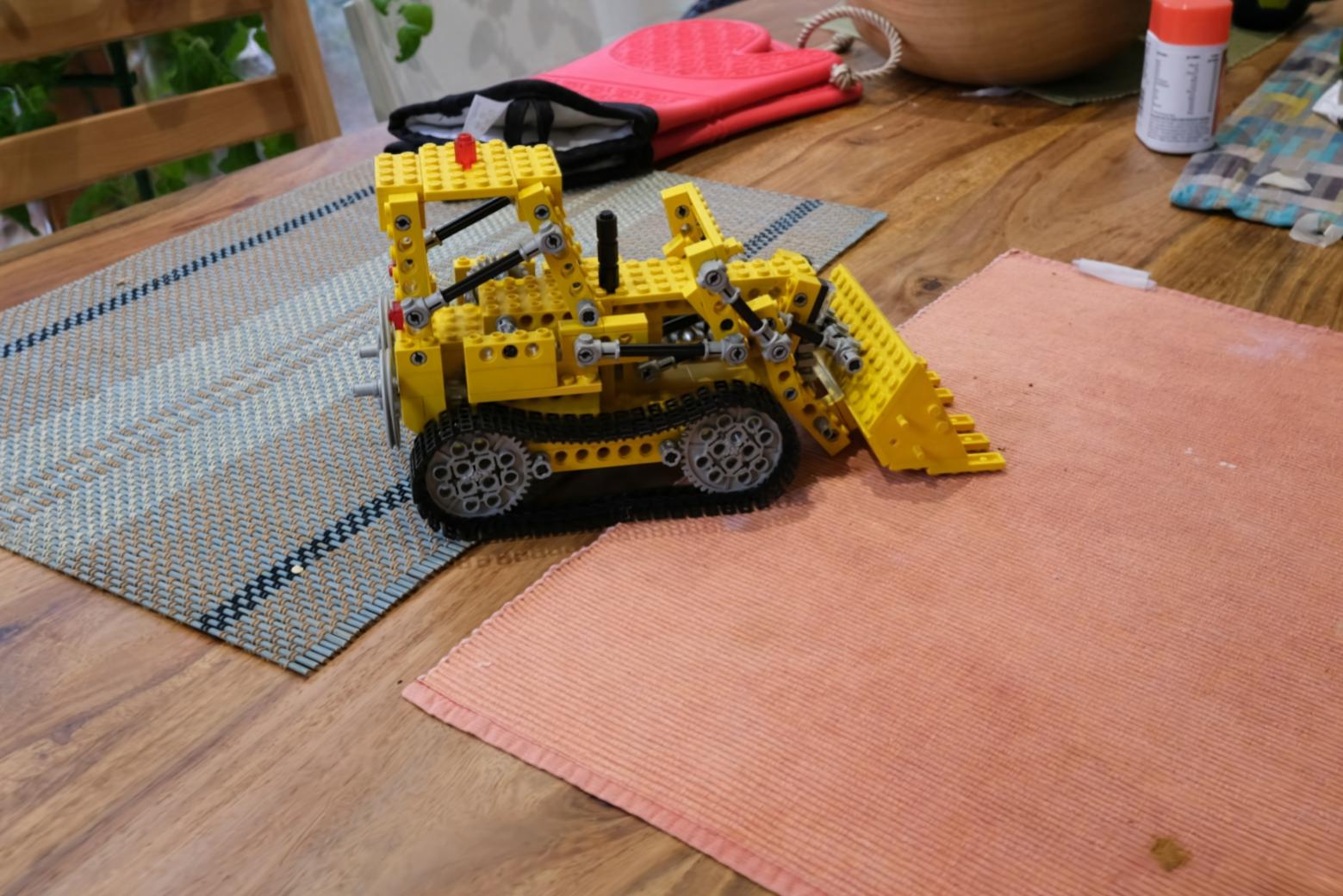}
        &
        \includegraphics[width=\tmpcolwidth,trim={0px 150px 0px 0px},clip]{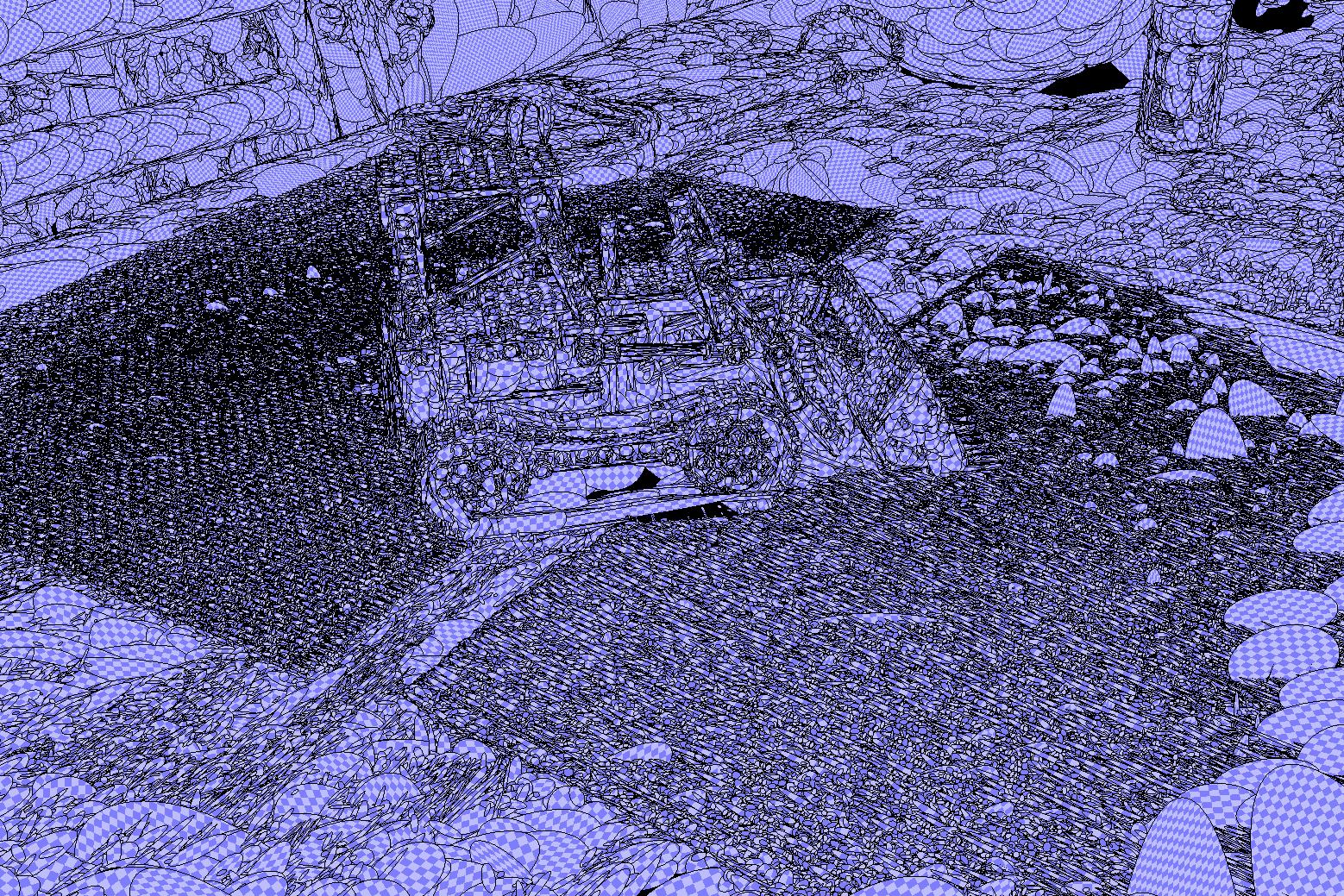}\\
        &
        \includegraphics[width=\tmpcolwidth,trim={0px 150px 0px 0px},clip]{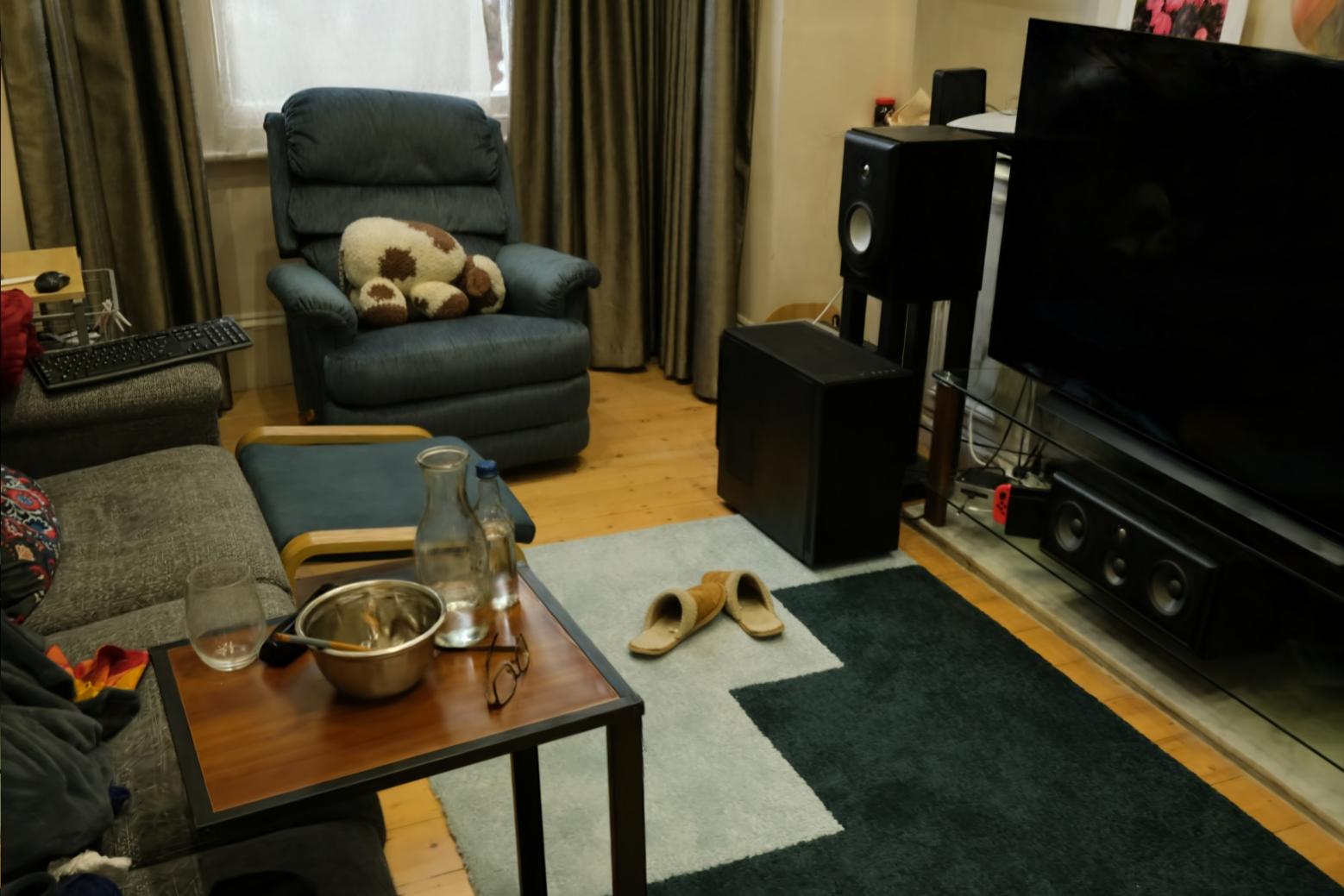}
        &
        \includegraphics[width=\tmpcolwidth,trim={0px 150px 0px 0px},clip]{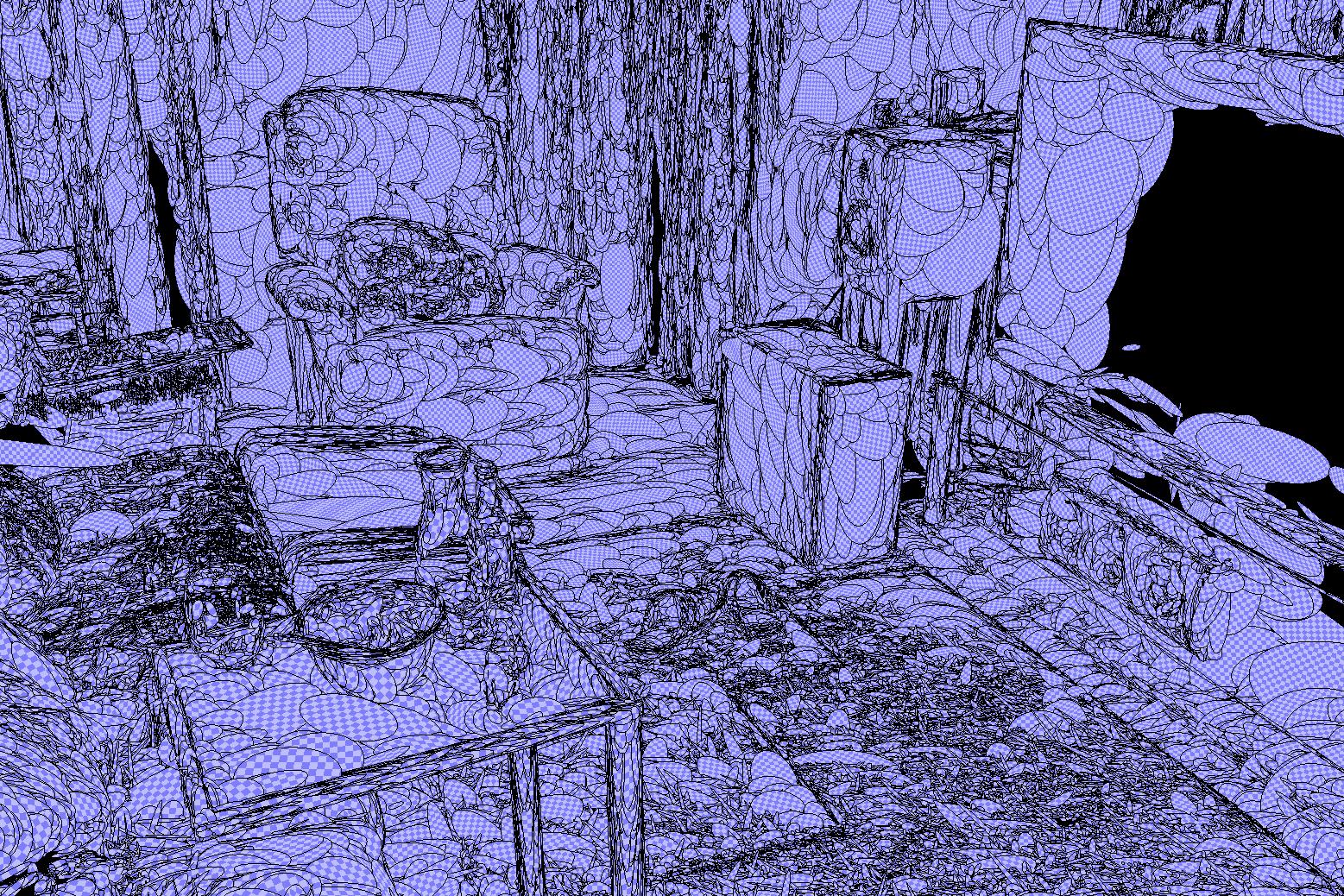}\\
    \end{tabular}
	\caption{\textbf{MipNeRF-360 indoor scenes.} We provide test-set renders and texel visualizations of \ours with $10^8$ texels applied to the indoor scenes of MipNeRF-360. Each cell of the checkered Gaussians indicates a single texel. Note that the texel visualizations omit Gaussians with opacity $< 0.5$. \label{fig:m360_indoor}}
    \vspace{-1em}
\end{figure*}

\begin{table*}[t]
  \begin{center}
  {{
    \setlength\tabcolsep{8.0pt}
\begin{tabular}{lccccccccc}
\toprule
Method & Chair & Drums & Ficus & Hotdog & Lego & Materials & Mic & Ship & Mean\\ %
\midrule
3DGS & 35.90 & 26.16 & 34.85 & 37.70 & 35.78 & 30.00 & 35.42 & 30.90 & 33.34\\
2DGS & 35.41 & 26.12 & 35.39 & 37.47 & 35.25 & 29.74 & 35.20 & 30.66 & 33.15\\
\cmidrule(l{2pt}r{2pt}){2-9}\cmidrule(l{2pt}r{2pt}){10-10}
Texture-GS & 30.20 & 24.93 & 30.63 & 32.52 & 28.77 & 26.26 & 30.81 & 27.63 & 28.97\\
\ours & 35.51 & 26.06 & 35.65 & 37.49 & 35.51 & 29.72 & 35.28 & 30.80 & 33.25\\
\midrule
3DGS & 0.987 & 0.955 & 0.987 & 0.985 & 0.983 & 0.960 & 0.992 & 0.907 & 0.969\\
2DGS & 0.987 & 0.954 & 0.988 & 0.985 & 0.981 & 0.958 & 0.991 & 0.903 & 0.968\\
\cmidrule(l{2pt}r{2pt}){2-9}\cmidrule(l{2pt}r{2pt}){10-10}
Texture-GS & 0.950 & 0.937 & 0.968 & 0.962 & 0.935 & 0.925 & 0.975 & 0.853 & 0.938\\
\ours & 0.987 & 0.954 & 0.988 & 0.985 & 0.982 & 0.958 & 0.991 & 0.904 & 0.969\\
\midrule
3DGS & 0.012 & 0.037 & 0.012 & 0.020 & 0.016 & 0.034 & 0.006 & 0.107 & 0.030\\
2DGS & 0.009 & 0.039 & 0.008 & 0.014 & 0.012 & 0.021 & 0.006 & 0.087 & 0.024\\
\cmidrule(l{2pt}r{2pt}){2-9}\cmidrule(l{2pt}r{2pt}){10-10}
Texture-GS & 0.049 & 0.056 & 0.023 & 0.042 & 0.047 & 0.055 & 0.020 & 0.146 & 0.055\\
\ours & 0.009 & 0.038 & 0.008 & 0.013 & 0.011 & 0.020 & 0.006 & 0.085 & 0.024\\
\midrule
3DGS & 270K & 350K & 290K & 150K & 320K & 280K & 310K & 330K & 290K\\
2DGS & 100K & 140K & 50K & 70K & 160K & 130K & 150K & 160K & 120K\\
\cmidrule(l{2pt}r{2pt}){2-9}\cmidrule(l{2pt}r{2pt}){10-10}
Texture-GS & 60K & 80K & 20K & 40K & 100K & 60K & 30K & 40K & 50K\\
\ours & 100K & 140K & 50K & 70K & 160K & 130K & 150K & 160K & 120K\\
\bottomrule
\end{tabular}
}}
\end{center}
\caption{Novel view synthesis metrics for the synthetic Blender dataset. We report PSNR $\uparrow$, SSIM $\uparrow$, LPIPS $\downarrow$\editadd{, and number of Gaussians}, respectively, for individual scenes.\label{tab:supp_nvs_syn}}

  \begin{center}
    {\footnotesize{
    \setlength\tabcolsep{4.5pt}
\begin{tabular}{lcccccccccccccccc}
\toprule
Method & 24 & 37 & 40 & 55 & 63 & 65 & 69 & 83 & 97 & 105 & 106 & 110 & 114 & 118 & 122 & Mean\\
\midrule
3DGS & 30.48 & 26.97 & 29.57 & 31.76 & 35.44 & 31.29 & 28.24 & 38.42 & 30.08 & 34.12 & 35.07 & 34.61 & 31.08 & 37.59 & 38.34 & 32.87\\
2DGS & 29.99 & 26.18 & 28.77 & 31.30 & 34.35 & 30.76 & 27.90 & 37.72 & 29.77 & 33.46 & 34.93 & 33.32 & 30.58 & 36.69 & 37.60 & 32.22\\
\cmidrule(l{2pt}r{2pt}){2-16}\cmidrule(l{2pt}r{2pt}){17-17}
Texture-GS & 27.63 & 25.31 & 27.83 & 27.04 & 33.62 & 30.02 & 27.72 & 37.22 & 28.43 & 31.91 & 32.73 & 30.41 & 28.95 & 34.50 & 34.65 & 30.53\\
\ours & 30.77 & 26.91 & 29.63 & 31.85 & 35.28 & 31.17 & 28.10 & 38.42 & 30.21 & 33.92 & 35.27 & 34.55 & 31.16 & 37.65 & 38.23 & 32.87\\
\midrule
3DGS & 0.937 & 0.919 & 0.917 & 0.963 & 0.971 & 0.965 & 0.938 & 0.981 & 0.950 & 0.963 & 0.963 & 0.970 & 0.951 & 0.975 & 0.980 & 0.956\\
2DGS & 0.909 & 0.888 & 0.881 & 0.950 & 0.958 & 0.958 & 0.927 & 0.972 & 0.937 & 0.945 & 0.949 & 0.952 & 0.936 & 0.963 & 0.970 & 0.940\\
\cmidrule(l{2pt}r{2pt}){2-16}\cmidrule(l{2pt}r{2pt}){17-17}
Texture-GS & 0.875 & 0.886 & 0.847 & 0.893 & 0.949 & 0.955 & 0.908 & 0.970 & 0.923 & 0.924 & 0.927 & 0.936 & 0.907 & 0.944 & 0.955 & 0.920\\
\ours & 0.937 & 0.921 & 0.916 & 0.965 & 0.971 & 0.965 & 0.935 & 0.980 & 0.949 & 0.961 & 0.962 & 0.968 & 0.954 & 0.974 & 0.979 & 0.956\\
\midrule
3DGS & 0.056 & 0.064 & 0.103 & 0.039 & 0.033 & 0.050 & 0.076 & 0.030 & 0.057 & 0.049 & 0.054 & 0.060 & 0.053 & 0.037 & 0.025 & 0.052\\
2DGS & 0.089 & 0.087 & 0.147 & 0.045 & 0.074 & 0.093 & 0.144 & 0.059 & 0.098 & 0.094 & 0.096 & 0.093 & 0.095 & 0.074 & 0.060 & 0.090\\
\cmidrule(l{2pt}r{2pt}){2-16}\cmidrule(l{2pt}r{2pt}){17-17}
Texture-GS & 0.086 & 0.080 & 0.154 & 0.089 & 0.051 & 0.061 & 0.123 & 0.035 & 0.065 & 0.093 & 0.080 & 0.118 & 0.103 & 0.063 & 0.047 & 0.083\\
\ours & 0.032 & 0.042 & 0.079 & 0.029 & 0.022 & 0.039 & 0.069 & 0.020 & 0.040 & 0.037 & 0.039 & 0.040 & 0.042 & 0.024 & 0.018 & 0.038\\
\midrule
3DGGS & 670K & 700K & 1010K & 740K & 130K & 190K & 210K & 80K & 360K & 220K & 280K & 110K & 300K & 170K & 180K & 360K\\
2DGS & 320K & 440K & 590K & 330K & 80K & 90K & 110K & 50K & 180K & 120K & 130K & 60K & 110K & 100K & 90K & 190K\\
\cmidrule(l{2pt}r{2pt}){2-16}\cmidrule(l{2pt}r{2pt}){17-17}
Texture-GS & 60K & 100K & 120K & 110K & 20K & 30K & 50K & 20K & 60K & 40K & 50K & 20K & 50K & 40K & 40K & 50K\\
\ours & 320K & 440K & 590K & 330K & 80K & 90K & 110K & 50K & 180K & 120K & 130K & 60K & 110K & 100K & 90K & 190K\\
\bottomrule
\end{tabular}
}}
\end{center}
\caption{Novel view synthesis metrics for individual scenes on the DTU dataset. We report PSNR $\uparrow$, SSIM $\uparrow$, LPIPS $\downarrow$\editadd{, and number of Gaussians}, respectively.\label{tab:supp_nvs_dtu}}
\end{table*}

\begin{figure*}[!h]
    \newcommand{\vlabel}[1]{\vspace{3pt}#1\vspace{1pt}}
    \setlength\tmpcolwidth{1.0in}
	\setlength{\tabcolsep}{0pt}
    \renewcommand{\arraystretch}{0}
	\centering
	\begin{tabular}{M{0.2in}M{\tmpcolwidth}M{\tmpcolwidth}M{\tmpcolwidth}M{\tmpcolwidth}M{\tmpcolwidth}M{\tmpcolwidth}}
        & \vlabel{2DGS} & \vlabel{\ours} & \vlabel{2DGS} & \vlabel{\ours} & \vlabel{2DGS} & \vlabel{\ours}\\
        \multirow{3}{*}{\rotatebox{90}{$\xleftarrow{\makebox[1.3in]{Number of Gaussians}}$}}
        &
        \includegraphics[width=\tmpcolwidth,trim={20px 80px 20px 80px},clip]{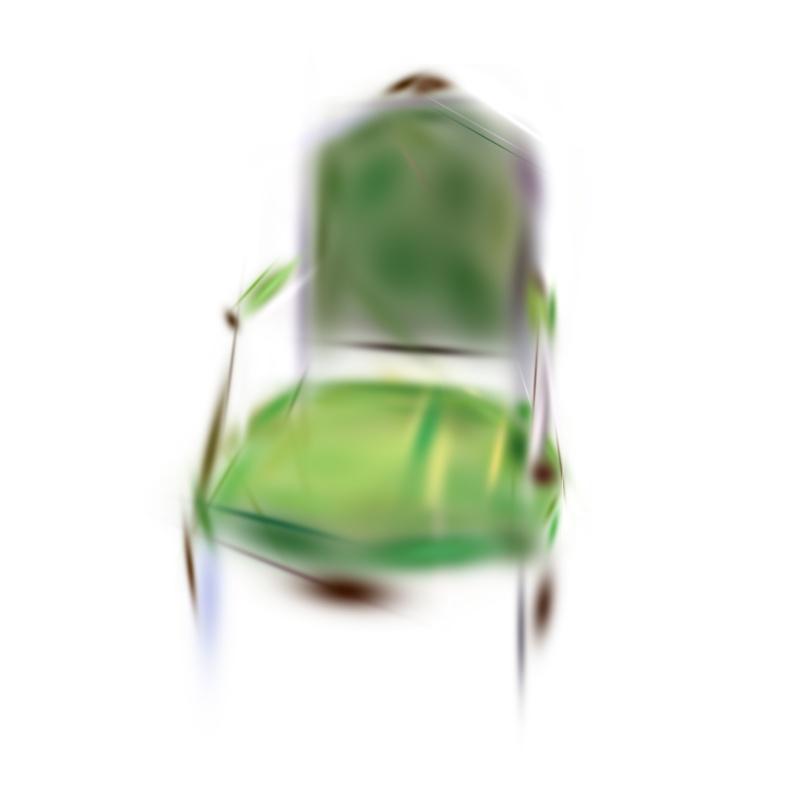}
        & 
        \includegraphics[width=\tmpcolwidth,trim={20px 80px 20px 80px},clip]{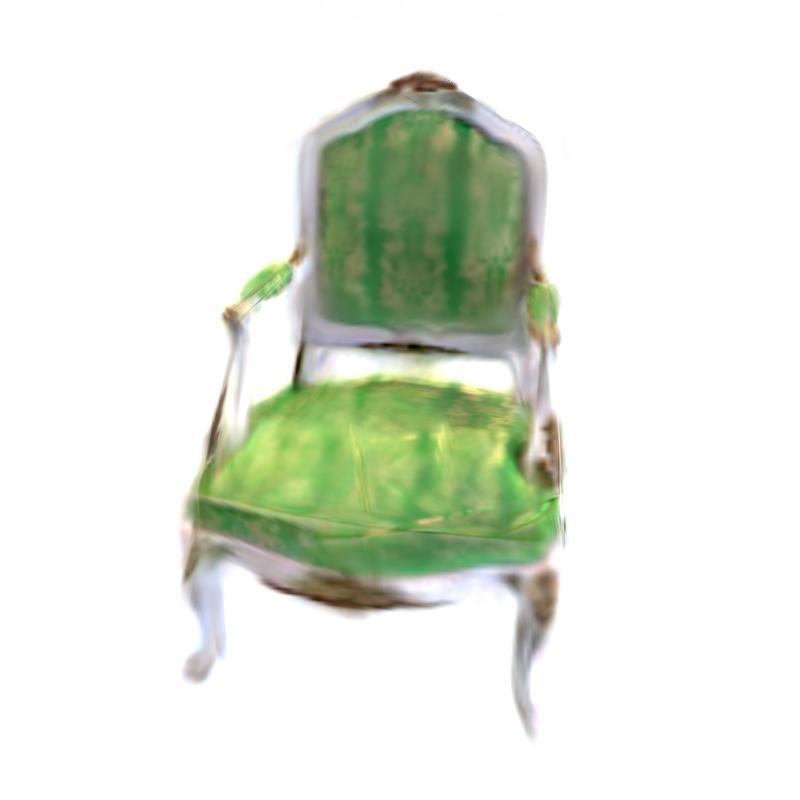}
        & 
        \includegraphics[width=\tmpcolwidth,trim={20px 80px 20px 80px},clip]{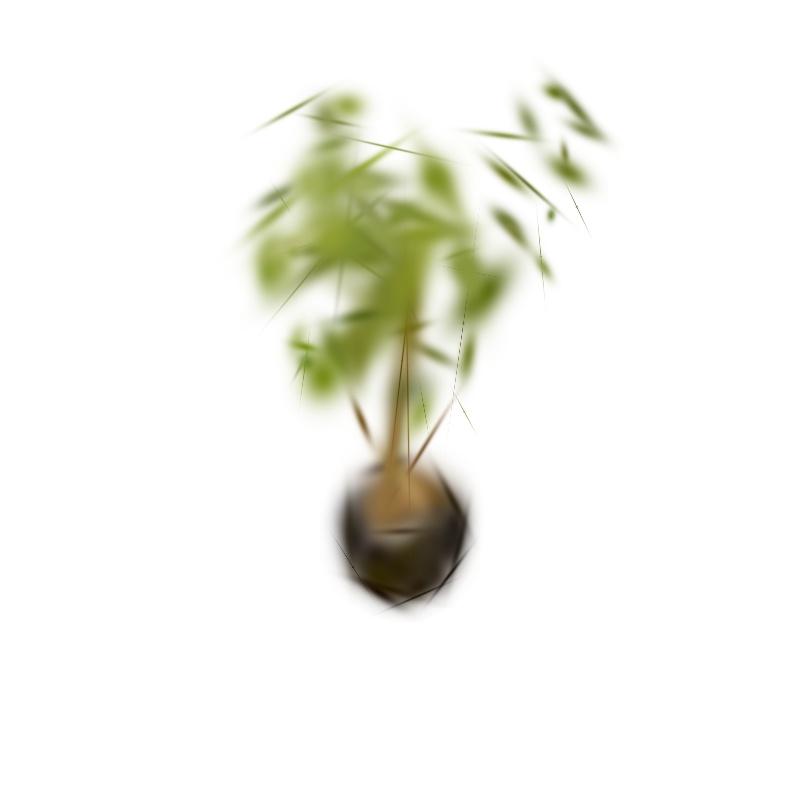}
        & 
        \includegraphics[width=\tmpcolwidth,trim={20px 80px 20px 80px},clip]{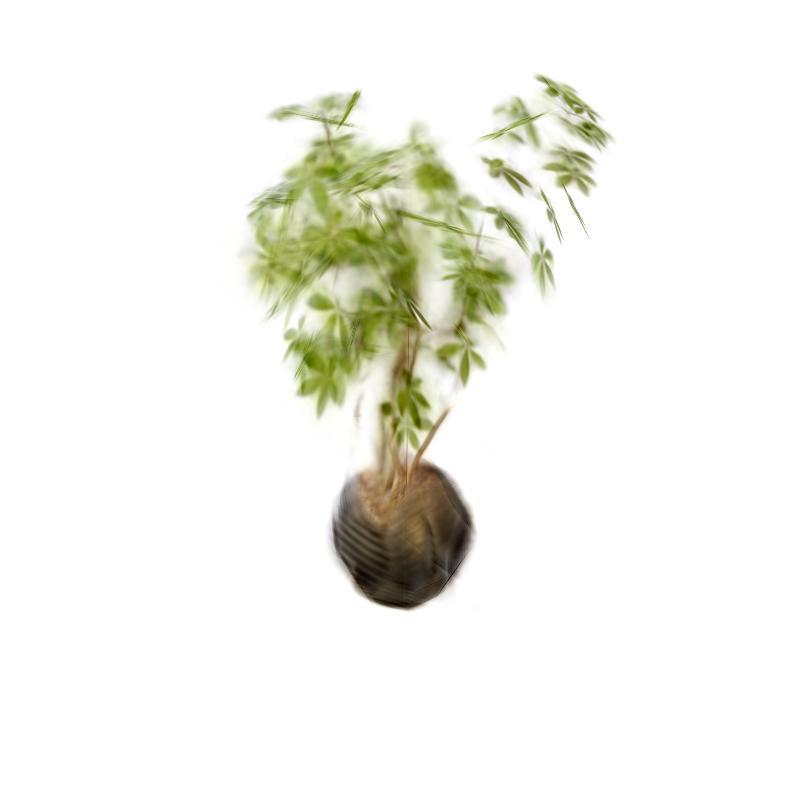}
        & 
        \includegraphics[width=\tmpcolwidth,trim={20px 80px 20px 80px},clip]{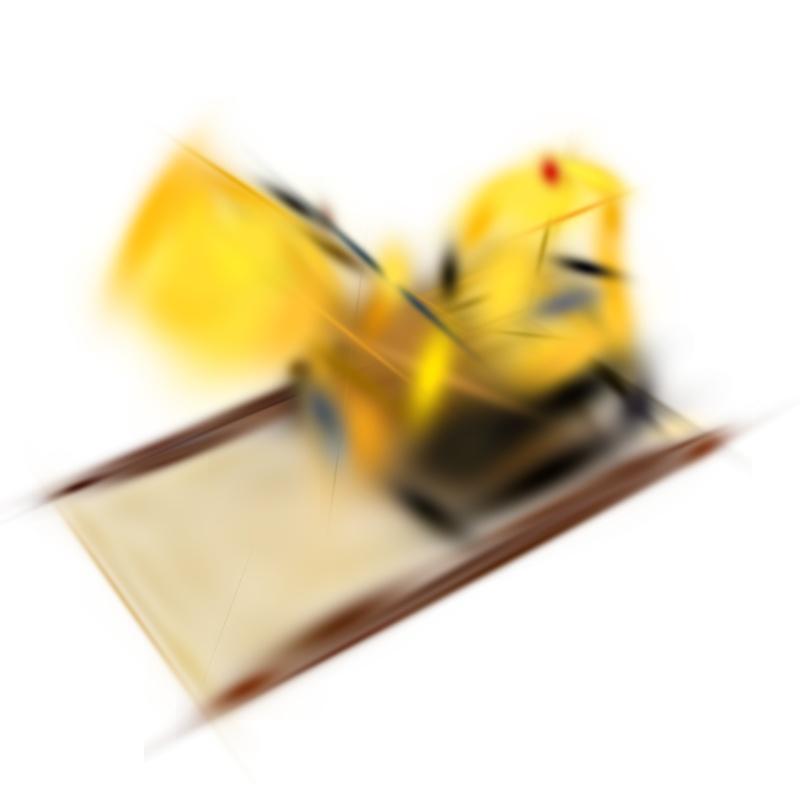}
        & 
        \includegraphics[width=\tmpcolwidth,trim={20px 80px 20px 80px},clip]{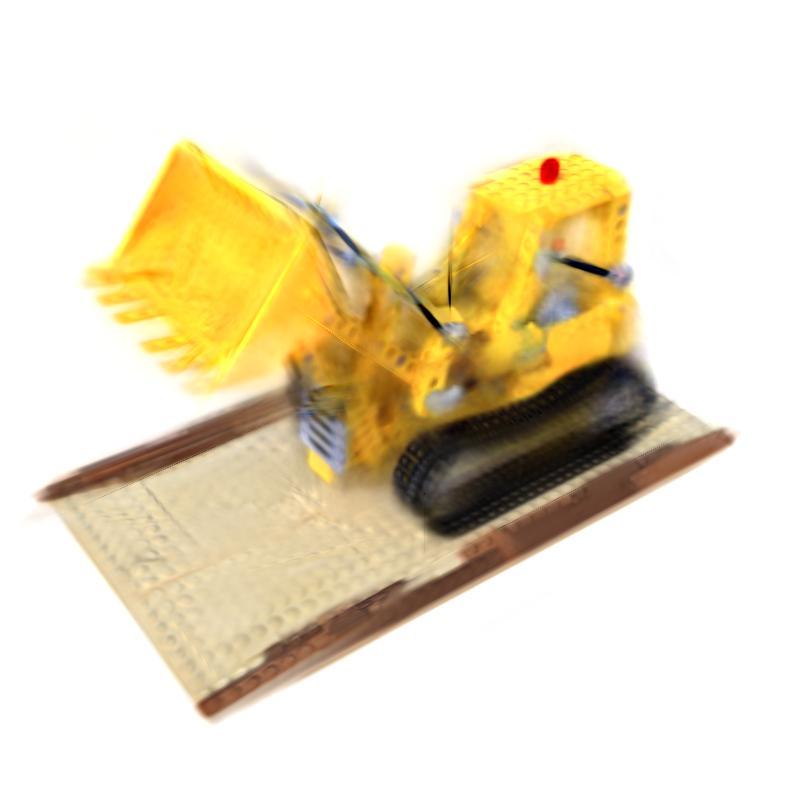}\\
        &
        \includegraphics[width=\tmpcolwidth,trim={20px 80px 20px 80px},clip]{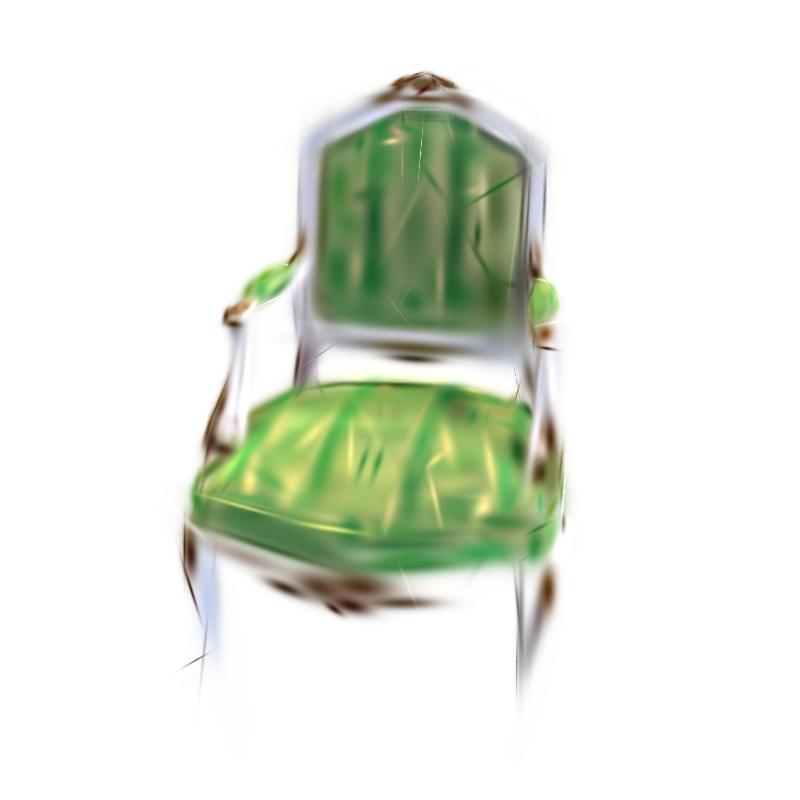}
        & 
        \includegraphics[width=\tmpcolwidth,trim={20px 80px 20px 80px},clip]{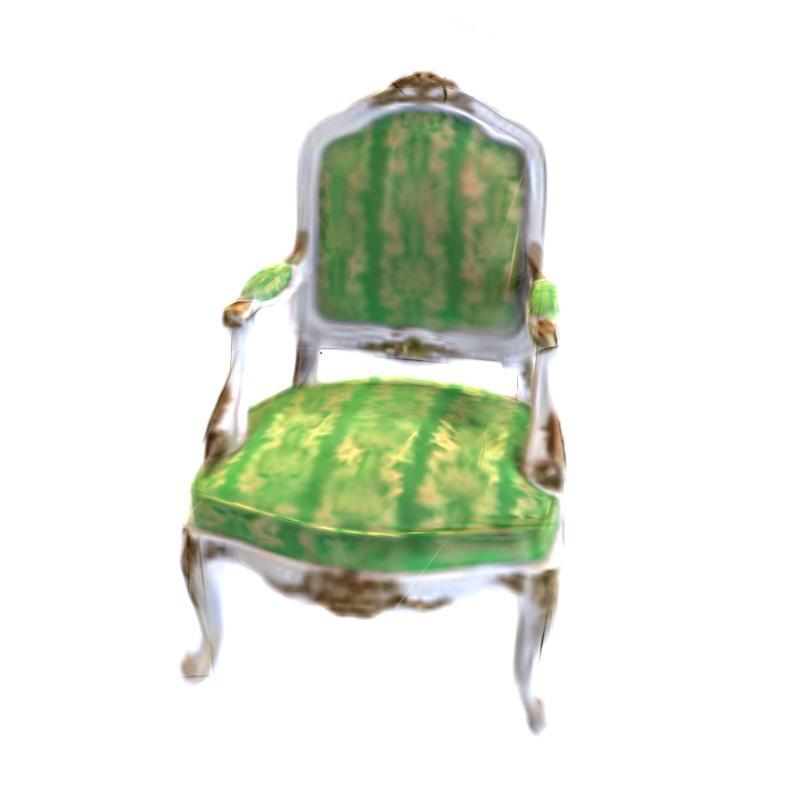}
        & 
        \includegraphics[width=\tmpcolwidth,trim={20px 80px 20px 80px},clip]{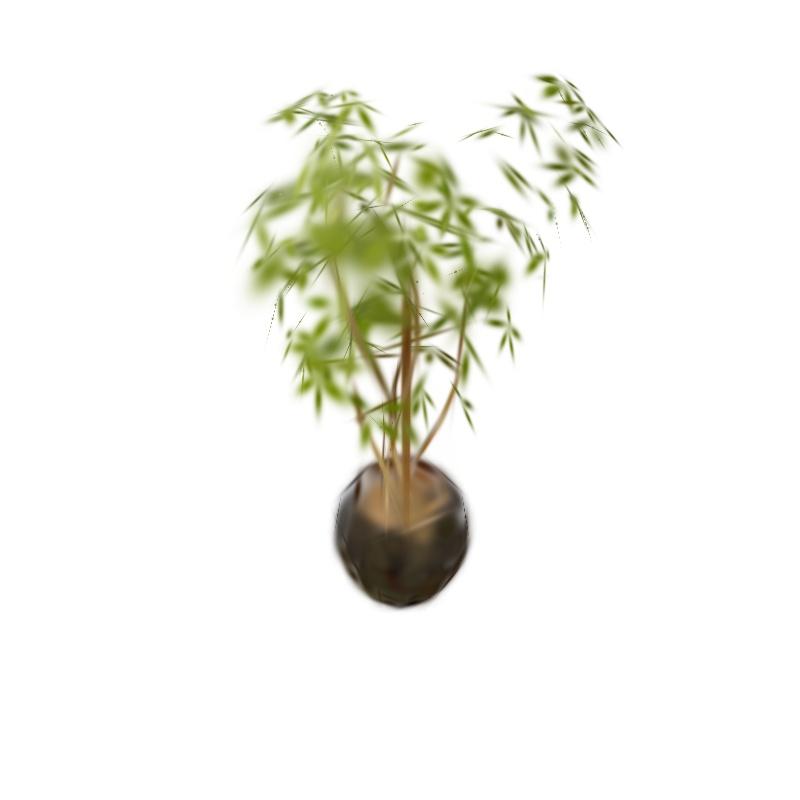}
        & 
        \includegraphics[width=\tmpcolwidth,trim={20px 80px 20px 80px},clip]{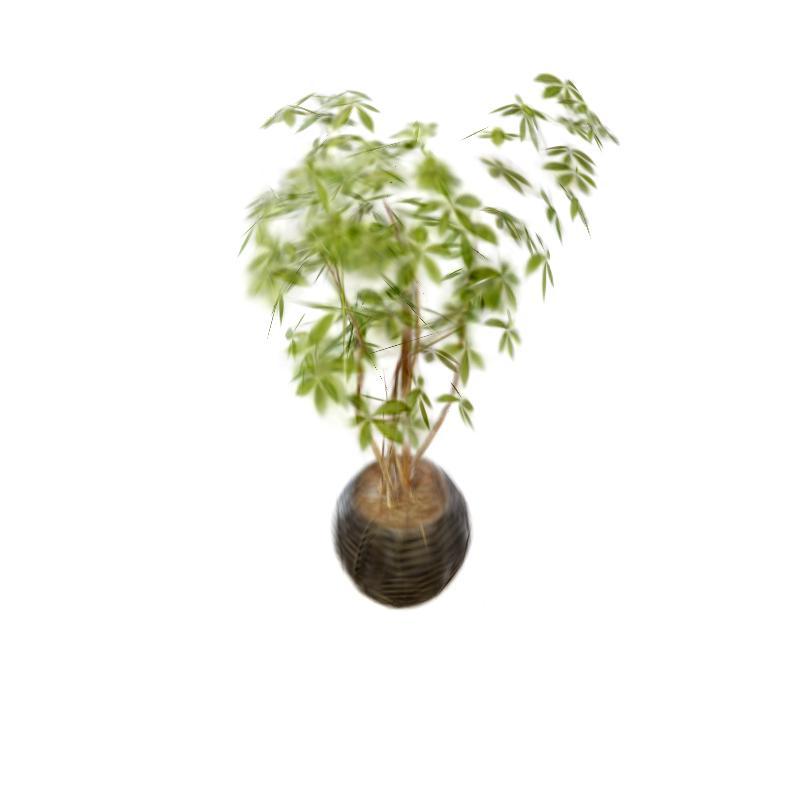}
        & 
        \includegraphics[width=\tmpcolwidth,trim={20px 80px 20px 80px},clip]{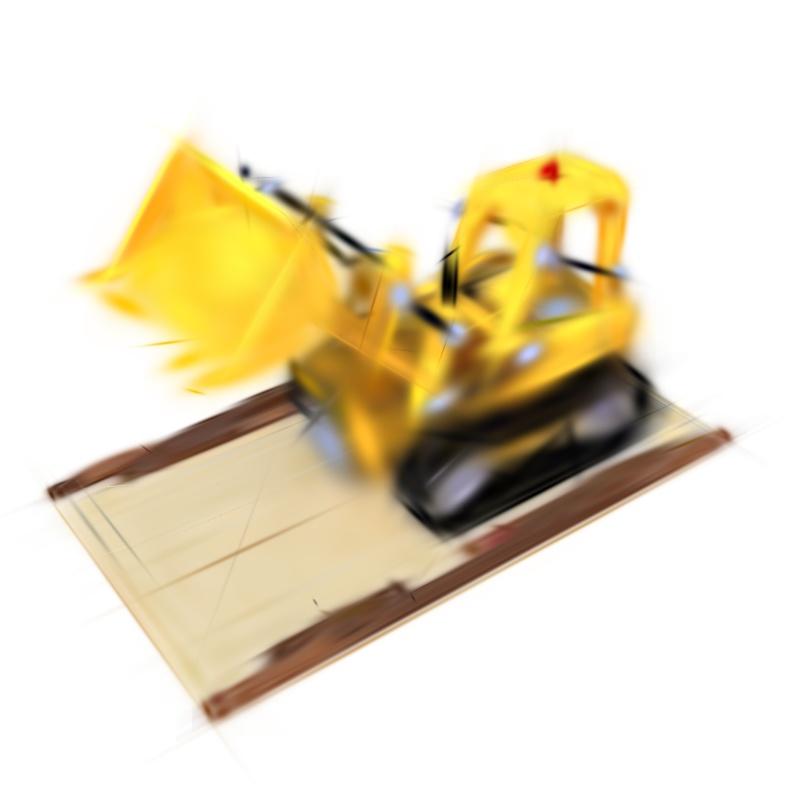}
        & 
        \includegraphics[width=\tmpcolwidth,trim={20px 80px 20px 80px},clip]{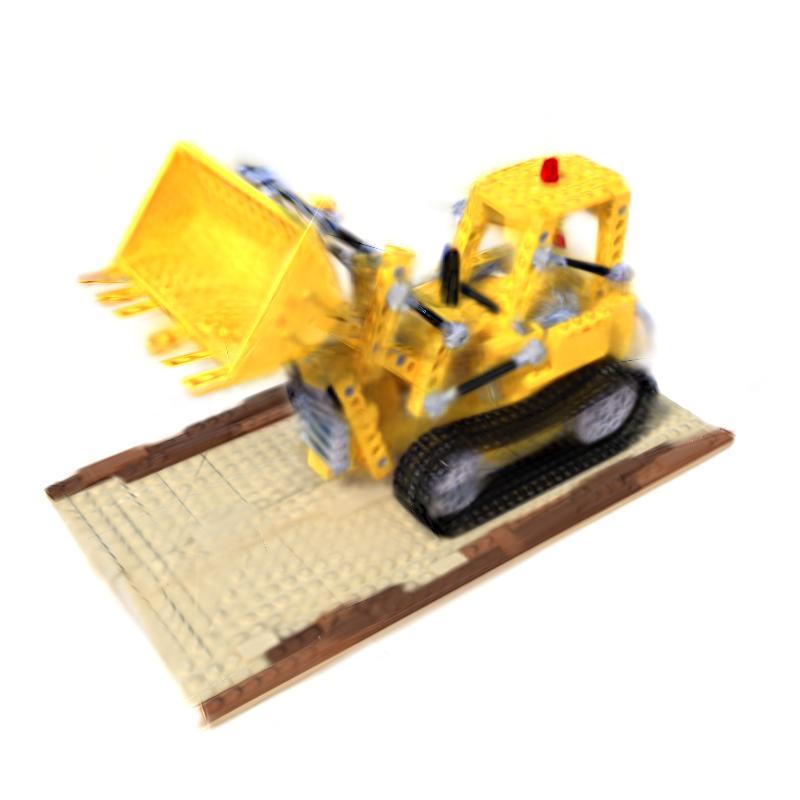}\\
        &
        \includegraphics[width=\tmpcolwidth,trim={20px 80px 20px 80px},clip]{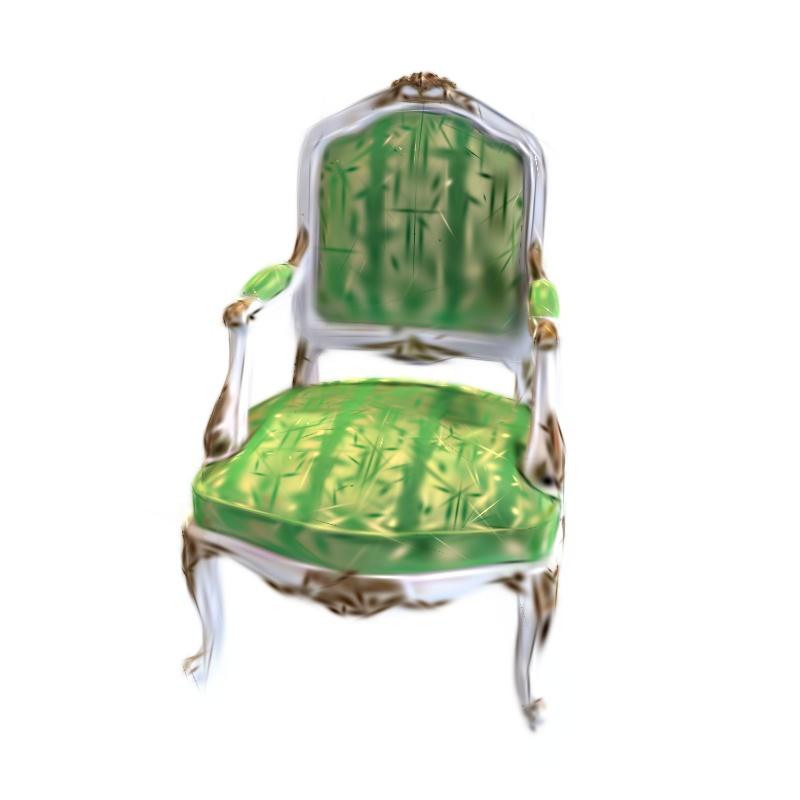}
        & 
        \includegraphics[width=\tmpcolwidth,trim={20px 80px 20px 80px},clip]{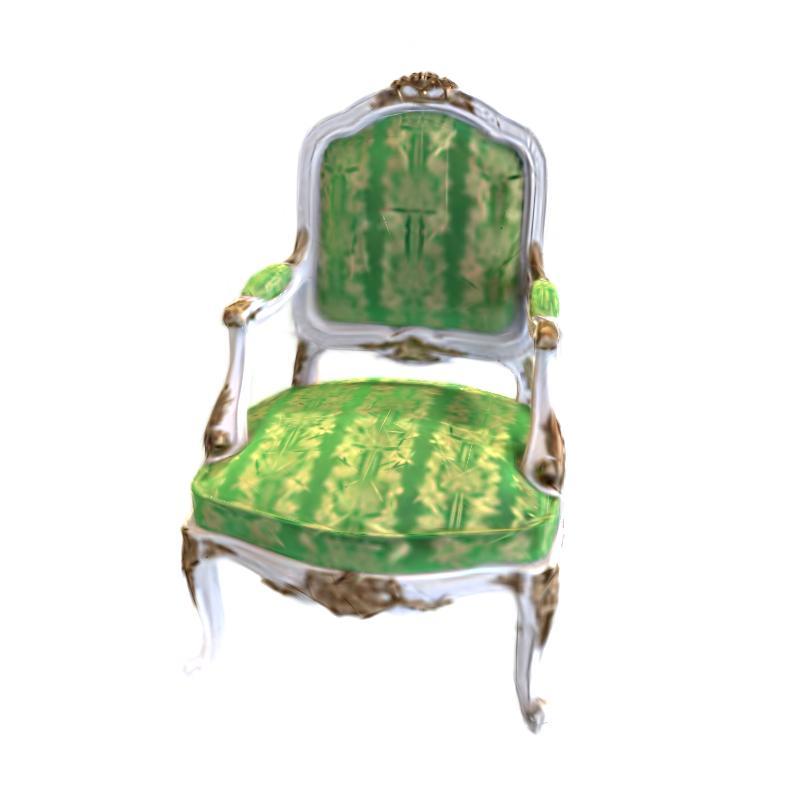}
        & 
        \includegraphics[width=\tmpcolwidth,trim={20px 80px 20px 80px},clip]{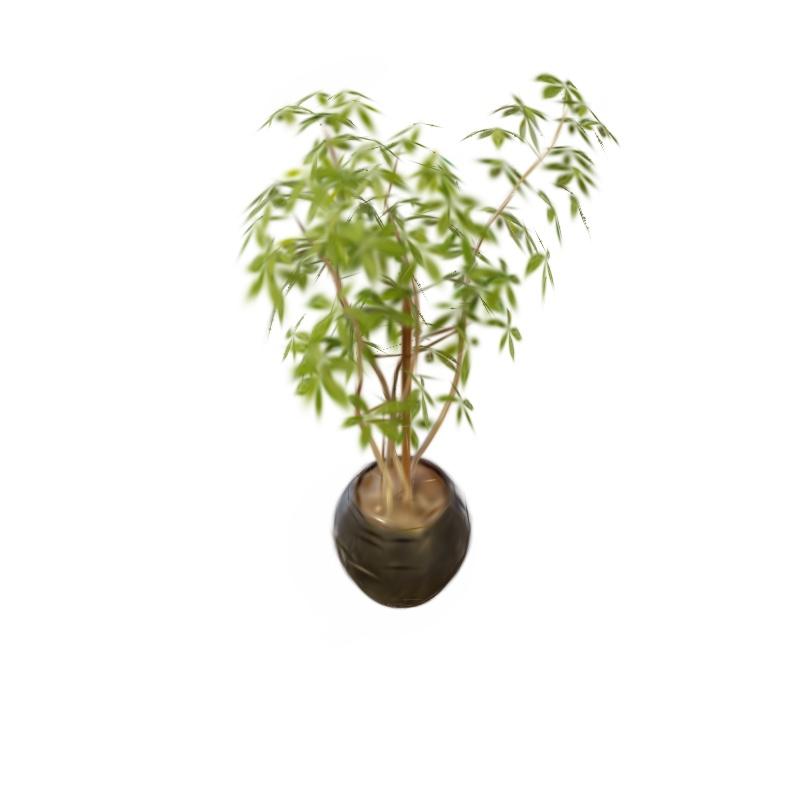}
        & 
        \includegraphics[width=\tmpcolwidth,trim={20px 80px 20px 80px},clip]{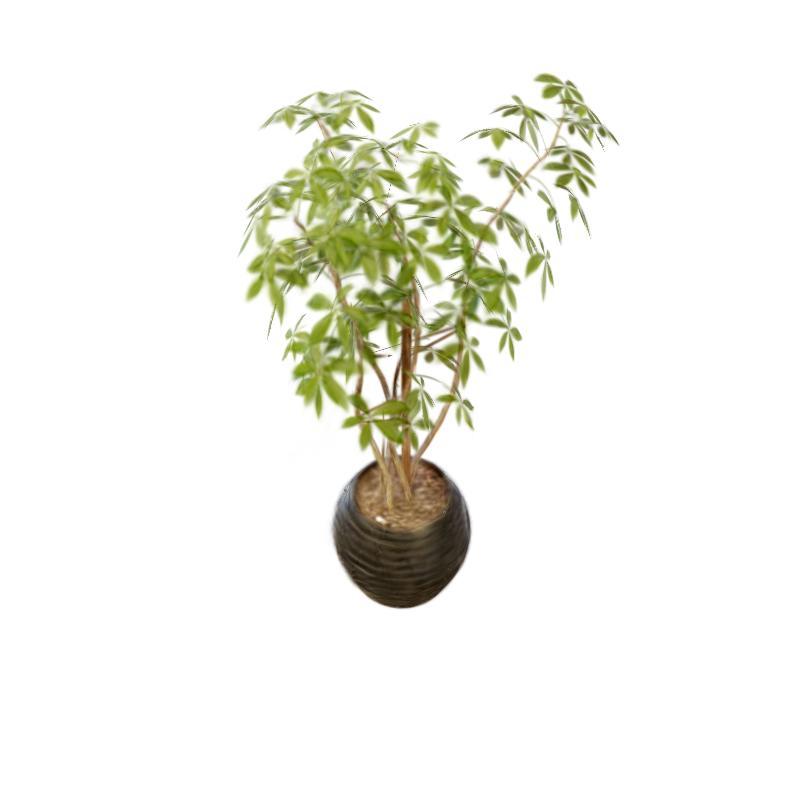}
        & 
        \includegraphics[width=\tmpcolwidth,trim={20px 80px 20px 80px},clip]{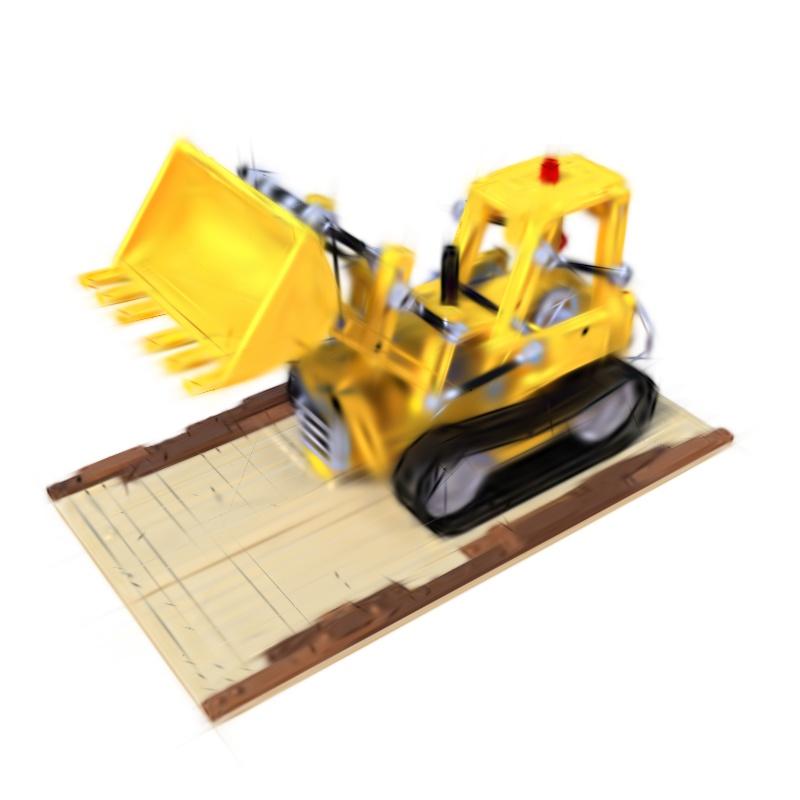}
        & 
        \includegraphics[width=\tmpcolwidth,trim={20px 80px 20px 80px},clip]{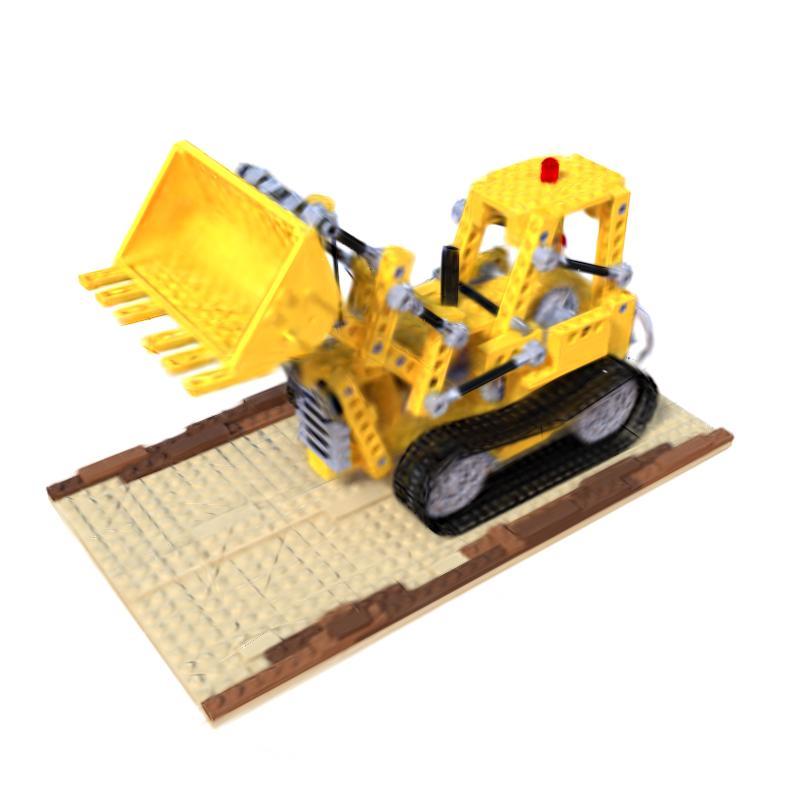}\\
        & \vlabel{2DGS} & \vlabel{\ours} & \vlabel{2DGS} & \vlabel{\ours} &  & \\
        \multirow{3}{*}{\rotatebox{90}{$\xleftarrow{\makebox[1.3in]{Number of Gaussians}}$}}
        &
        \includegraphics[width=\tmpcolwidth,trim={20px 80px 20px 80px},clip]{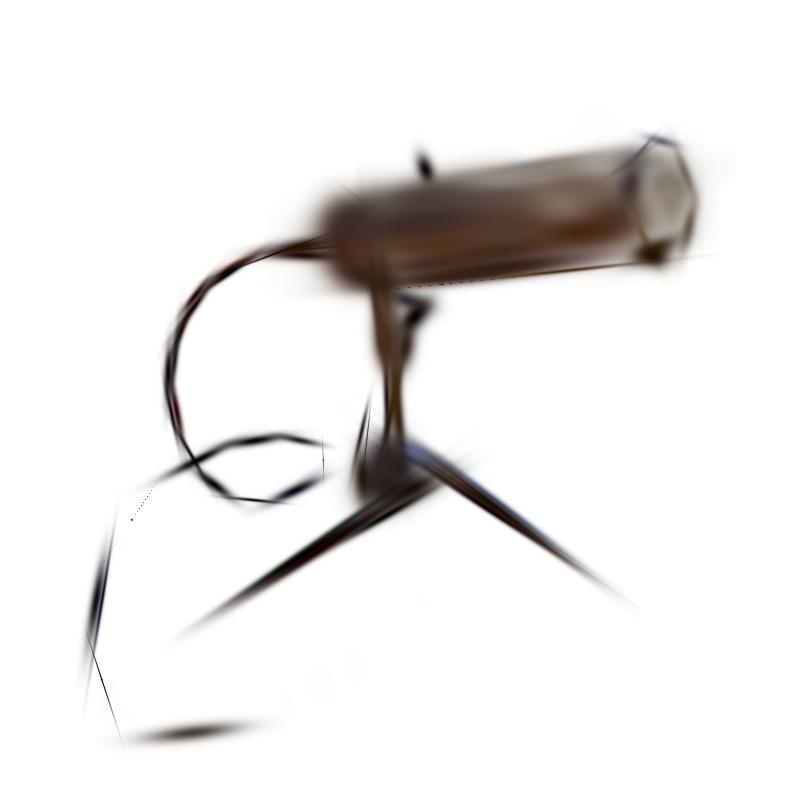}
        & 
        \includegraphics[width=\tmpcolwidth,trim={20px 80px 20px 80px},clip]{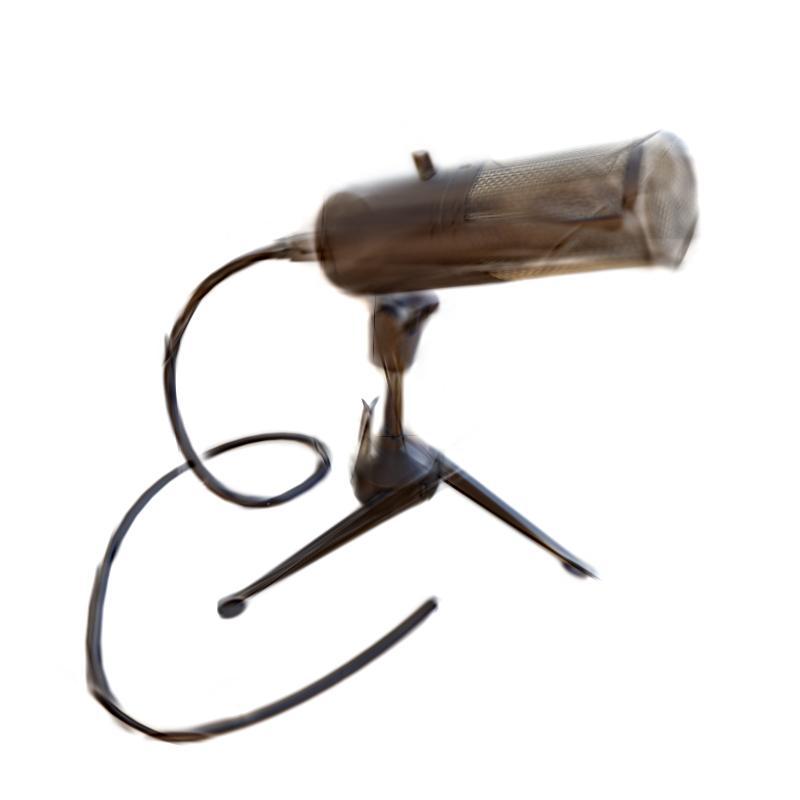}
        &
        \includegraphics[width=\tmpcolwidth,trim={20px 80px 20px 80px},clip]{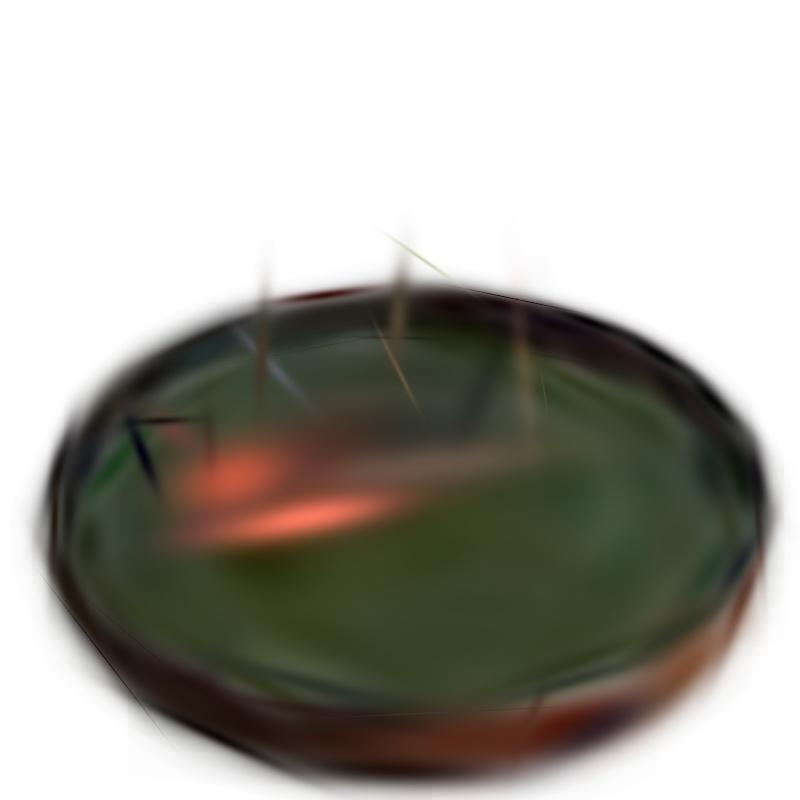}
        & 
        \includegraphics[width=\tmpcolwidth,trim={20px 80px 20px 80px},clip]{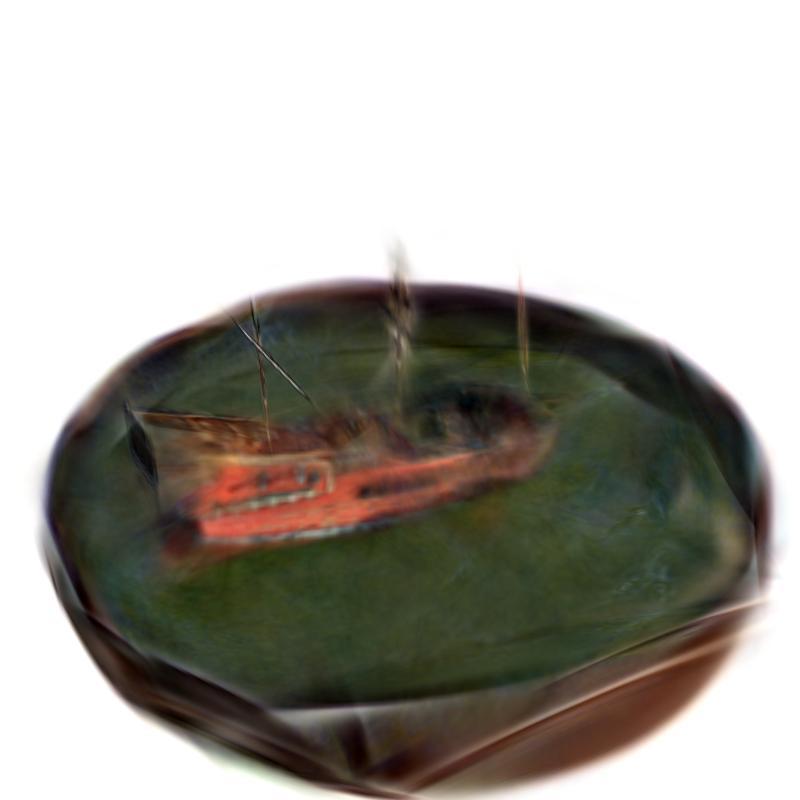}
        &
        &\\
        &
        \includegraphics[width=\tmpcolwidth,trim={20px 80px 20px 80px},clip]{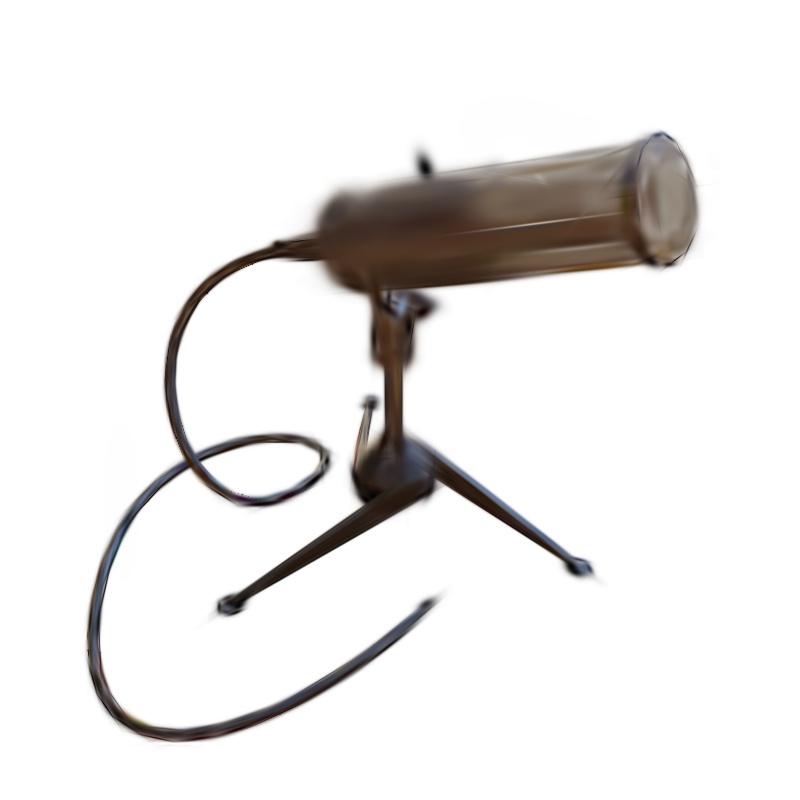}
        & 
        \includegraphics[width=\tmpcolwidth,trim={20px 80px 20px 80px},clip]{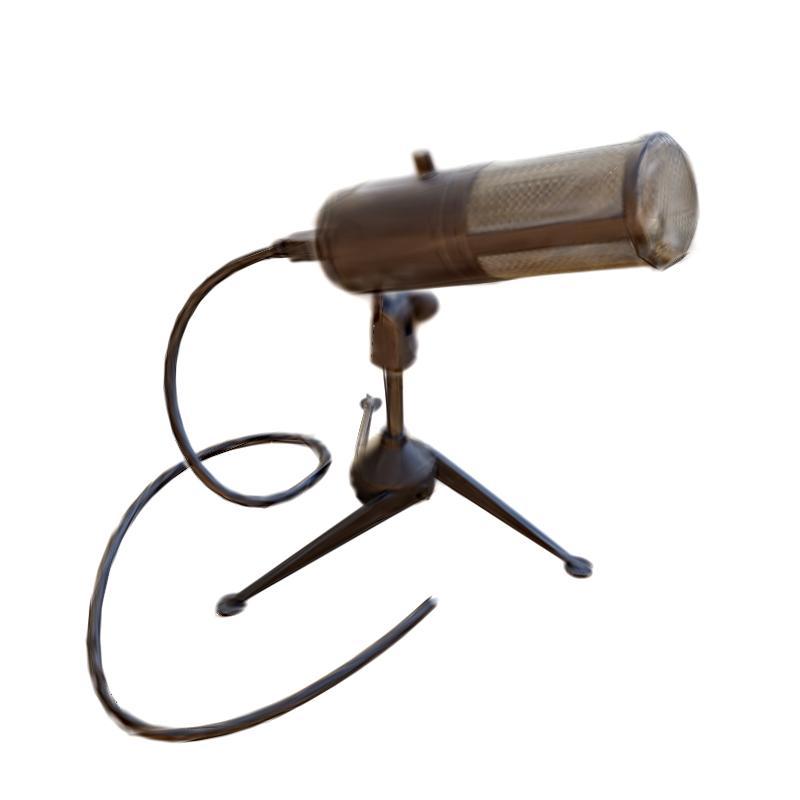}
        &
        \includegraphics[width=\tmpcolwidth,trim={20px 80px 20px 80px},clip]{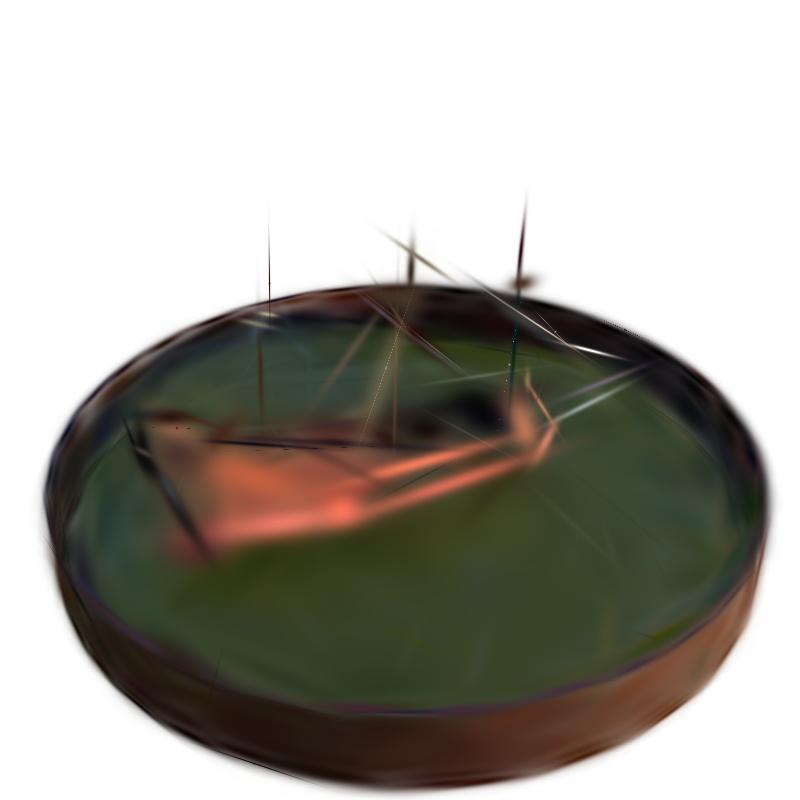}
        & 
        \includegraphics[width=\tmpcolwidth,trim={20px 80px 20px 80px},clip]{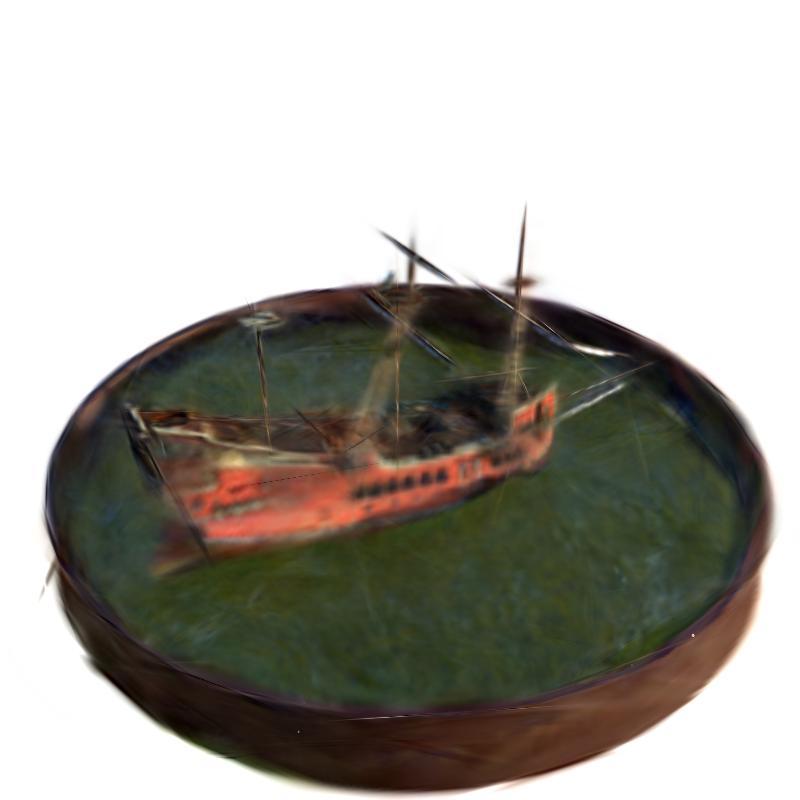}
        &
        &\\
        &
        \includegraphics[width=\tmpcolwidth,trim={20px 80px 20px 80px},clip]{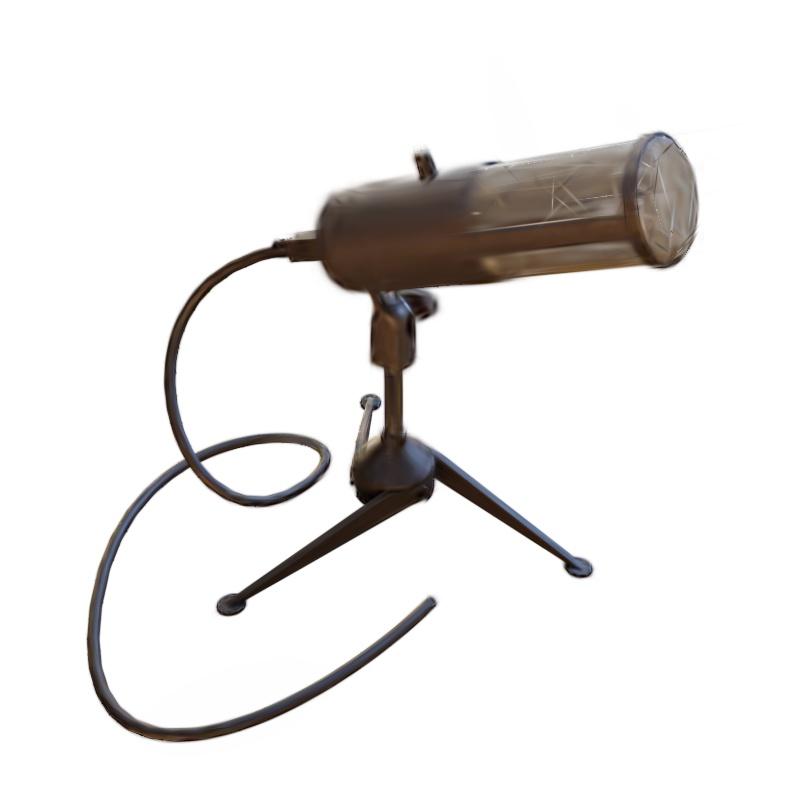}
        & 
        \includegraphics[width=\tmpcolwidth,trim={20px 80px 20px 80px},clip]{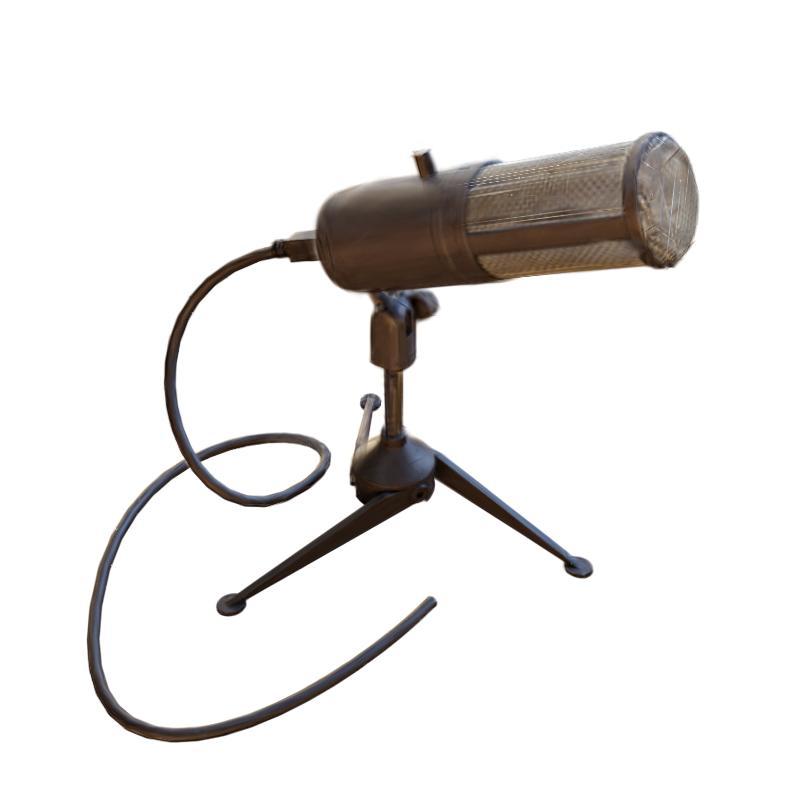}
        &
        \includegraphics[width=\tmpcolwidth,trim={20px 80px 20px 80px},clip]{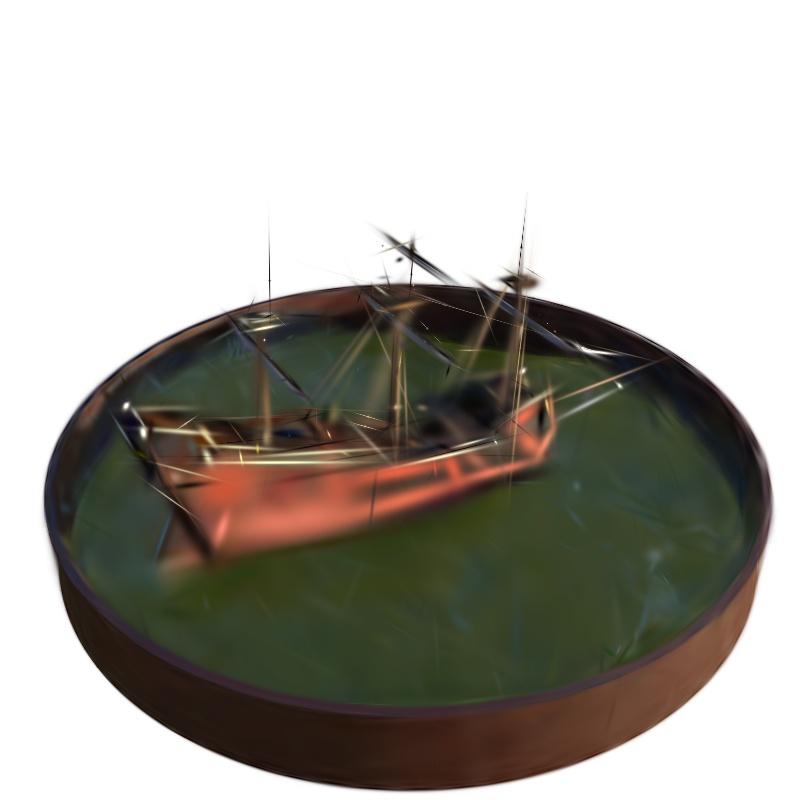}
        & 
        \includegraphics[width=\tmpcolwidth,trim={20px 80px 20px 80px},clip]{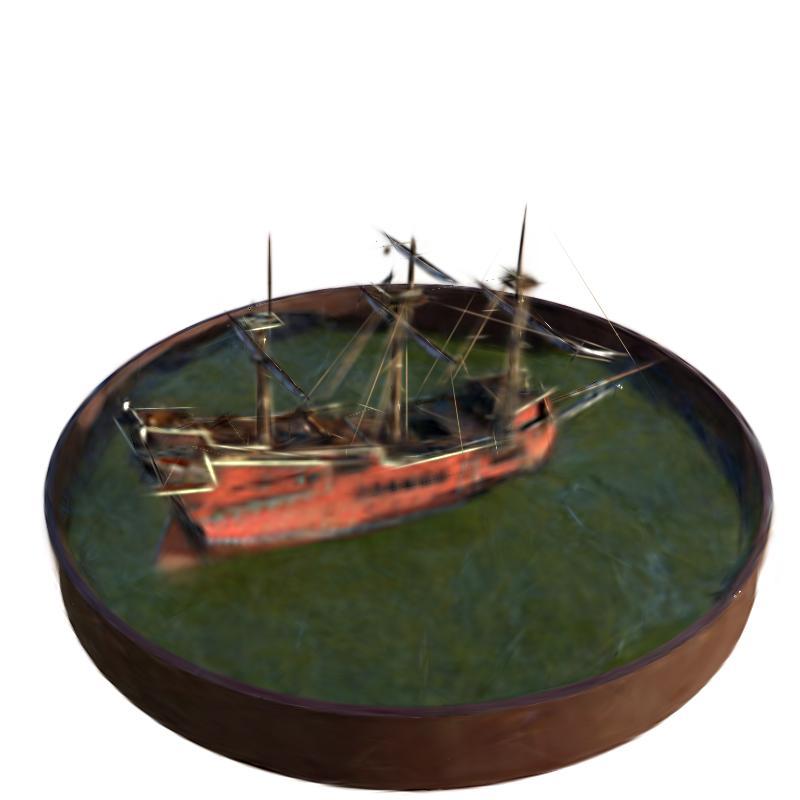}
        &
        &
    \end{tabular}
	\caption{\textbf{Novel view synthesis for discrete levels of detail.} We show test set renders of \ours and 2DGS models in three settings of levels of detail: with 128, 512, and 2048 Gaussians. We show our results on the Blender synthetic scenes not shown in the main paper.\label{fig:lod_supp_blender}}
    \vspace{-1em}
\end{figure*}

\begin{figure*}[!h]
    \newcommand{\vlabel}[1]{\vspace{3pt}#1\vspace{1pt}}
    \setlength\tmpcolwidth{1.0in}
	\setlength{\tabcolsep}{0pt}
    \renewcommand{\arraystretch}{0}
	\centering
	\begin{tabular}{M{0.2in}M{\tmpcolwidth}M{\tmpcolwidth}M{\tmpcolwidth}M{\tmpcolwidth}M{\tmpcolwidth}M{\tmpcolwidth}}
        & \vlabel{2DGS} & \vlabel{\ours} & \vlabel{2DGS} & \vlabel{\ours} & \vlabel{2DGS} & \vlabel{\ours}\\
        \multirow{3}{*}{\rotatebox{90}{$\xleftarrow{\makebox[1.1in]{Number of Gaussians}}$}}
        &
        \includegraphics[width=\tmpcolwidth,trim={40px 160px 40px 0px},clip]{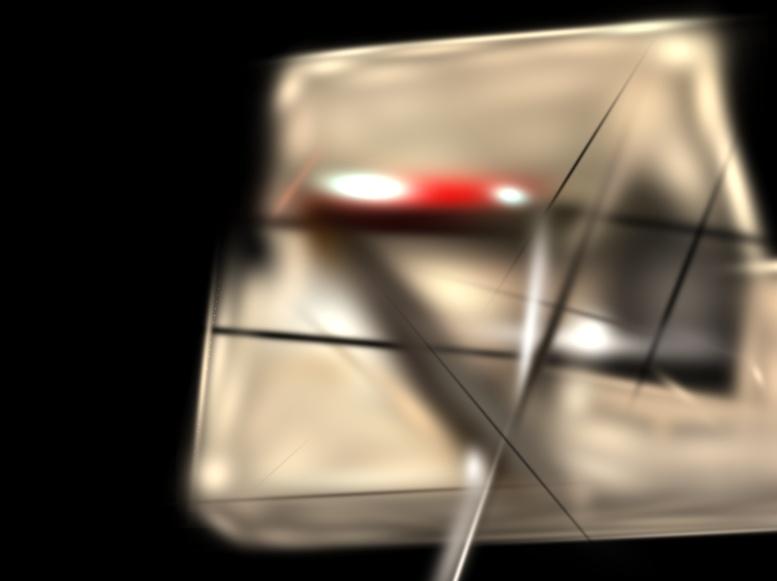}
        & \includegraphics[width=\tmpcolwidth,trim={40px 160px 40px 0px},clip]{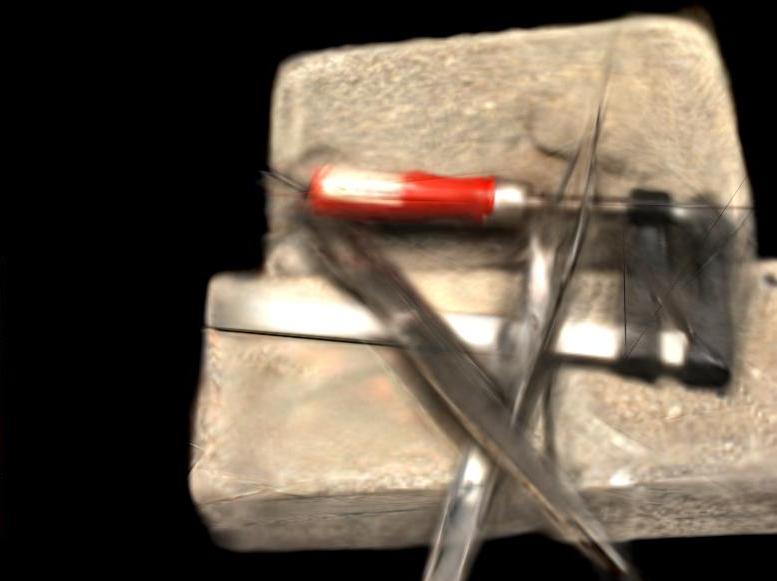}
        &
        \includegraphics[width=\tmpcolwidth,trim={40px 160px 40px 0px},clip]{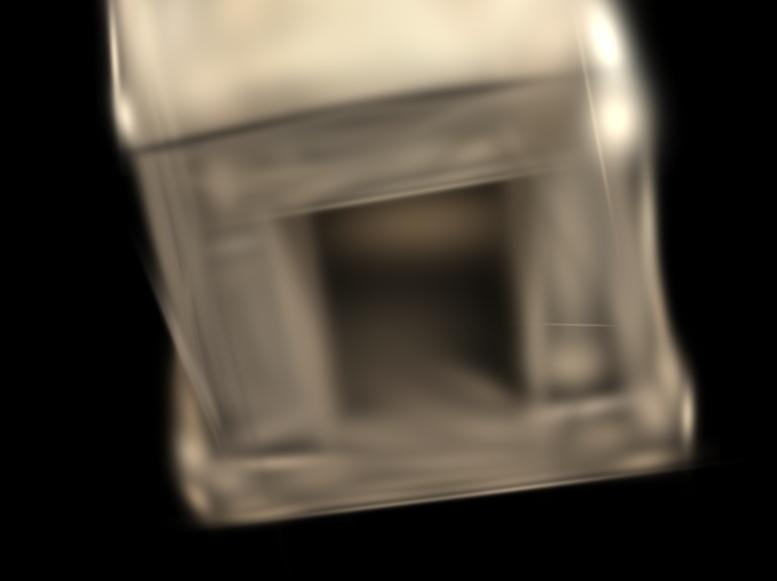}
        & \includegraphics[width=\tmpcolwidth,trim={40px 160px 40px 0px},clip]{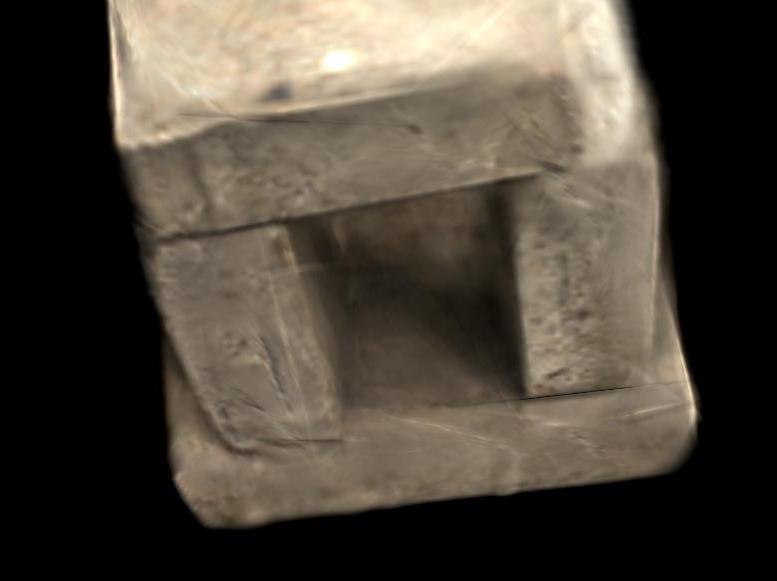}
        &
        \includegraphics[width=\tmpcolwidth,trim={40px 160px 40px 0px},clip]{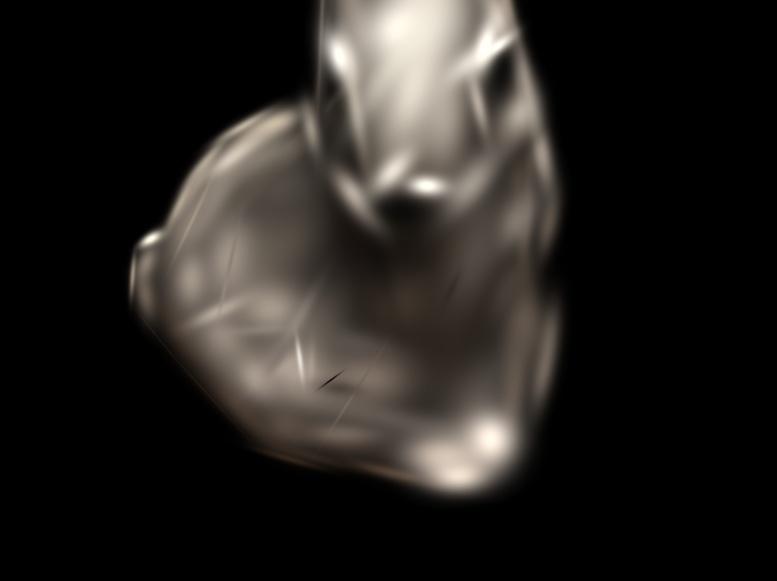}
        & \includegraphics[width=\tmpcolwidth,trim={40px 160px 40px 0px},clip]{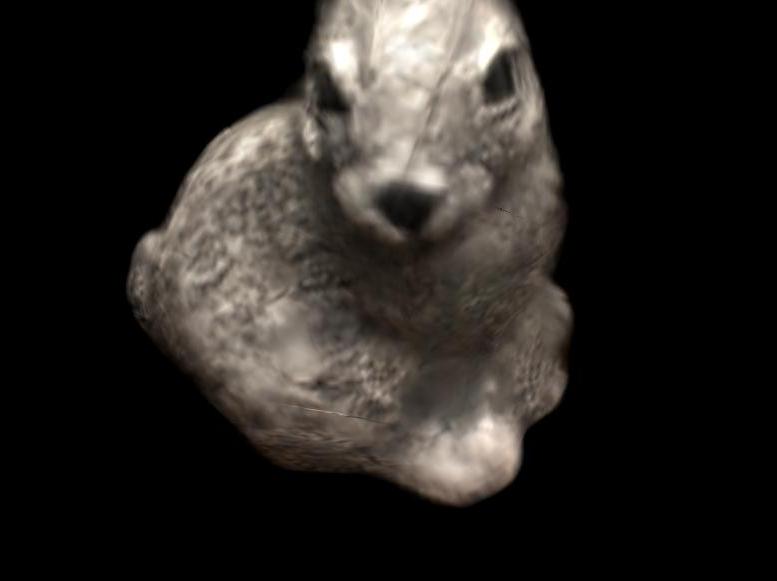}\\
                &
        \includegraphics[width=\tmpcolwidth,trim={40px 160px 40px 0px},clip]{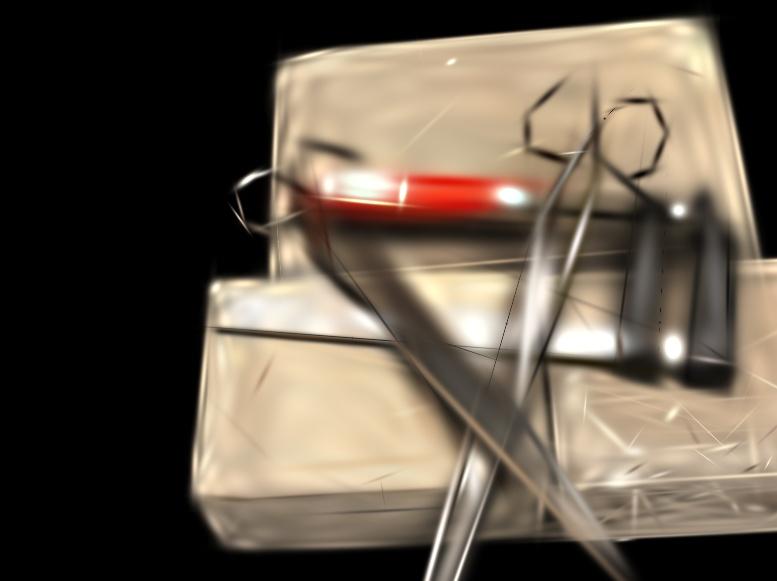}
        & \includegraphics[width=\tmpcolwidth,trim={40px 160px 40px 0px},clip]{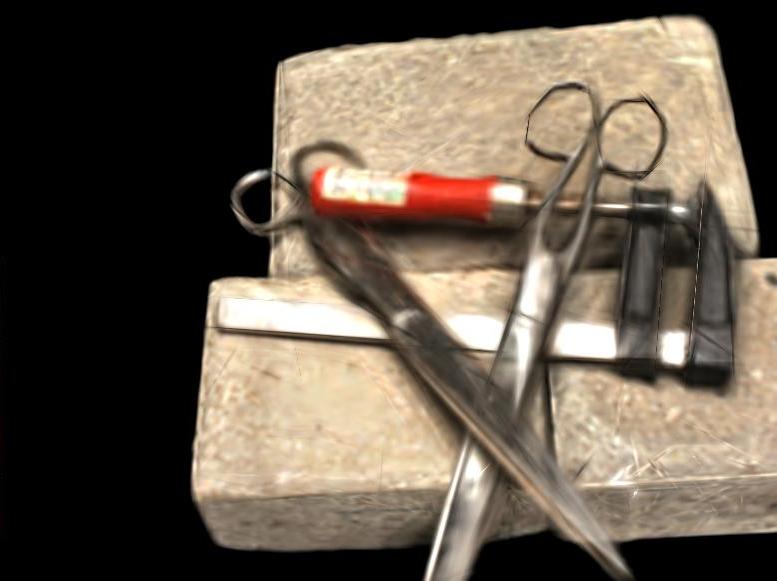}
        &
        \includegraphics[width=\tmpcolwidth,trim={40px 160px 40px 0px},clip]{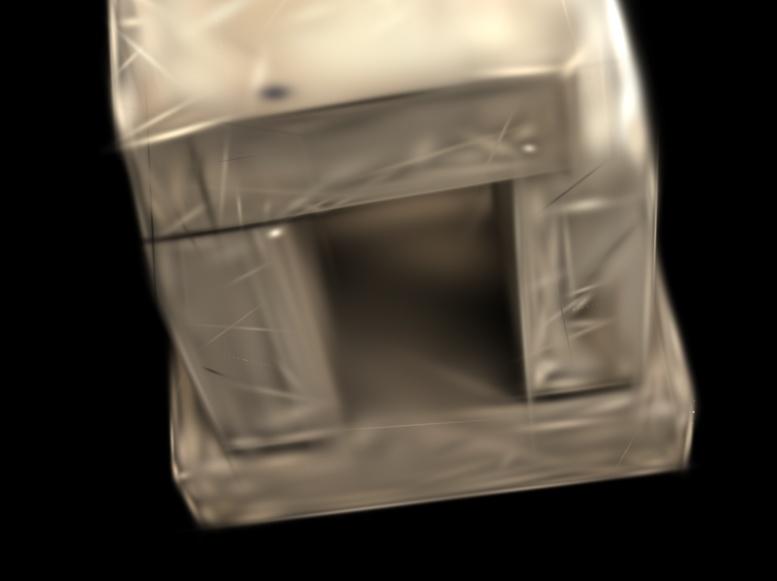}
        & \includegraphics[width=\tmpcolwidth,trim={40px 160px 40px 0px},clip]{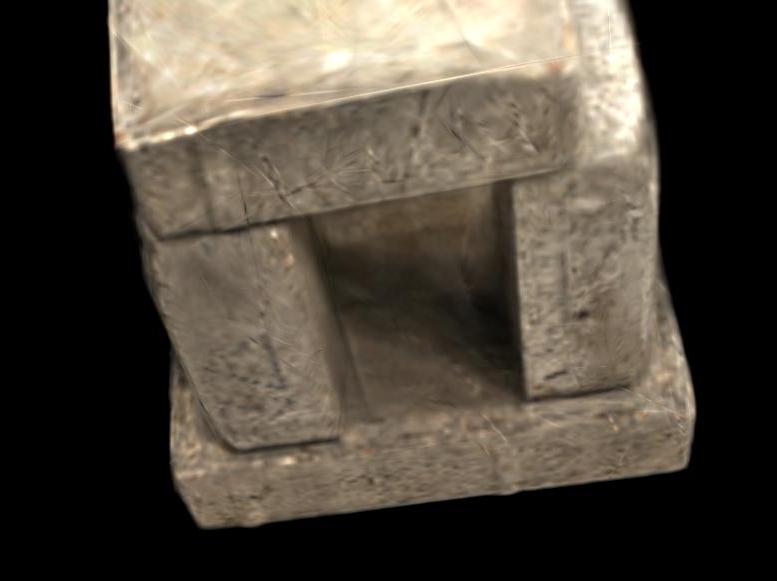}
        &
        \includegraphics[width=\tmpcolwidth,trim={40px 160px 40px 0px},clip]{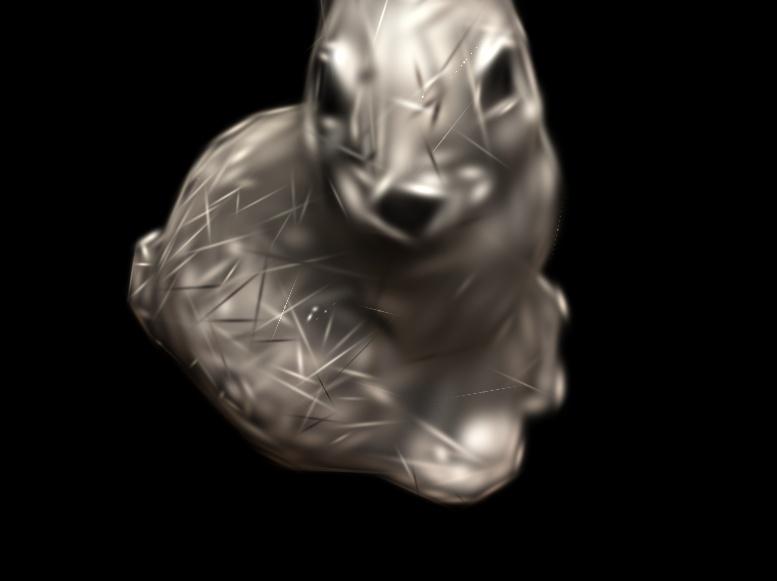}
        & \includegraphics[width=\tmpcolwidth,trim={40px 160px 40px 0px},clip]{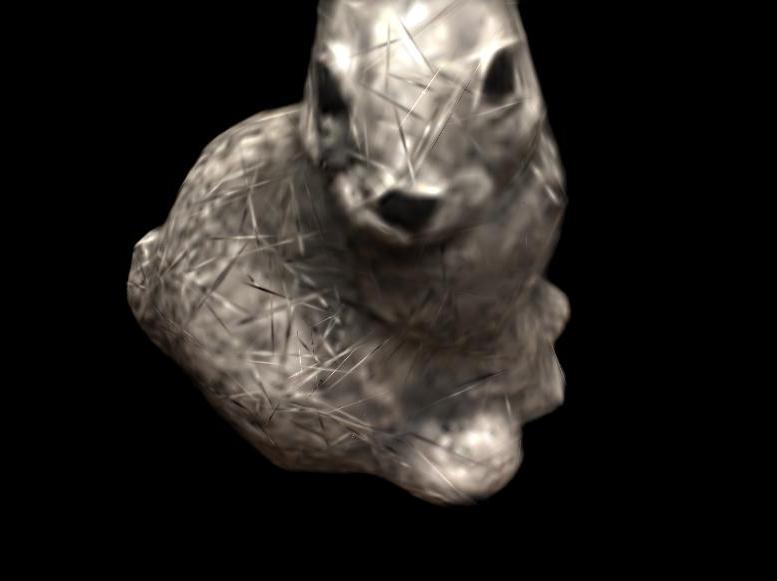}\\
                &
        \includegraphics[width=\tmpcolwidth,trim={40px 160px 40px 0px},clip]{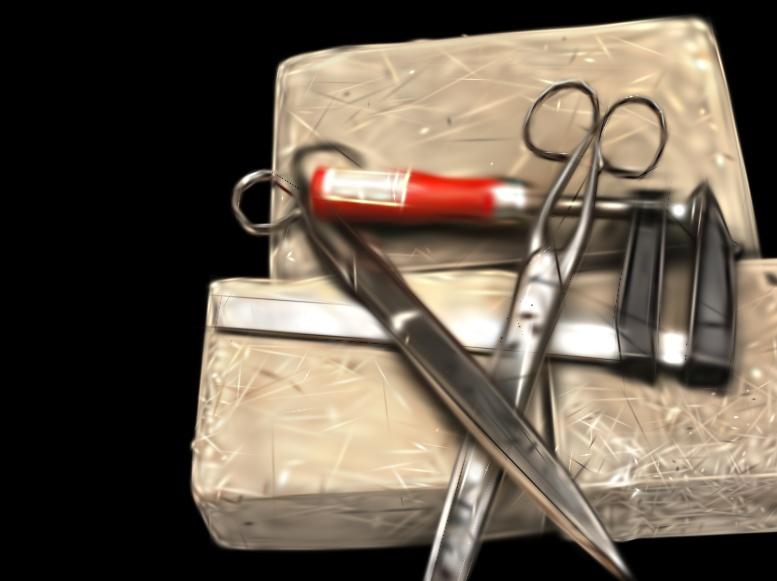}
        & \includegraphics[width=\tmpcolwidth,trim={40px 160px 40px 0px},clip]{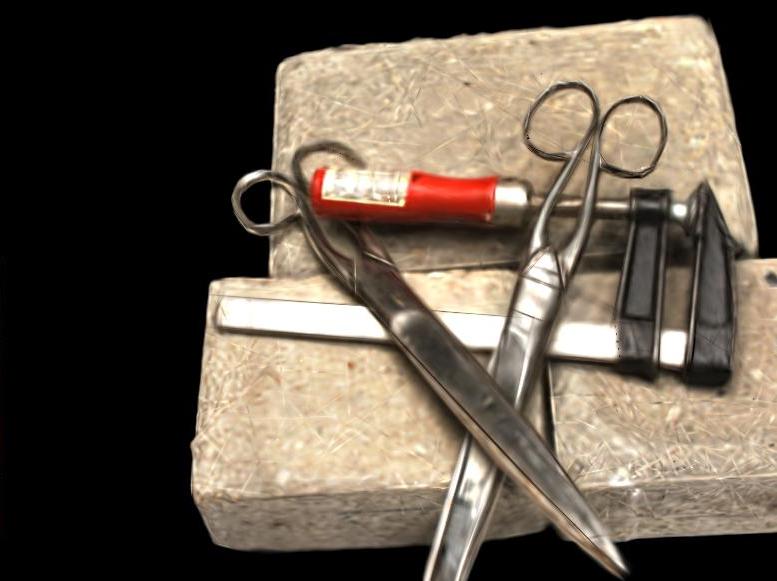}
        &
        \includegraphics[width=\tmpcolwidth,trim={40px 160px 40px 0px},clip]{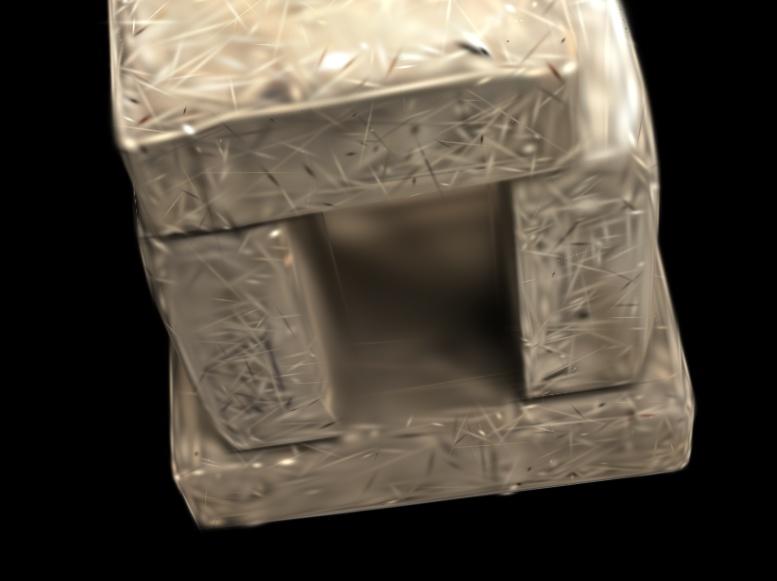}
        & \includegraphics[width=\tmpcolwidth,trim={40px 160px 40px 0px},clip]{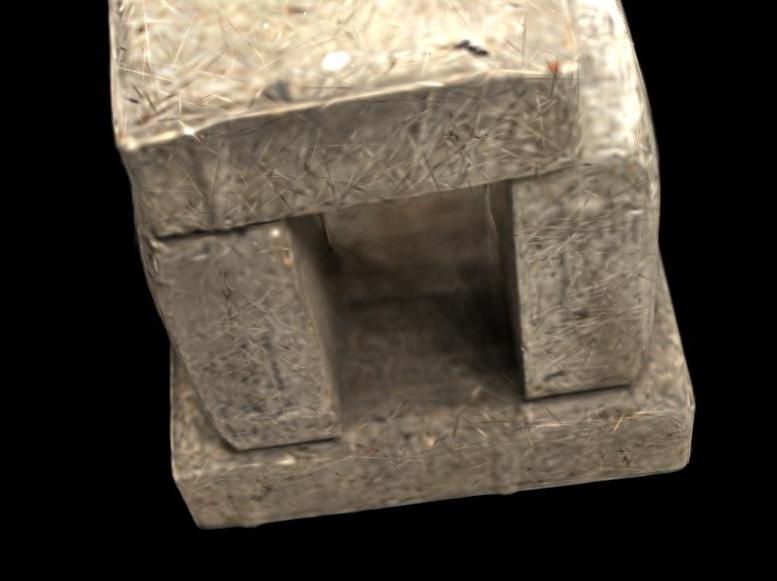}
        &
        \includegraphics[width=\tmpcolwidth,trim={40px 160px 40px 0px},clip]{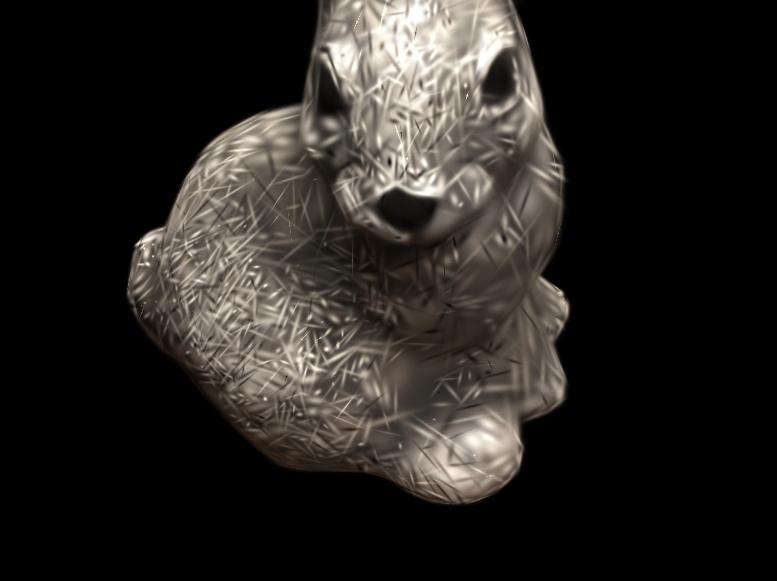}
        & \includegraphics[width=\tmpcolwidth,trim={40px 160px 40px 0px},clip]{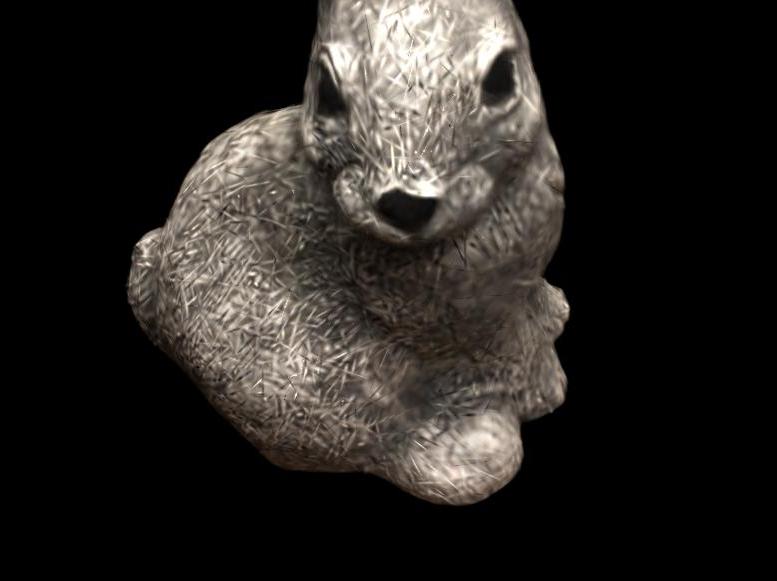}\\
        & \vlabel{2DGS} & \vlabel{\ours} & \vlabel{2DGS} & \vlabel{\ours} & \vlabel{2DGS} & \vlabel{\ours}\\
        \multirow{3}{*}{\rotatebox{90}{$\xleftarrow{\makebox[1.1in]{Number of Gaussians}}$}}
        &
        \includegraphics[width=\tmpcolwidth,trim={40px 160px 40px 0px},clip]{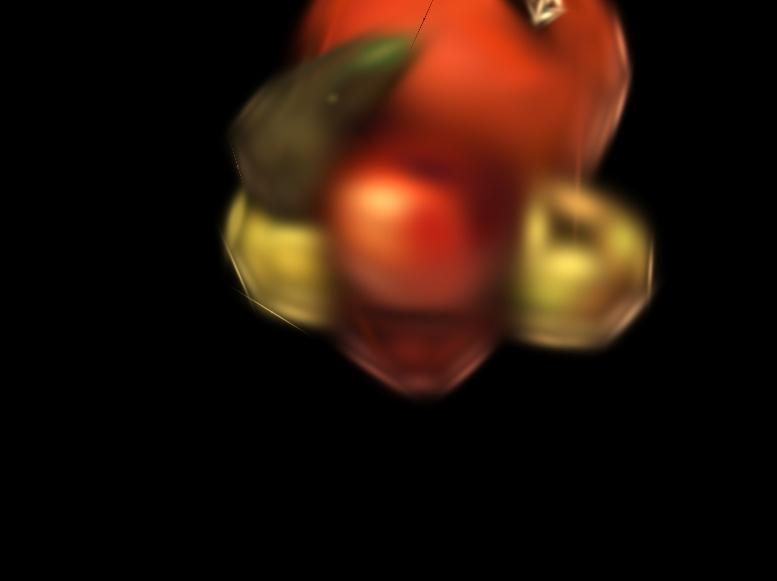}
        & \includegraphics[width=\tmpcolwidth,trim={40px 160px 40px 0px},clip]{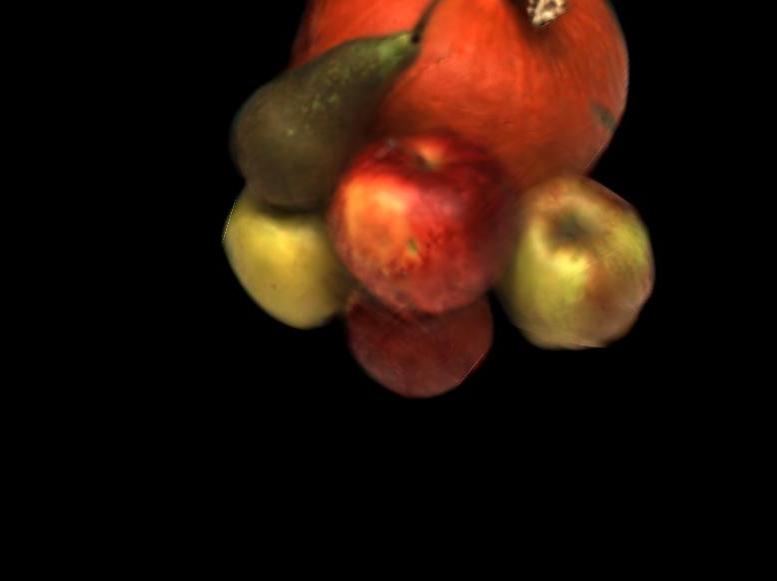}
        &
        \includegraphics[width=\tmpcolwidth,trim={40px 160px 40px 0px},clip]{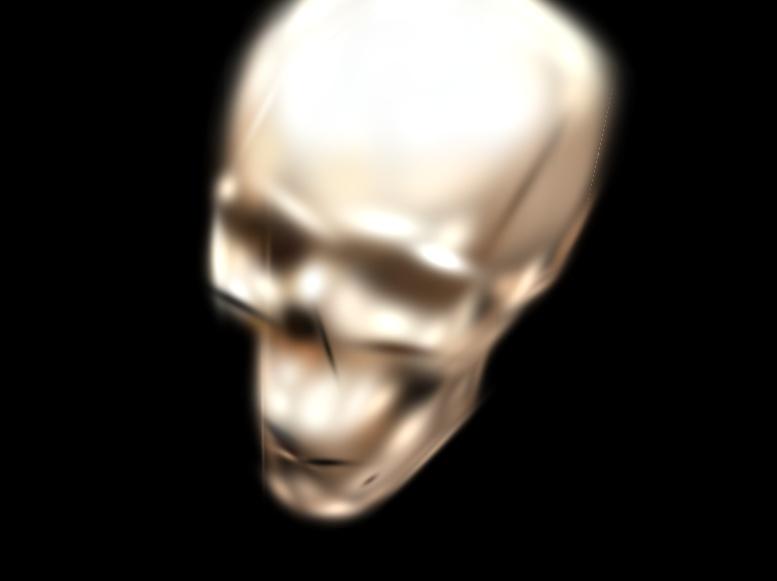}
        & \includegraphics[width=\tmpcolwidth,trim={40px 160px 40px 0px},clip]{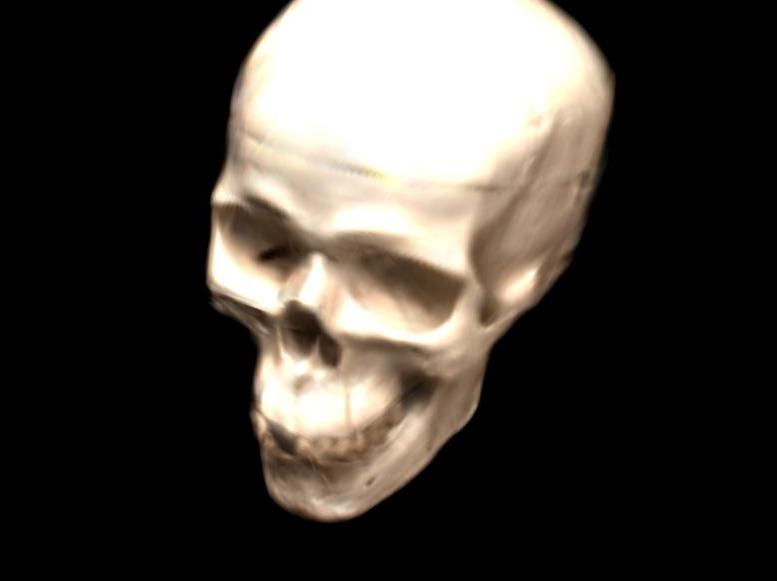}
        &
        \includegraphics[width=\tmpcolwidth,trim={40px 160px 40px 0px},clip]{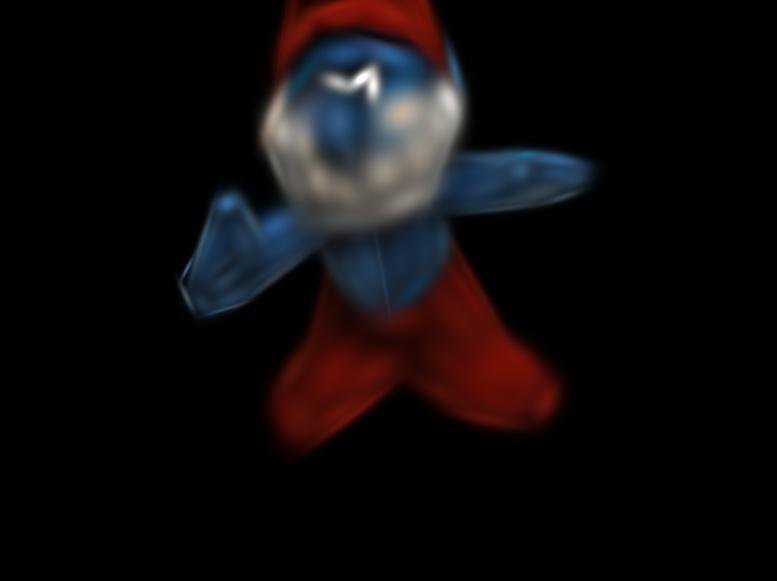}
        & \includegraphics[width=\tmpcolwidth,trim={40px 160px 40px 0px},clip]{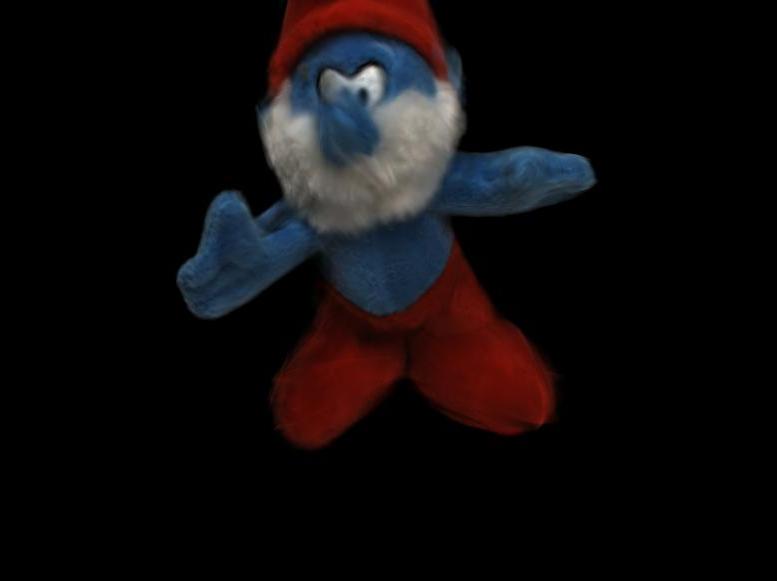}\\
                &
        \includegraphics[width=\tmpcolwidth,trim={40px 160px 40px 0px},clip]{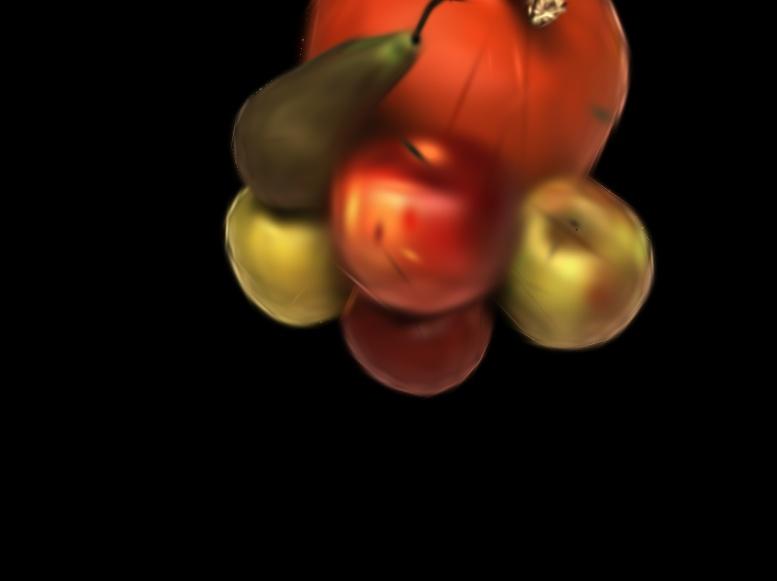}
        & \includegraphics[width=\tmpcolwidth,trim={40px 160px 40px 0px},clip]{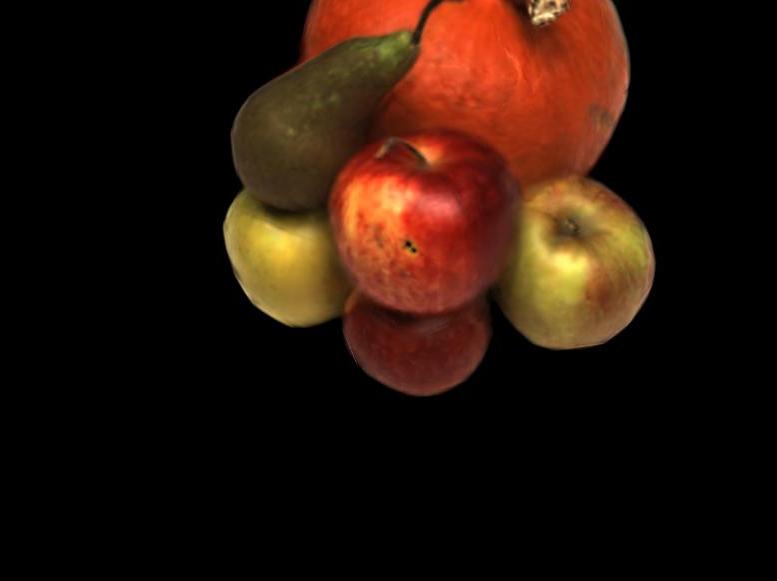}
        &
        \includegraphics[width=\tmpcolwidth,trim={40px 160px 40px 0px},clip]{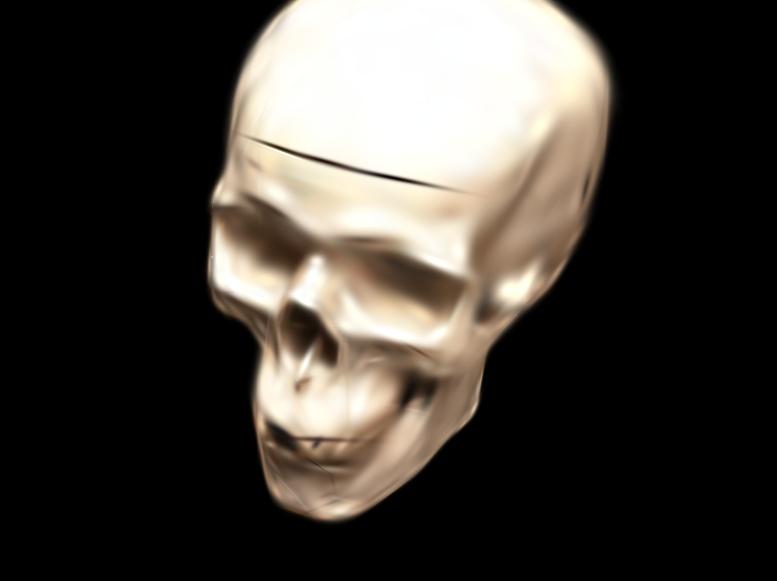}
        & \includegraphics[width=\tmpcolwidth,trim={40px 160px 40px 0px},clip]{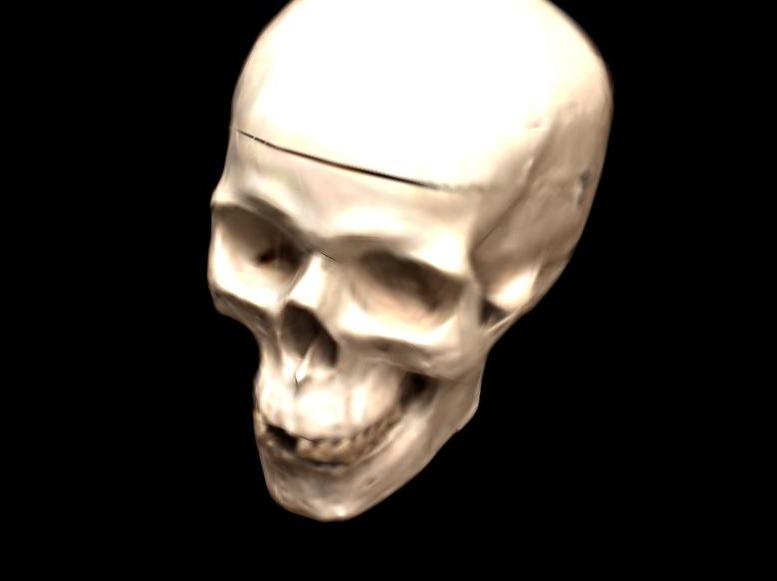}
        &
        \includegraphics[width=\tmpcolwidth,trim={40px 160px 40px 0px},clip]{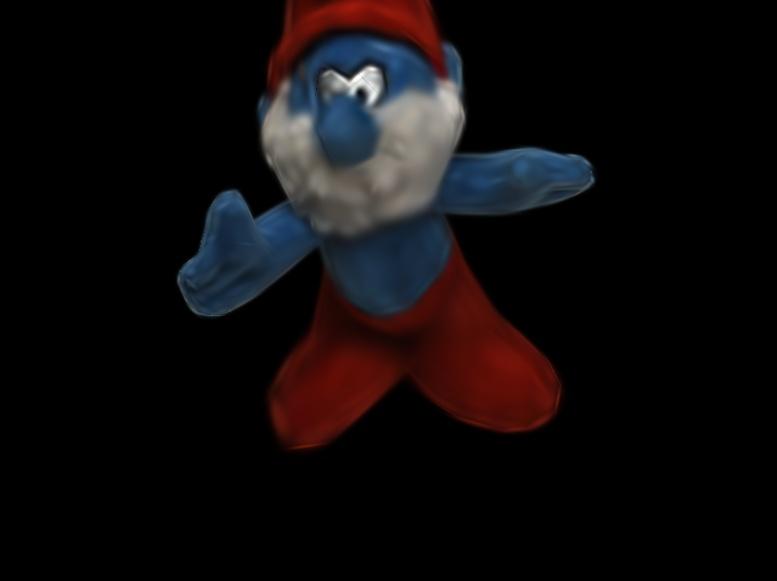}
        & \includegraphics[width=\tmpcolwidth,trim={40px 160px 40px 0px},clip]{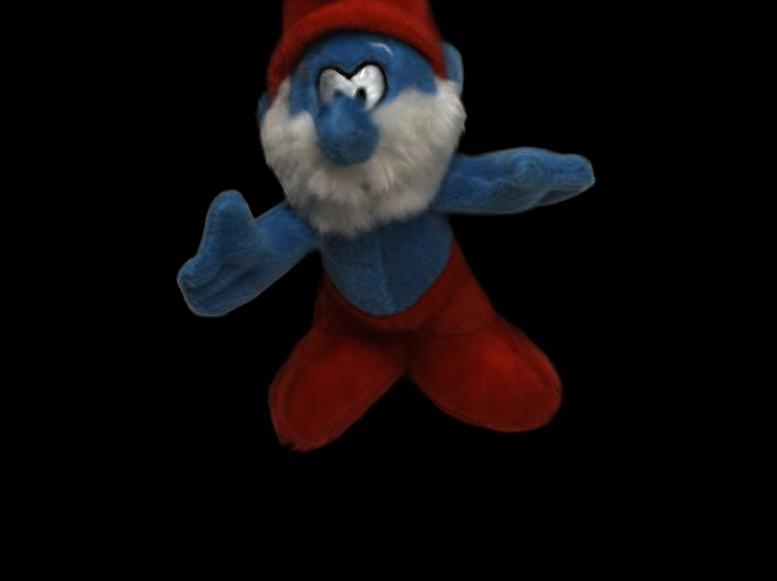}\\
                &
        \includegraphics[width=\tmpcolwidth,trim={40px 160px 40px 0px},clip]{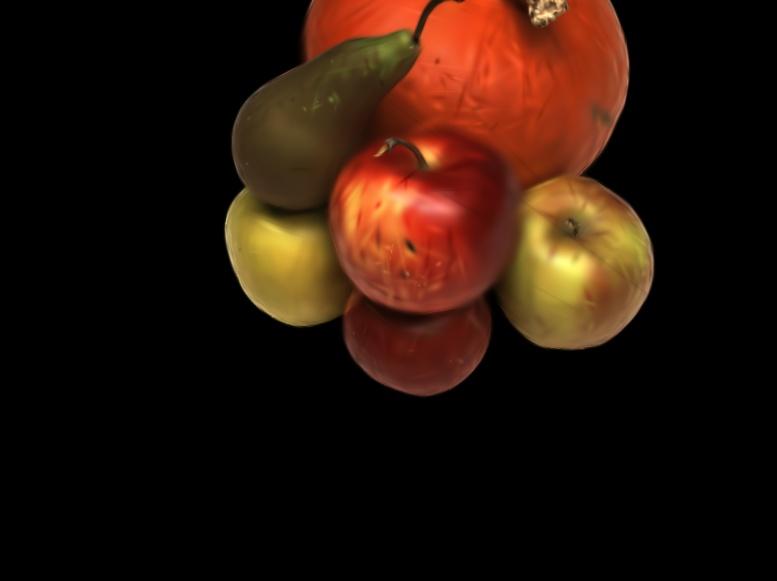}
        & \includegraphics[width=\tmpcolwidth,trim={40px 160px 40px 0px},clip]{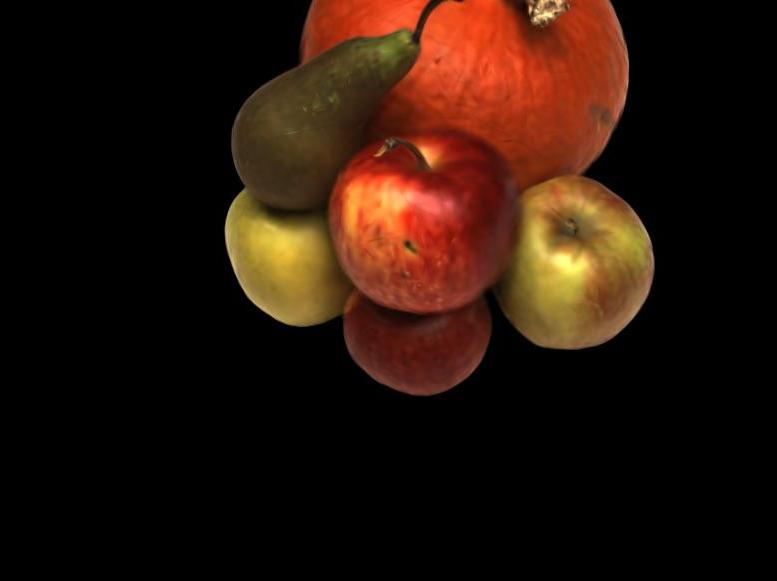}
        &
        \includegraphics[width=\tmpcolwidth,trim={40px 160px 40px 0px},clip]{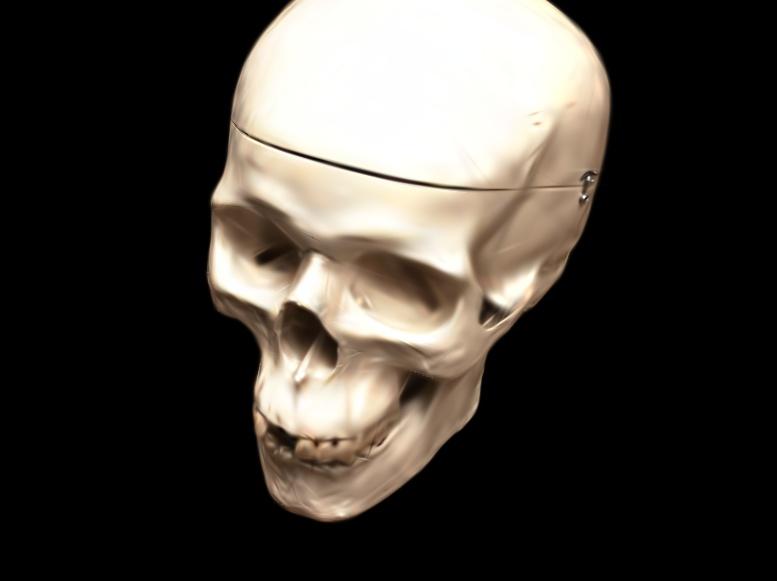}
        & \includegraphics[width=\tmpcolwidth,trim={40px 160px 40px 0px},clip]{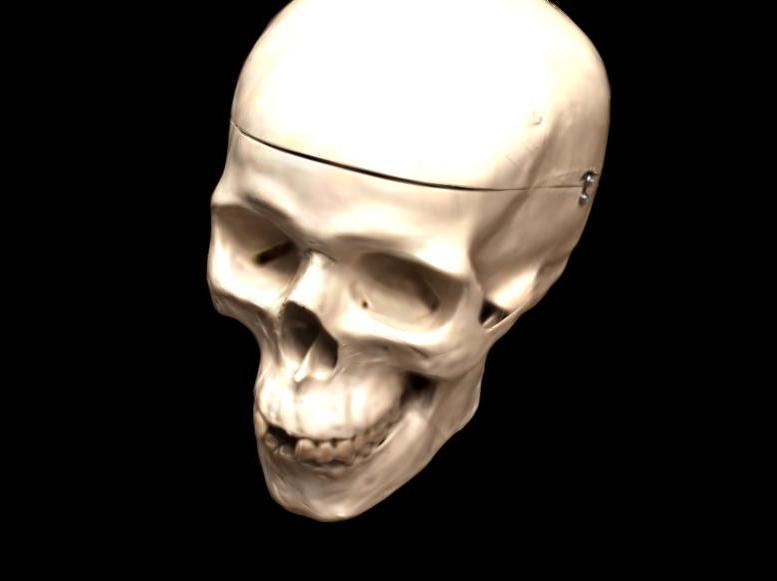}
        &
        \includegraphics[width=\tmpcolwidth,trim={40px 160px 40px 0px},clip]{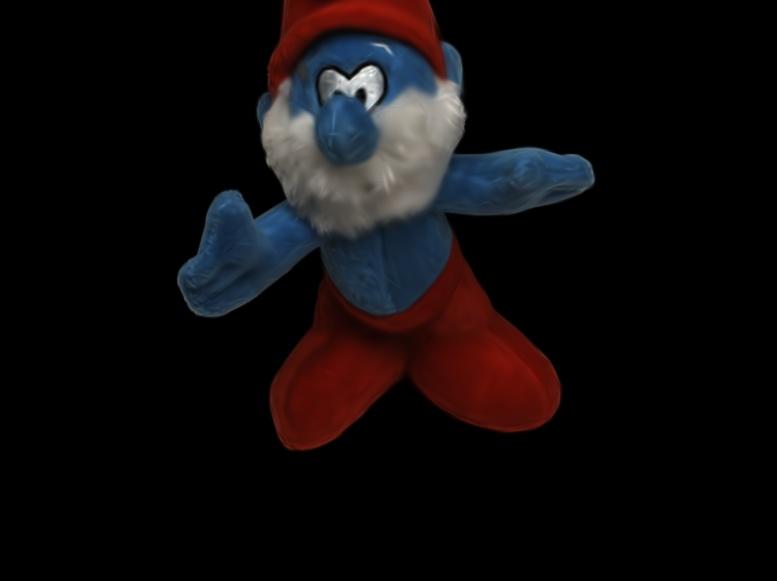}
        & \includegraphics[width=\tmpcolwidth,trim={40px 160px 40px 0px},clip]{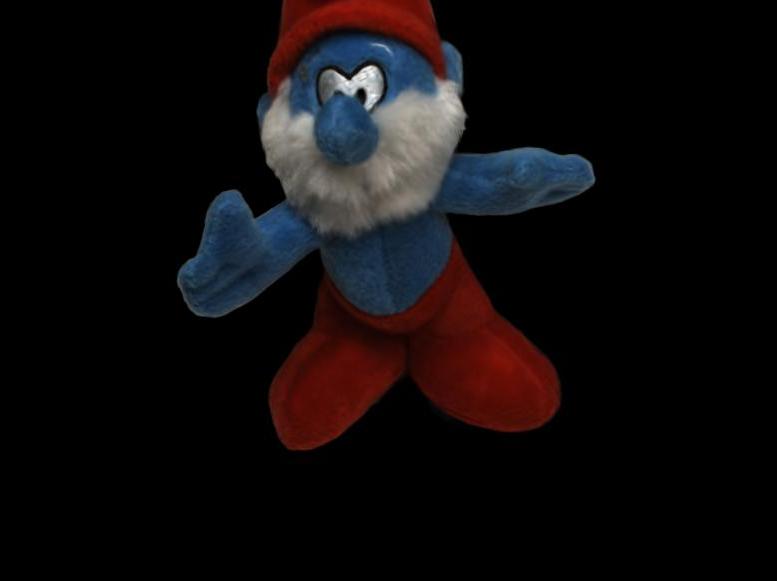}\\
        & \vlabel{2DGS} & \vlabel{\ours} & \vlabel{2DGS} & \vlabel{\ours} & \vlabel{2DGS} & \vlabel{\ours}\\
        \multirow{3}{*}{\rotatebox{90}{$\xleftarrow{\makebox[1.1in]{Number of Gaussians}}$}}
        &
        \includegraphics[width=\tmpcolwidth,trim={40px 160px 40px 0px},clip]{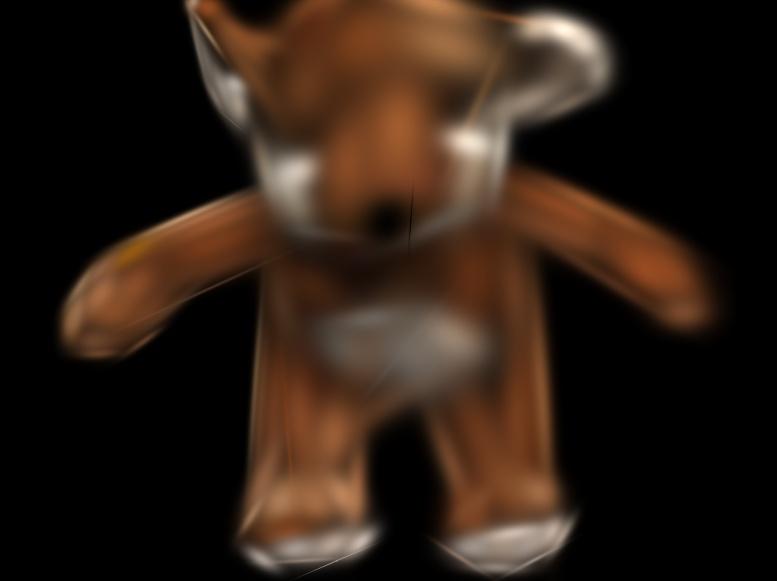}
        & \includegraphics[width=\tmpcolwidth,trim={40px 160px 40px 0px},clip]{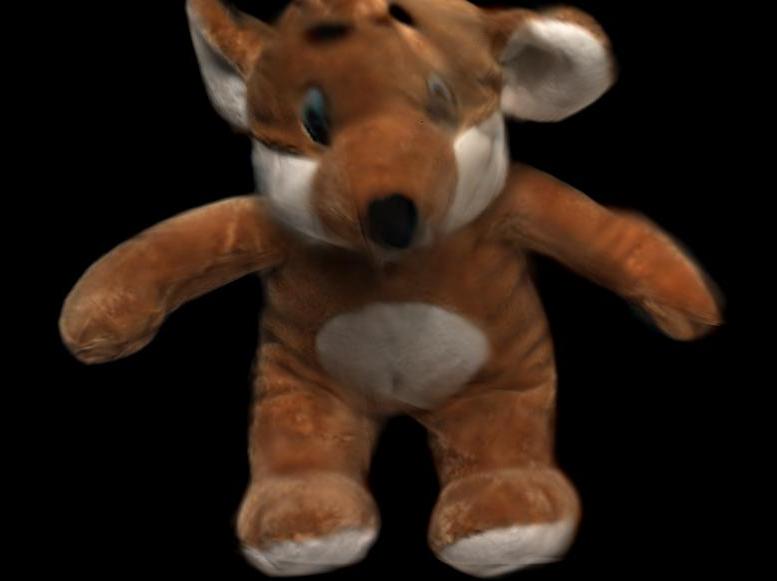}
        &
        \includegraphics[width=\tmpcolwidth,trim={40px 160px 40px 0px},clip]{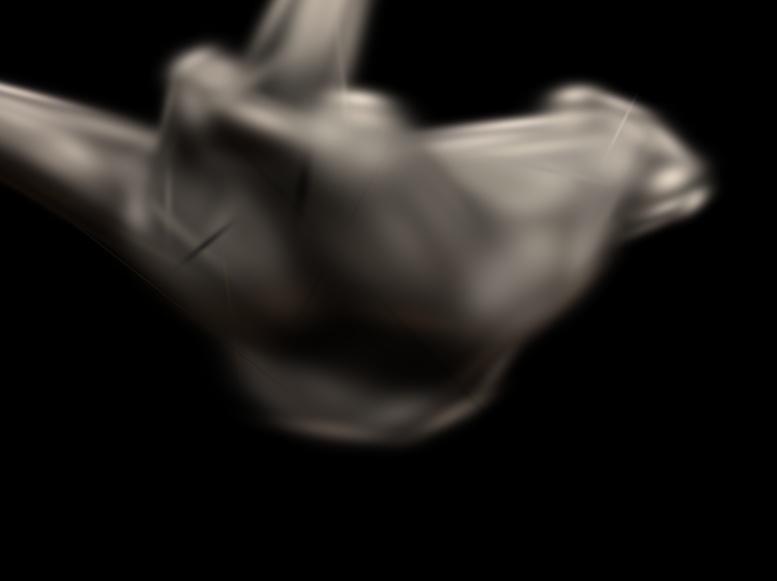}
        & \includegraphics[width=\tmpcolwidth,trim={40px 160px 40px 0px},clip]{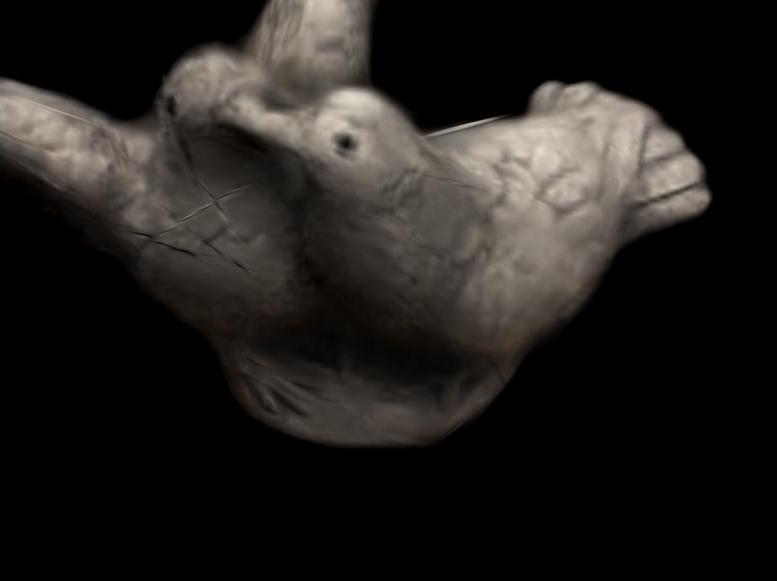}
        &
        \includegraphics[width=\tmpcolwidth,trim={40px 160px 40px 0px},clip]{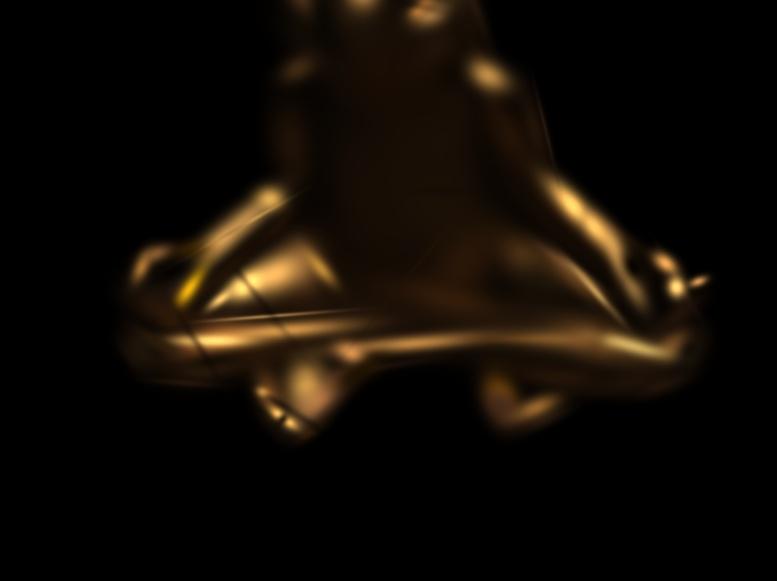}
        & \includegraphics[width=\tmpcolwidth,trim={40px 160px 40px 0px},clip]{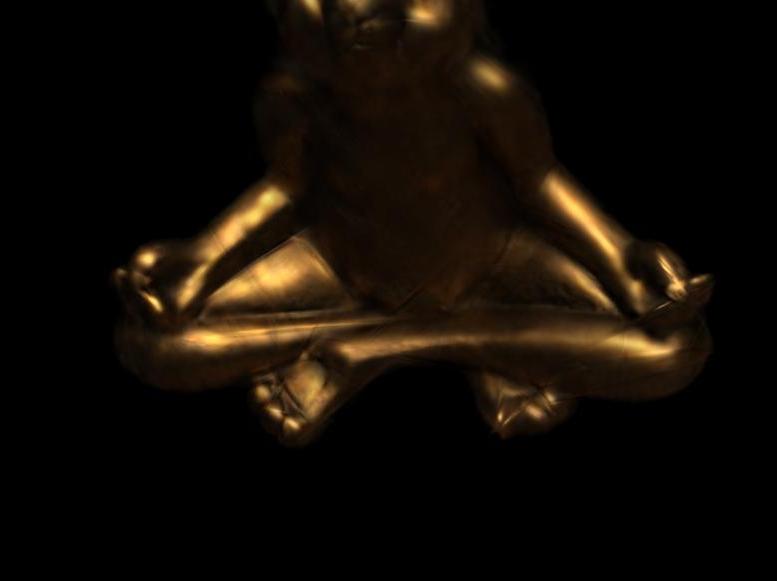}\\
                &
        \includegraphics[width=\tmpcolwidth,trim={40px 160px 40px 0px},clip]{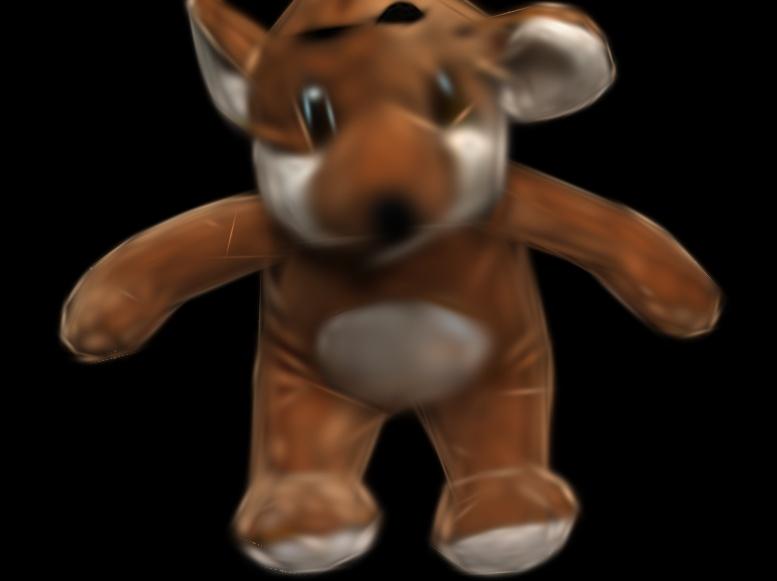}
        & \includegraphics[width=\tmpcolwidth,trim={40px 160px 40px 0px},clip]{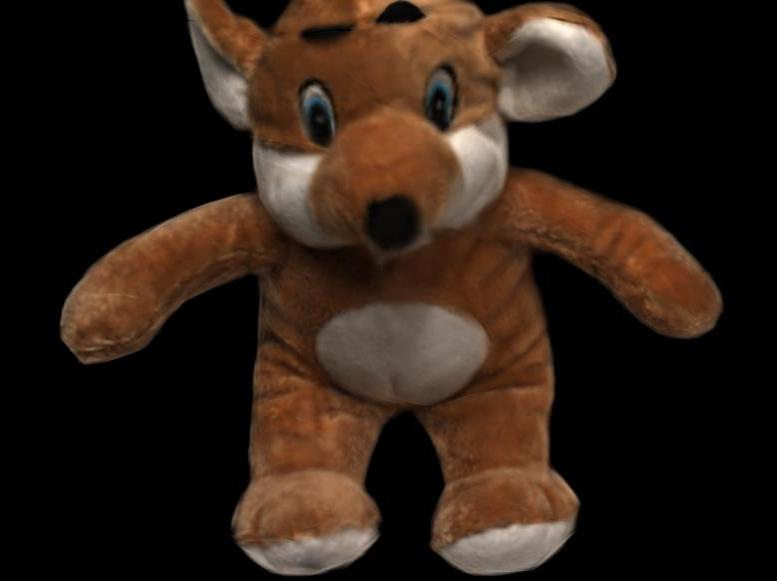}
        &
        \includegraphics[width=\tmpcolwidth,trim={40px 160px 40px 0px},clip]{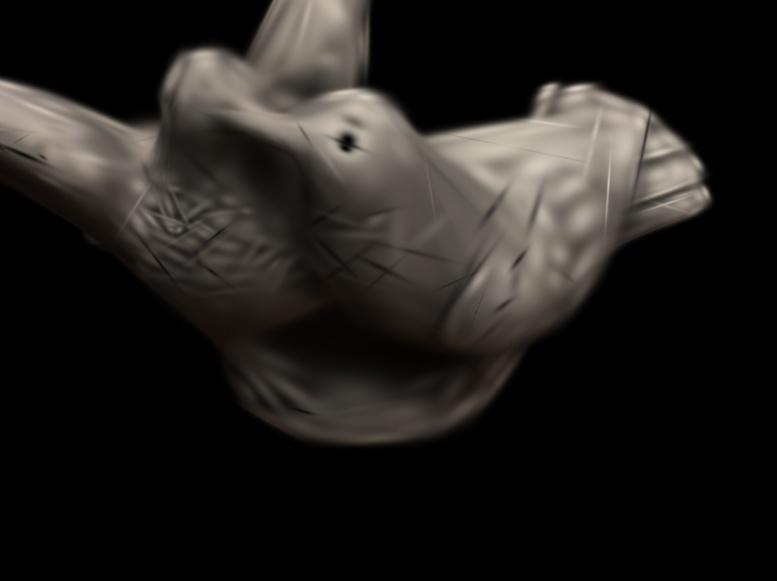}
        & \includegraphics[width=\tmpcolwidth,trim={40px 160px 40px 0px},clip]{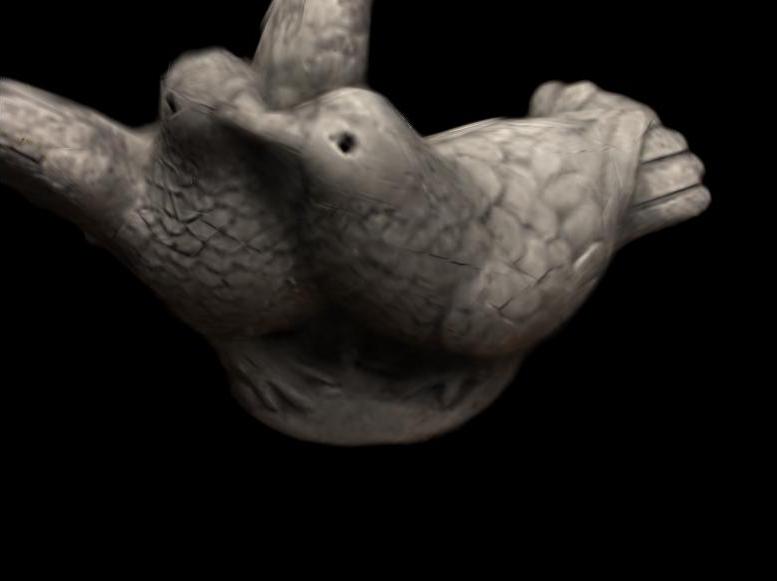}
        &
        \includegraphics[width=\tmpcolwidth,trim={40px 160px 40px 0px},clip]{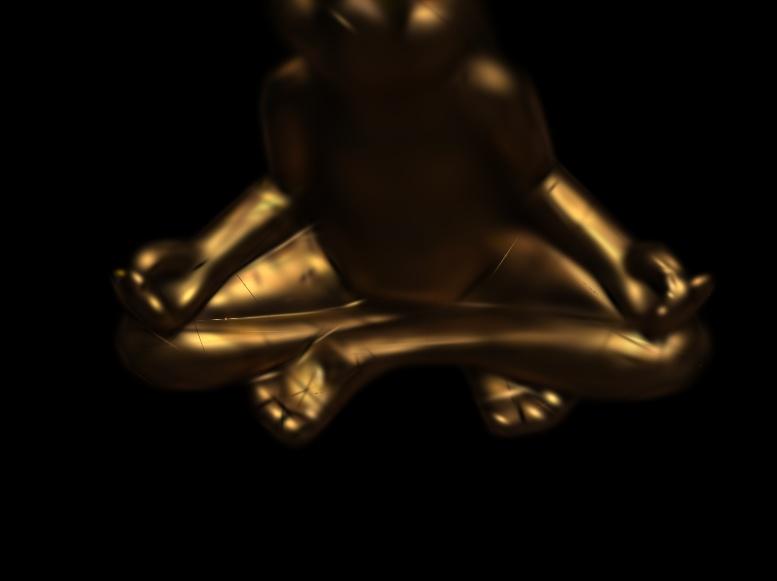}
        & \includegraphics[width=\tmpcolwidth,trim={40px 160px 40px 0px},clip]{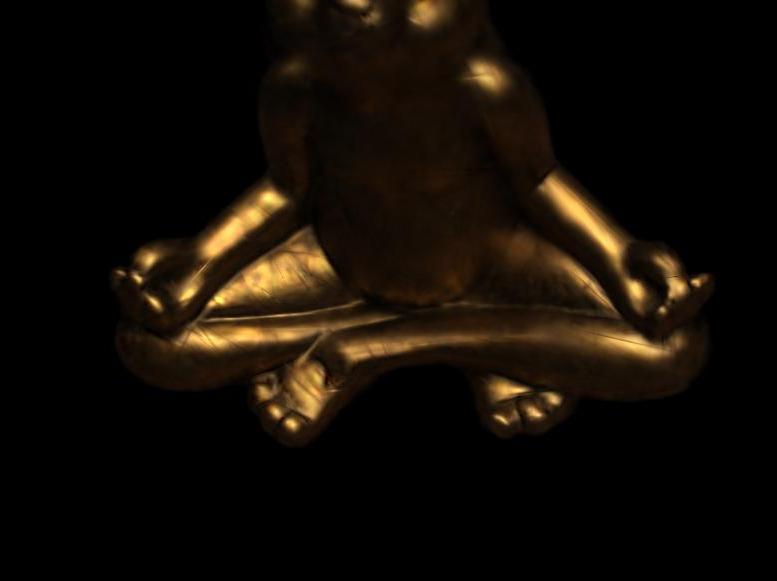}\\
                &
        \includegraphics[width=\tmpcolwidth,trim={40px 160px 40px 0px},clip]{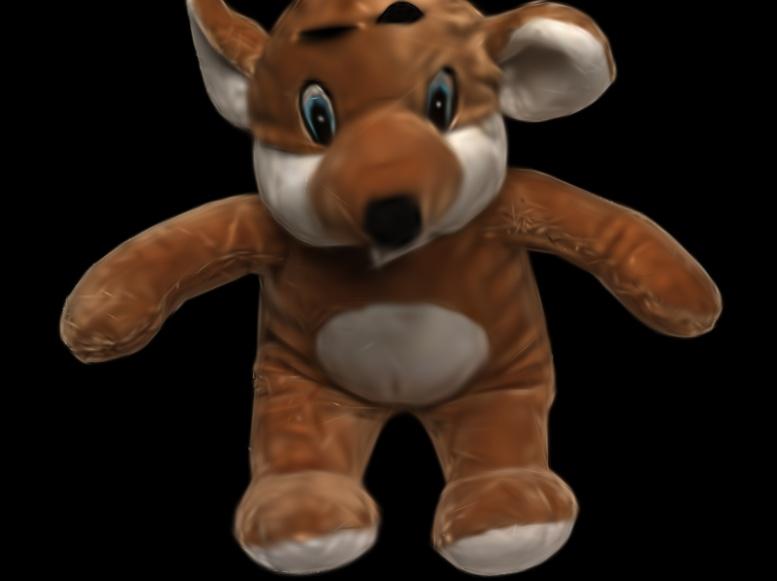}
        & \includegraphics[width=\tmpcolwidth,trim={40px 160px 40px 0px},clip]{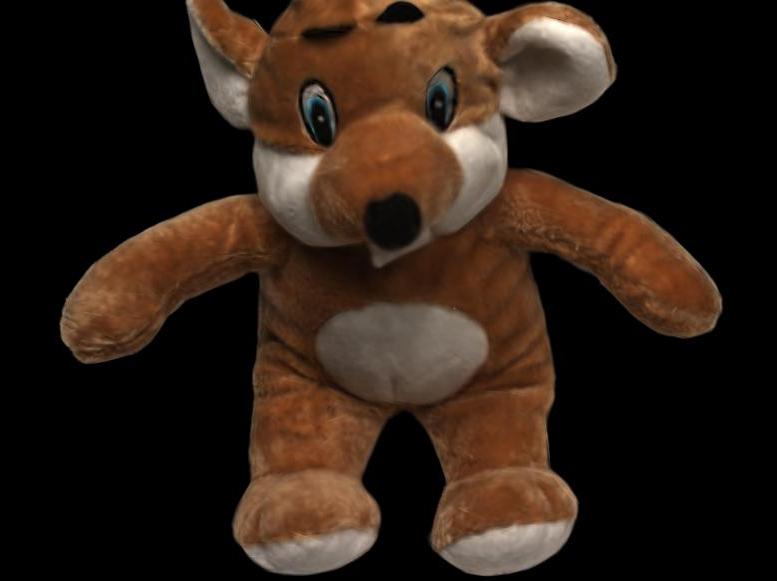}
        &
        \includegraphics[width=\tmpcolwidth,trim={40px 160px 40px 0px},clip]{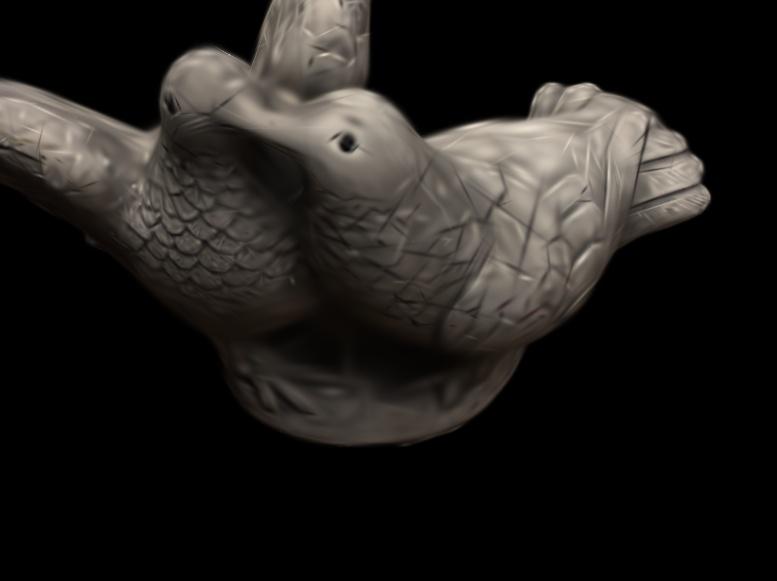}
        & \includegraphics[width=\tmpcolwidth,trim={40px 160px 40px 0px},clip]{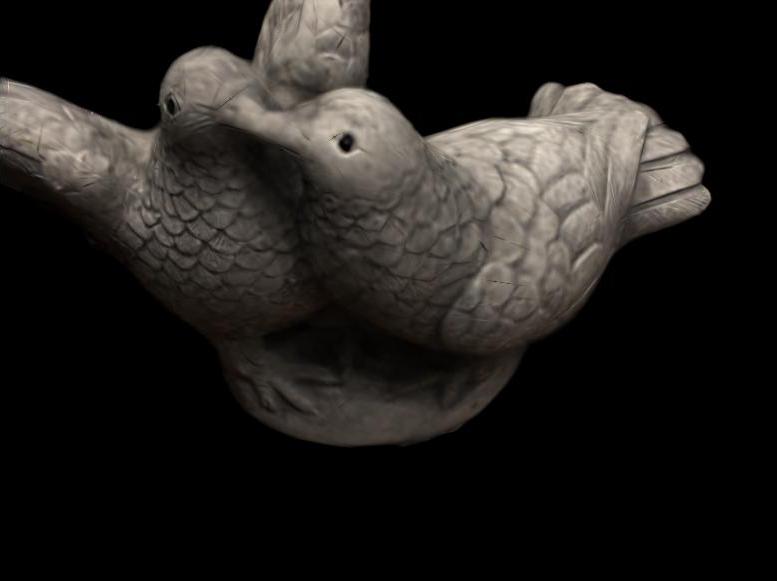}
        &
        \includegraphics[width=\tmpcolwidth,trim={40px 160px 40px 0px},clip]{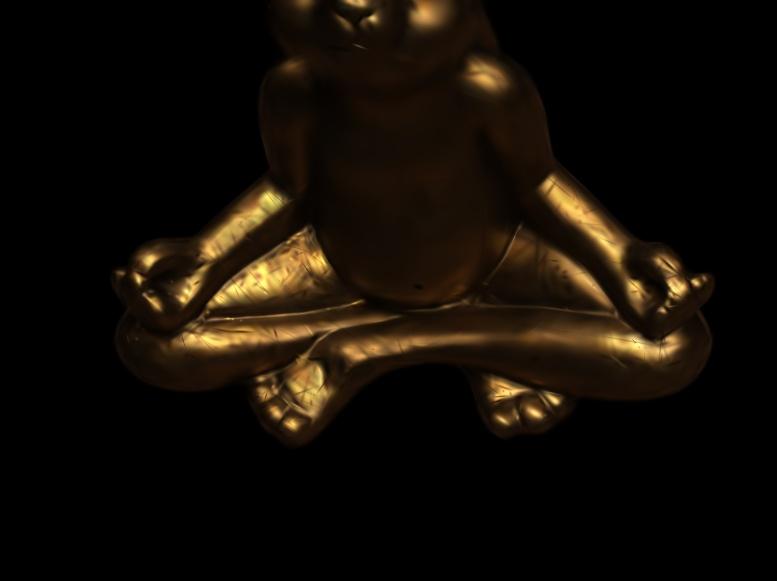}
        & \includegraphics[width=\tmpcolwidth,trim={40px 160px 40px 0px},clip]{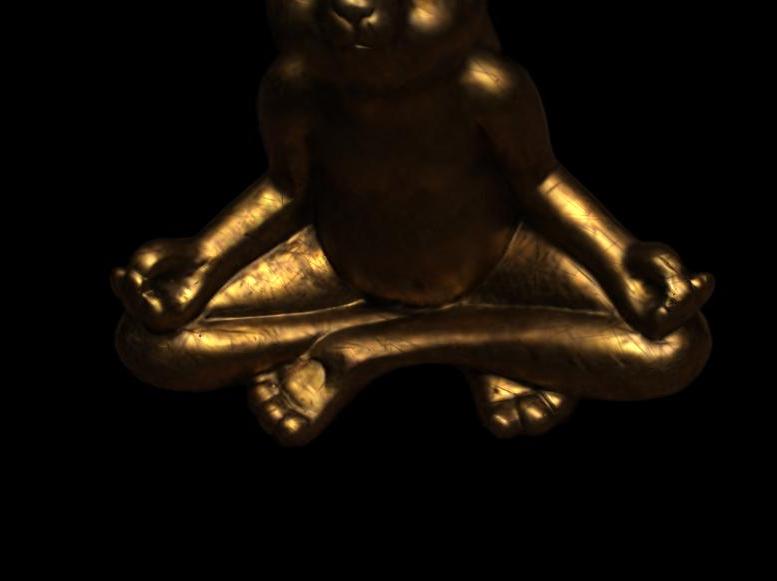}\\
        & \vlabel{2DGS} & \vlabel{\ours} & \vlabel{2DGS} & \vlabel{\ours} & \vlabel{2DGS} & \vlabel{\ours}\\
        \multirow{3}{*}{\rotatebox{90}{$\xleftarrow{\makebox[1.1in]{Number of Gaussians}}$}}
        &
        \includegraphics[width=\tmpcolwidth,trim={40px 160px 40px 0px},clip]{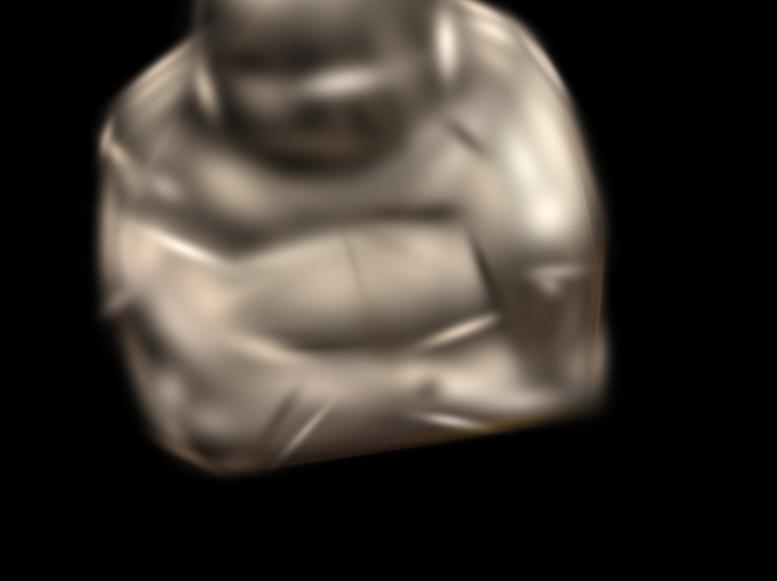}
        & \includegraphics[width=\tmpcolwidth,trim={40px 160px 40px 0px},clip]{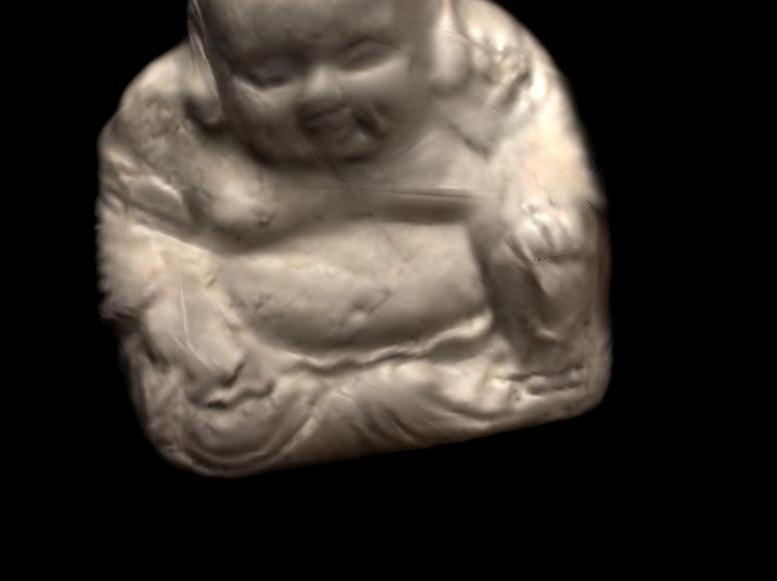}
        &
        \includegraphics[width=\tmpcolwidth,trim={40px 160px 40px 0px},clip]{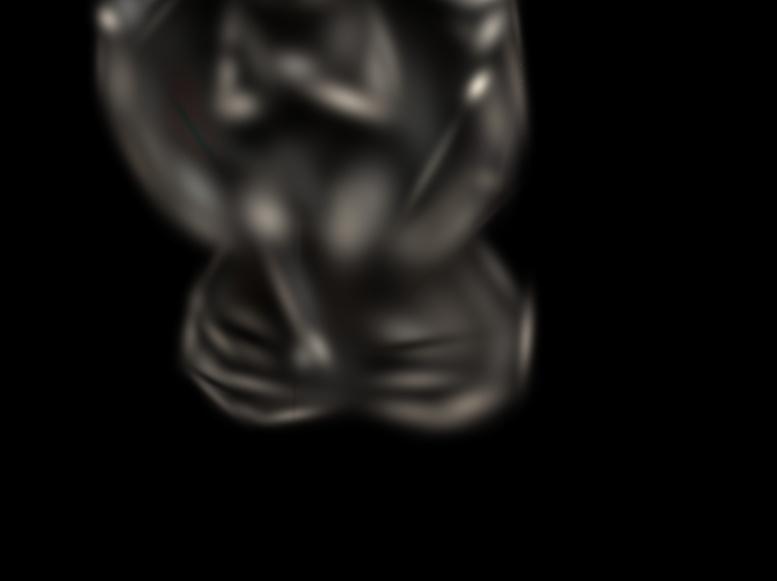}
        & \includegraphics[width=\tmpcolwidth,trim={40px 160px 40px 0px},clip]{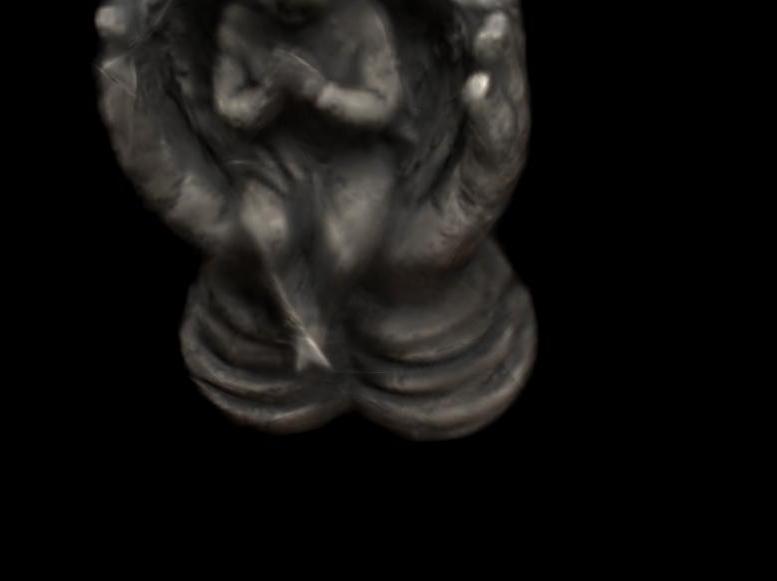}
        &
        \includegraphics[width=\tmpcolwidth,trim={40px 160px 40px 0px},clip]{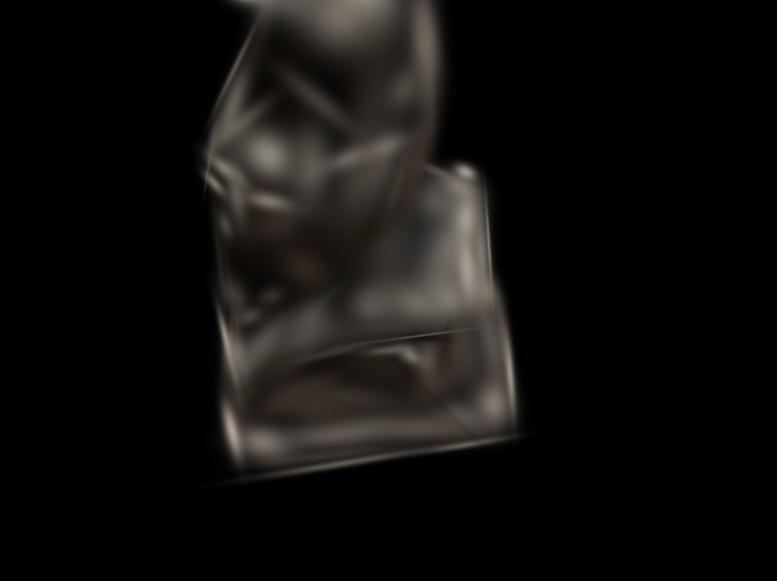}
        & \includegraphics[width=\tmpcolwidth,trim={40px 160px 40px 0px},clip]{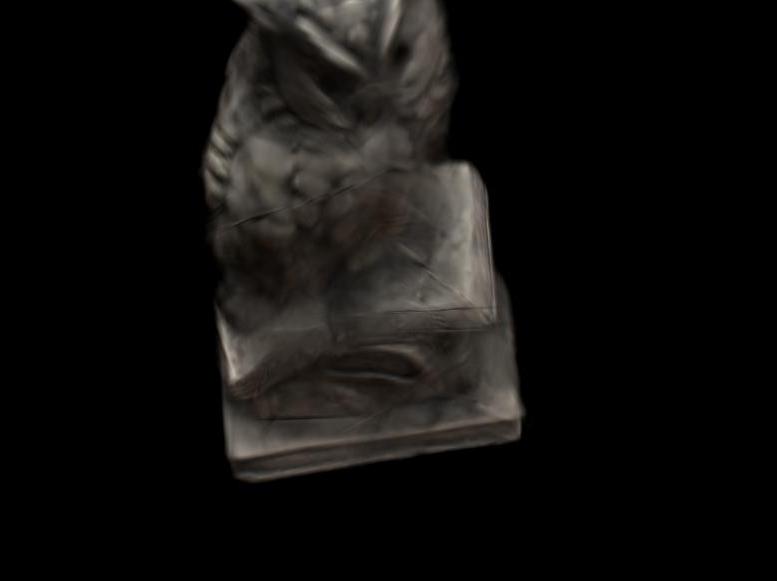}\\
                &
        \includegraphics[width=\tmpcolwidth,trim={40px 160px 40px 0px},clip]{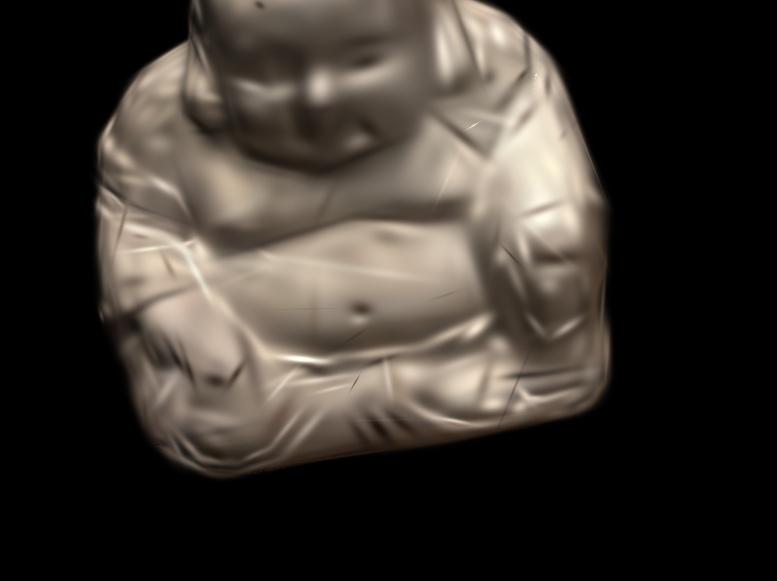}
        & \includegraphics[width=\tmpcolwidth,trim={40px 160px 40px 0px},clip]{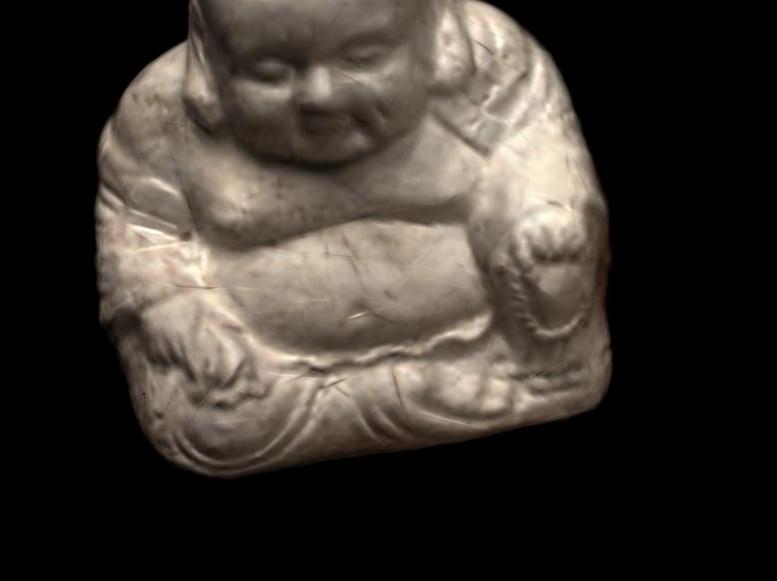}
        &
        \includegraphics[width=\tmpcolwidth,trim={40px 160px 40px 0px},clip]{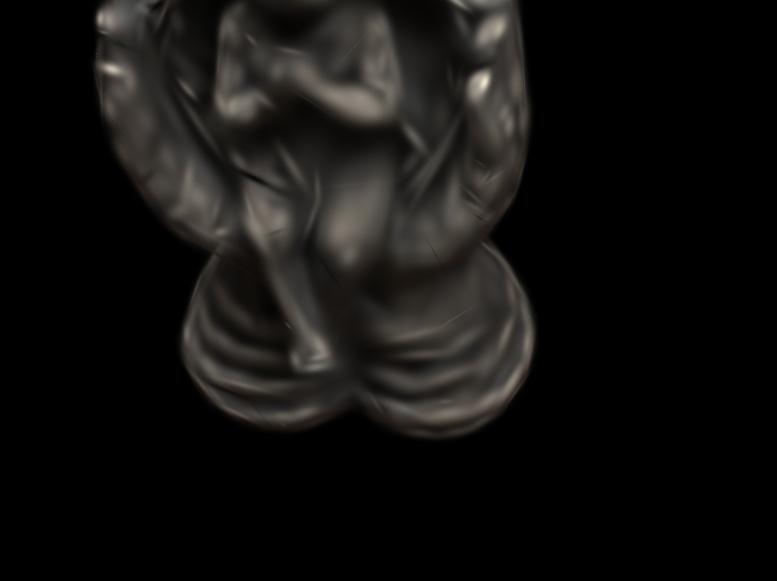}
        & \includegraphics[width=\tmpcolwidth,trim={40px 160px 40px 0px},clip]{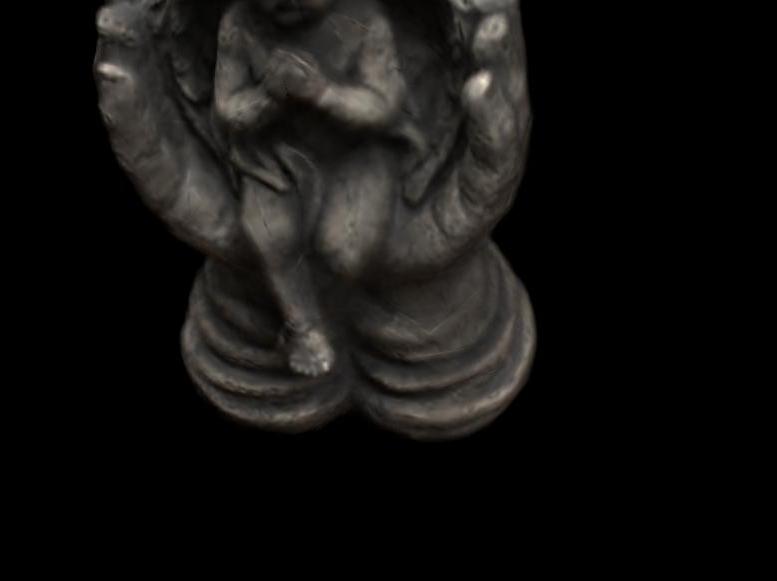}
        &
        \includegraphics[width=\tmpcolwidth,trim={40px 160px 40px 0px},clip]{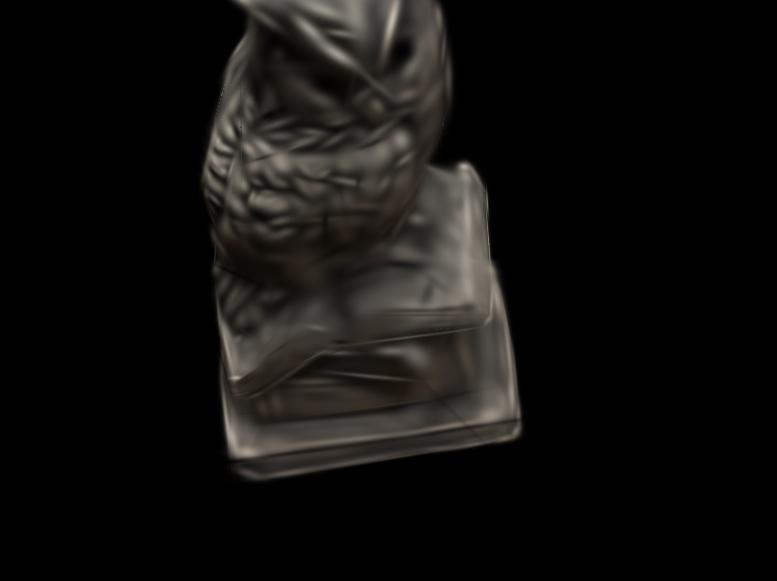}
        & \includegraphics[width=\tmpcolwidth,trim={40px 160px 40px 0px},clip]{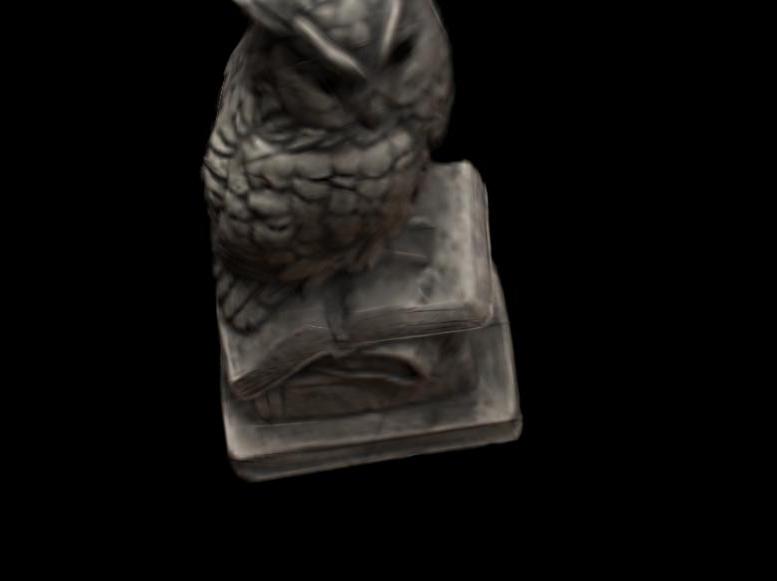}\\
                &
        \includegraphics[width=\tmpcolwidth,trim={40px 160px 40px 0px},clip]{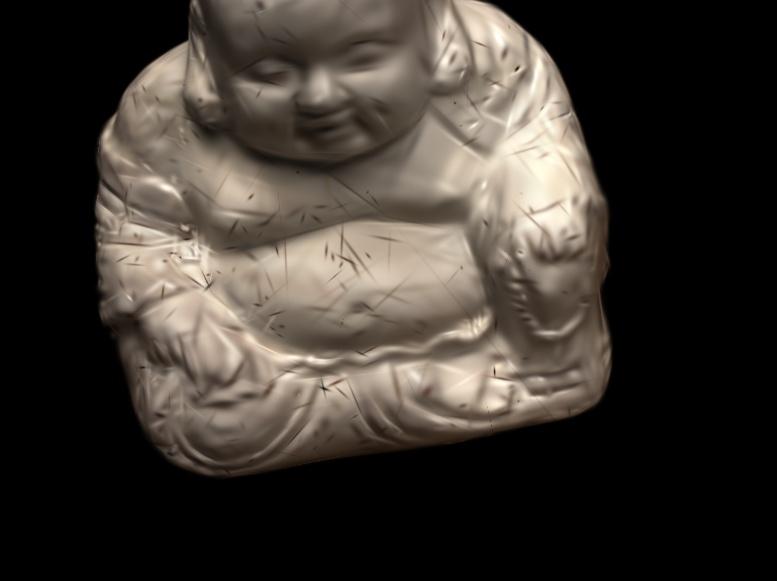}
        & \includegraphics[width=\tmpcolwidth,trim={40px 160px 40px 0px},clip]{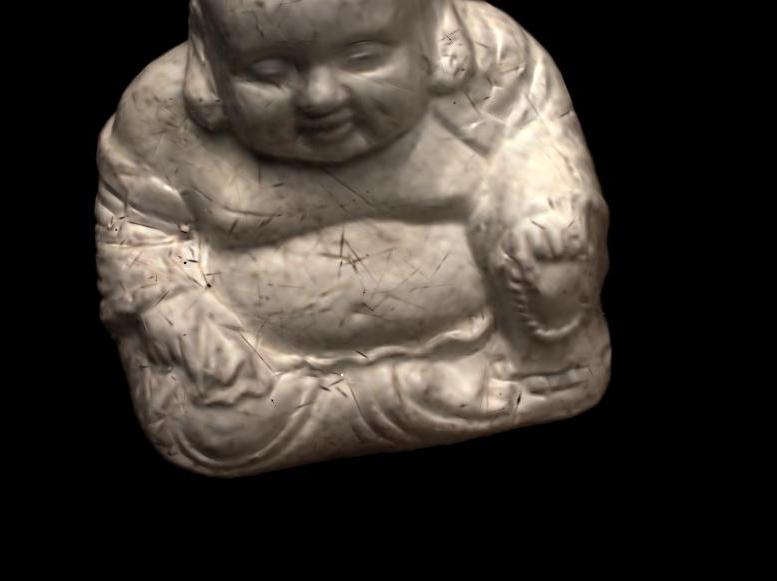}
        &
        \includegraphics[width=\tmpcolwidth,trim={40px 160px 40px 0px},clip]{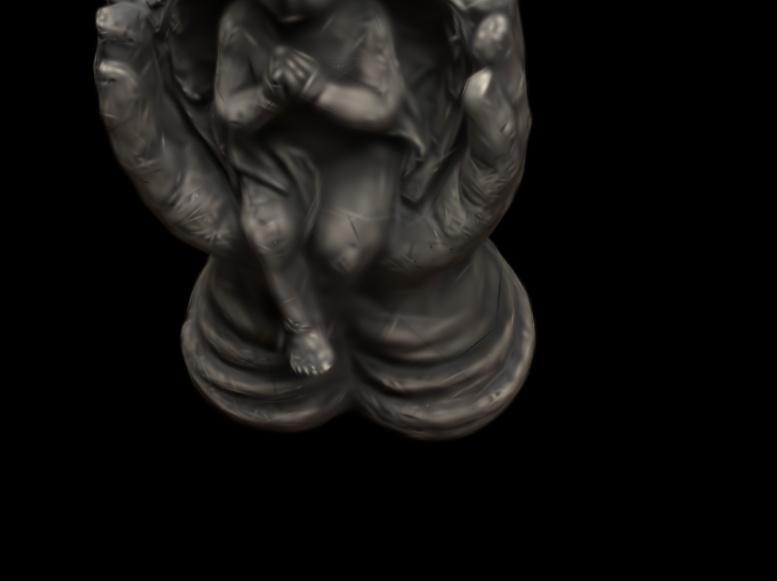}
        & \includegraphics[width=\tmpcolwidth,trim={40px 160px 40px 0px},clip]{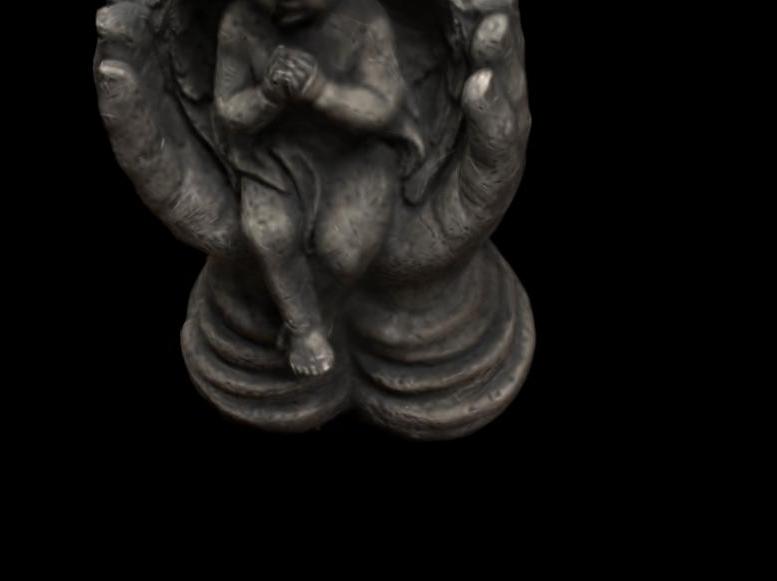}
        &
        \includegraphics[width=\tmpcolwidth,trim={40px 160px 40px 0px},clip]{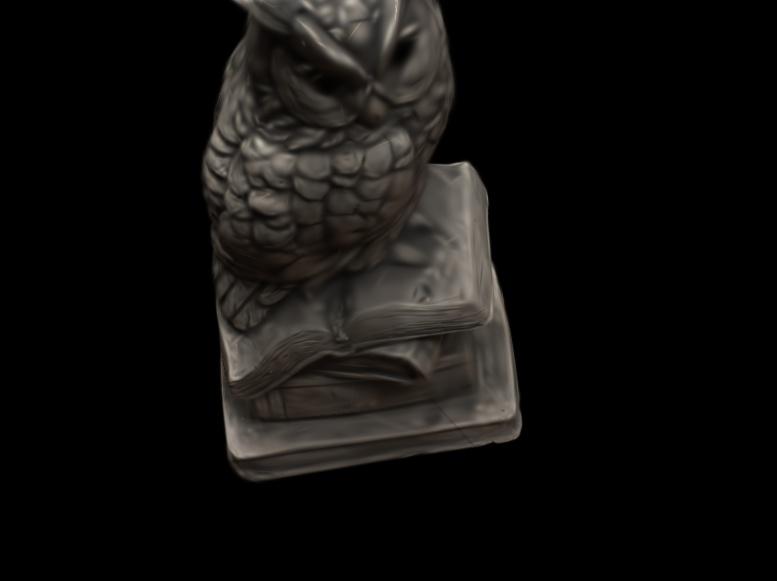}
        & \includegraphics[width=\tmpcolwidth,trim={40px 160px 40px 0px},clip]{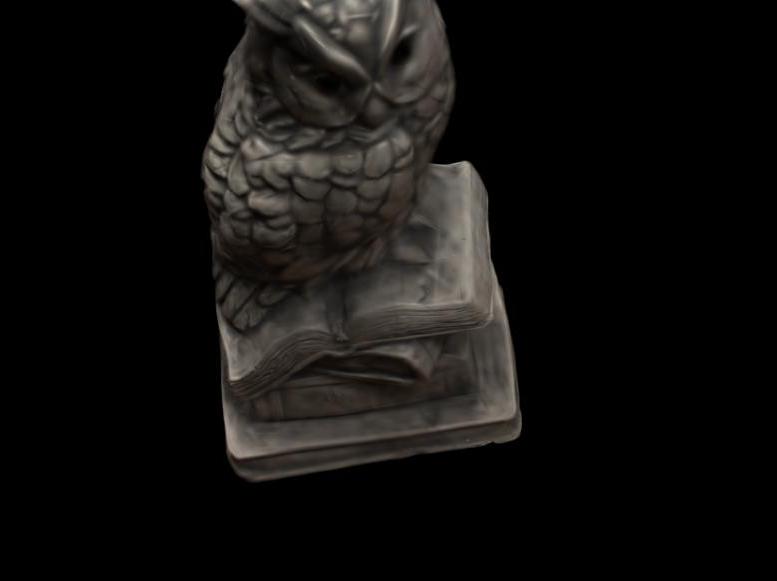}
    \end{tabular}
	\caption{\textbf{Novel view synthesis for discrete levels of detail.} We show test set renders of \ours and 2DGS models in three settings of levels of detail: with 128, 512, and 2048 Gaussians. We show our results on the DTU scenes not shown in the main paper.\label{fig:lod_supp_dtu}}
    \vspace{-1em}
\end{figure*}

\end{document}